\documentclass{article}

\usepackage{preprint}

\usepackage[utf8]{inputenc}
\usepackage[T1]{fontenc}
\usepackage{url}
\usepackage{microtype}
\usepackage{xcolor}

\usepackage{amsmath}
\usepackage{amsfonts}
\usepackage{amssymb}
\usepackage{mathtools}

\usepackage{booktabs}
\usepackage{longtable}
\usepackage{multirow}
\usepackage{makecell}

\usepackage{graphicx}
\usepackage{float}
\usepackage{caption}
\usepackage{subcaption}

\usepackage[
    colorlinks,
    citecolor=mila-purple,
    urlcolor=mila-purple,
    linkcolor=mila-purple
]{hyperref}
\usepackage[capitalize,noabbrev]{cleveref}

\newcommand{\tablestyle}[1][4pt]{%
  \small\setlength{\tabcolsep}{#1}\renewcommand{\arraystretch}{1.06}}
\definecolor{tabchange}{HTML}{505760}
\newcommand{\scorechange}[1]{{\scriptsize\textcolor{tabchange}{(#1)}}}

\definecolor{echodark}{HTML}{432355}
\definecolor{echolight}{HTML}{7E569B}
\newcommand{\EchoMoE}{%
  \textcolor{echodark}{E}%
  \textcolor{echolight}{cho}%
  \textcolor{echodark}{M}%
  \textcolor{echolight}{oE}%
}

\title{Communicate the Change: Efficient Expert Parallelism via Activation Compression}

\author{
  Fengyuan Liu$^{\text{*} \ 1,2}$\,\,
  Jay Gala$^{\text{*} \ 1,2}$\,\,
  Siva Reddy$^{1,2,4}$\,\,
  Dzmitry Bahdanau$^{1,2,3}$\,\,
  \vspace{0.4em} \\
  {\normalsize
  $^1$Mila - Quebec AI Institute \,\,$^2$McGill University \,\,$^3$Periodic Labs \,\,$^4$Canada CIFAR AI Chair
  \vspace{0.5em}\\

  \Affilfont
  $^*$Equal Contribution. Author ordering determined by coin flip. \\\ \ {Correspondence:  \texttt{\{fengyuan.liu,jay.gala\}@mila.quebec}}
  }
}

\begin{document}

\maketitle

\begin{abstract}
Scaling has driven the success of large language models (LLM), with sparse mixture-of-experts (MoE) architectures enabling greater parameter capacity without a proportional increase in computation. At large scales, MoE models rely on expert parallelism to distribute experts across devices. Each MoE layer requires two synchronous collective communication steps during inference: dispatching hidden states to the activated experts and combining the resulting expert outputs. As compute throughput has outpaced interconnect bandwidth, these synchronous collectives have become a major bottleneck to MoE scaling and efficiency. We address this challenge with \emph{\EchoMoE}, a lightweight architecture modification that retrofits existing MoE models, requiring $8{,}500\times$ fewer training tokens than full pretraining. Motivated by the observation that transformer blocks update the residual stream incrementally, we use a hidden state from an earlier layer as a reference and communicate only the compressed difference between it and the current state, rather than communicating each hidden state in full. EchoMoE periodically refreshes these reference states and transmits only the compressed differences synchronously, reducing the traffic on latency-critical paths without significant computational overhead. EchoMoE reduces inter-node communication volume by $58\%$ and improves throughput by $22.6\%$, while preserving average performance within 3.8 percentage points across 18 benchmarks. EchoMoE further improves inference throughput by up to $2.38\times$ over bandwidth-constrained Ethernet, paving the way for wide-area distributed inference. EchoMoE also tolerates quantization of the communicated representations without measurable quality loss. We release an efficient implementation for applying EchoMoE to pretrained MoE models, together with an inference integration for vLLM.

\vspace{3mm}
{\small\sffamily \textbf{Webpage:} \href{https://mcgill-nlp.github.io/EchoMoE/}{\textbf{\texttt{mcgill-nlp.github.io/EchoMoE}}}} \smallskip \\
{\small\sffamily \textbf{Code Repository:} \href{https://github.com/McGill-NLP/EchoMoE}{\textbf{\texttt{github.com/McGill-NLP/EchoMoE}}}} \smallskip \\
{\small\sffamily \textbf{HuggingFace Models:} \href{https://huggingface.co/McGill-NLP/EchoMoE}{\textbf{\texttt{huggingface.co/McGill-NLP/EchoMoE}}}}

\end{abstract}

\section{Introduction}
\label{sec:intro}

\begin{figure}[t]
\centering
\includegraphics[width=0.56\textwidth]{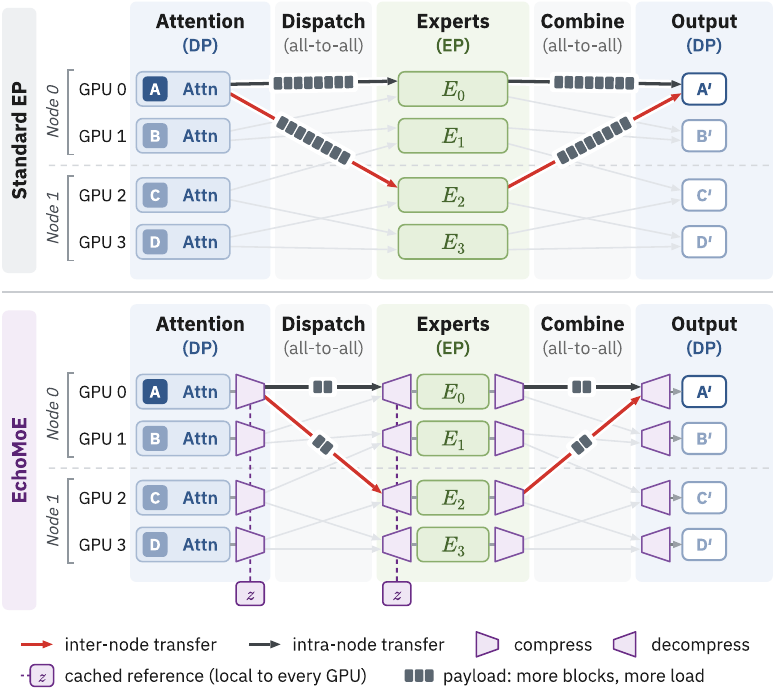}
\includegraphics[width=0.42\textwidth]{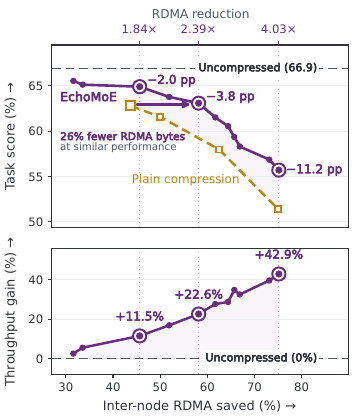}
\caption{\textbf{EchoMoE reduces EP communication and improves inference throughput.}
\textbf{Left:} Standard EP dispatches full activations to selected experts and returns full expert outputs to each token's owner rank. EchoMoE instead compresses dispatch activations using cached cross-layer references and compresses expert outputs before combine. Token A’s activation communication path is highlighted.
\textbf{Right:} EchoMoE saves $58\%$ of RDMA bytes and increases throughput by $22.6\%$ at a $3.8$ pp task score drop. At similar task performance, EchoMoE sends $26\%$ less RDMA traffic than plain compression, which directly compresses dispatch activations. RDMA savings include reference transfers at EP32 over $4$ nodes, with EP communication in BF16. Throughput uses EP32, $128$ requests per GPU, $4096$ input and $256$ output tokens and balanced expert assignment.}
\label{fig:pareto}
\end{figure}

Scaling has become a central paradigm in Large Language Models (LLM) development, with empirical scaling laws driving a continued push toward larger models \citep{kaplan2020scalinglawsneurallanguage, hoffmann2022trainingcomputeoptimallargelanguage}. Mixture-of-Experts (MoE) model \citep{shazeer2017outrageouslylargeneuralnetworks} enable substantial growth in parameter count while keeping per-token computation relatively modest through sparse expert activation.
Serving frontier MoE language models such as Kimi K3 \citep{kimiteam2026kimik3openfrontier} and DeepSeek-V4 \citep{deepseekai2026deepseekv4highlyefficientmilliontoken} requires distributing experts across many accelerators (e.g., GPUs) using Expert Parallelism (EP) \citep{rajbhandari2022deepspeedmoeadvancingmixtureofexpertsinference}. In standard EP (\Cref{fig:pareto} top left), at every MoE layer, each token's activation is sent to the devices hosting its selected experts (\emph{dispatch}), and the expert outputs are sent back (\emph{combine}). When the model is sharded across multiple nodes, both transfers carry the hidden state across the network (RDMA). For DeepSeek-V3 \citep{deepseekv3} served with 128 GPUs (EP128), a single $128$K-token prefill moves up to $812$\ GiB of activations between nodes \citep{Zhao_2025} and these collectives account for $60\%$ of transformer-layer latency, and still $32\%$ after overlapping communication with computation \citep{Perplexity_Research_2025}. As accelerator compute grows faster than interconnect bandwidth \citep{10477550}, EP communication increasingly limits MoE throughput and latency.

Prior work reduces this cost along two axes. The first redesigns the model architecture to reduce communication volume, either by communicating lower-dimensional activations or by organizing experts to localize more traffic within a node \citep{elango2026latentmoeoptimalaccuracyflop,jin2025bigmaccommunicationefficientmixtureofexpertsmodel,abdurrahman2026federationexpertscommunicationefficient,cai2025shortcutconnectedexpertparallelismaccelerating}. These designs are effective, but they require architectural changes before pretraining and cannot be directly applied to existing pretrained MoE models. The second leaves the model unchanged and reduces the cost of each communication through faster collective kernels or reduced-precision activations \citep{hwang2023tuteladaptivemixtureofexpertsscale,deepep2025}. These methods apply to off-the-shelf pretrained MoE models and are orthogonal to EchoMoE, as they reduce the cost per communicated element rather than the number of elements communicated. Regardless of precision, every MoE layer still transfers all $d$ activation dimensions on both collective operations.

Transformer blocks update the residual stream incrementally, causing nearby MoE layers to receive highly correlated inputs \citep{fang2025fatefastedgeinference, eliseev2023fastinferencemixtureofexpertslanguage, lee2024infinigenefficientgenerativeinference, Yu_2026}. We therefore hypothesize that successive full-width activations communicated across nodes during EP are highly redundant. Given a cached activation from an earlier layer, much of the current activation is predictable. Standard EP nevertheless transmits the full $d$-dimensional activation over RDMA at every layer. In this paper, we investigate whether this redundancy can be exploited to reduce EP communication in existing MoE models without retraining from scratch.

We introduce EchoMoE, which periodically caches a full-width activation from an earlier layer and reuses it as a reference for subsequent MoE layers. Rather than transmitting the full current activation, EchoMoE communicates only a compressed change relative to the cached activation, substantially reducing inter-node RDMA traffic (\Cref{fig:pareto} bottom left, \Cref{fig:arch}).

We evaluate EchoMoE on Qwen3-30B-A3B \citep{qwen3technicalreport} across $97$ trained variants and $18$ benchmarks spanning knowledge \& reasoning, mathematics, coding and instruction following (\Cref{fig:pareto}). One of our frontier configurations ({$s=4$, $r_d=512$, $r_c=768$}) reduces inter-node RDMA traffic by $58\%$ at EP32 across four nodes, while mean task performance drops by $3.8$ percentage points (pp). When served with vLLM \citep{kwon2023efficientmemorymanagementlarge} at 128 requests per GPU, EchoMoE lowers prefill latency by $23\%$ and decode latency by $7\%$, yielding a $22.6\%$ throughput improvement. At 512 requests per GPU, the decode gain rises to $11\%$. EchoMoE also composes cleanly with FP8 quantization, with no additional loss in task performance.
We empirically validate the effectiveness of cross-layer references by showing that EchoMoE substantially outperforms plain compression at comparable RDMA. We also develop an effective procedure for searching the frontier configuration space. We further apply EchoMoE to four other MoE models, roughly halving their RDMA traffic with a $2.4$--$5.9$ pp task performance drop (\Cref{tab:domain-results}).

\section{Expert-Parallel Communication}
\label{sec:epcomm}

\subsection{Mixture of Experts and Expert Parallelism}
\label{subsec:moe-ep}

An MoE layer replaces the dense feed-forward network with $N$ expert networks and a router. For each token activation $x\in\mathbb{R}^{d}$, the router scores all $N$ experts and selects the $K$ highest where typically $K \ll N$. Let $S$ index the selected experts and $E_i$ denote expert $i$. The MoE output is
$y=\sum_{i\in S} w_i\,E_i(x)$,
where the router weights $w_i$ are normalized over the selected experts. Each expert is a gated MLP with intermediate width $f$, and the resulting output $y$ is added to the residual stream. Increasing $N$ increases model capacity while keeping computation proportional to $K$.

At frontier scale, the experts of a layer are distributed across accelerators using EP. Throughout this paper, we use EP to refer to expert parallelism combined with data parallelism (DP), i.e., the EP+DP setting. Experts are partitioned across ranks, while non-expert parameters are replicated. At EP degree $P$, each rank stores $N/P$ experts from every MoE layer. Each sequence is associated with an owner rank which holds its residual state and executes the non-expert computation locally.

Because the experts selected for a token may reside on remote ranks, every MoE layer requires two collective communications. During dispatch, the owner sends the token activation $x$ to the ranks hosting the selected experts, which compute $E_i(x)$. During combine, the expert outputs are returned to the owner which aggregates them to form $y$. Both transfers are typically implemented with all-to-all collectives \citep{lepikhin2020gshard,hwang2023tuteladaptivemixtureofexpertsscale,deepep2025}, which route each token only between its owner and the ranks hosting its selected experts rather than replicating it across all ranks. Thus, a model with $L$ MoE layers performs $2L$ communication rounds per forward pass, with each routed activation still transmitted at the full width $d$.

\subsection{Expert-Parallel Communication Overheads}
\label{subsec:cost}

Dispatch and combine introduce communication dependencies and latency. For DeepSeek-V3 \citep{deepseekv3} at EP128 on H100 nodes, they account for roughly $60\%$ of transformer-layer latency in isolation, and still $32\%$ once overlapped with computation through microbatching \citep{Perplexity_Research_2025}.
Because inter-node links are much slower than intra-node interconnects, inter-node transfers dominate communication cost. Let $T$ denote the number of tokens processed across the EP group, $b$ the bytes per activation dimension, and $\bar m$ the average number of remote nodes reached by each token per collective communication.\footnote{$\bar m$ counts distinct remote nodes, not experts. Hierarchical forwarding sends one activation per node in each direction, so several selected experts on one node cost one transfer. DeepEP V1's low-latency kernels instead send one payload per expert \citep{deepep2025,Zhao_2025}. We exclude intra-node transfers and protocol overhead.} Each round sends one width $d$ activation to each of those nodes, so the total inter-node activation traffic across $L$ MoE layers is:
\begin{equation}
C_{\mathrm{base}}(T)
=\underbrace{b\,d\,L\,\bar m\,T}_{\text{dispatch}}
+\underbrace{b\,d\,L\,\bar m\,T}_{\text{combine}}
=2\,b\,d\,L\,\bar m\,T
\label{eq:baseline-communication}
\end{equation}

\begin{table}[t]
\centering
\small
\setlength{\tabcolsep}{4pt}
\begin{tabular}{lrrrrr}
\toprule
& \multicolumn{2}{c}{Architecture}
& \multicolumn{3}{c}{Inter-node RDMA Communication} \\
\cmidrule(lr){2-3}\cmidrule(lr){4-6}
Model & $d$ & $L$ & MiB/token & GiB/128K & Transfer (ms) \\
\midrule
Qwen3-30B-A3B   & 2048 & 48 & 1.125 & 144 & \phantom{0}97 \\
Qwen3-235B-A22B & 4096 & 94 & 4.406 & 564 & 379 \\
DeepSeek-V3     & 7168 & 58 & 4.758 & 609 & 409 \\
\bottomrule
\end{tabular}
\caption{Inter-node activation traffic and transfer time for a 131{,}072-token prefill on an EP32 group of 4 nodes at $400$ Gb/s per GPU, computed from \Cref{eq:baseline-communication}. All three models \citep{qwen3technicalreport,deepseekv3} activate eight experts per token. We consider MoE layers only.}
\label{tab:communication-cost}
\end{table}

Lower precision reduces $b$ and expert placement reduces $\bar m$, but $d$ is fixed by the architecture and every transfer carries all of it. Since a token's $K$ selected experts can span at most $K$ nodes, $\bar m$ eventually saturates, after which increasing the EP group size no longer reduces per-token traffic.

Communication time depends on both payload and number of collective calls. Each call incurs a fixed startup cost and the time required to transmit its payload at the available bandwidth. \Cref{tab:communication-cost} reports the traffic volume and the lower-bound communication time during prefill for two Qwen3 models and DeepSeek-V3.\footnote{We use bfloat16 communication ($b=2$) and maximum fan-out so every token reaches all three remote nodes ($\bar m=3$). Fewer destinations reduce traffic proportionally. Only DeepSeek-V3 has a shared expert and we assume it is replicated on every node so it adds no inter-node traffic. These estimates are lower bounds since they assume peak bandwidth on a balanced network and exclude startup overhead.} During prefill, all prompt tokens are processed in batches, yielding large payloads for which bandwidth-dependent transfer time dominates. During decoding, each pass processes one token per sequence, producing much smaller payloads, memory bottleneck and startup overhead are comparatively more significant \citep{Zhao_2025,deepep2025} (\Cref{app:deepep-scaling}).

\section{Similarity and Predictability in Dispatched Activations}
\label{sec:redundancy}

Each transformer block adds its output to the residual stream rather than overwriting it, so a token's hidden state changes gradually with depth and nearby MoE layers read highly similar inputs. Yet conventional expert parallelism still communicates the full $d$-dimensional activation at every layer, even when much of it is predictable from an earlier state.

\Cref{tab:moe-input-similarity} quantifies the similarity and predictability. We measure the cosine similarity between a token's inputs to adjacent MoE layers and also fit a per-dimension affine predictor from the earlier input to the later one and evaluate it on held-out tokens. Across Qwen3, DeepSeek-V3, GLM-5.3 \citep{glm5team2026glm5vibecodingagentic} and Kanana-2 models ($L\geq47$), adjacent MoE inputs have a mean cosine similarity above $90\%$, and the predictor recovers over $80\%$ of the variance in the later layer input at $k=1$ and over $50\%$ at $k=4$. Other models such as Gemma 4, Nemotron and Moonlight ($L\leq30$) are less predictable ($59.7$--$72.7\%$ at $k=1$), yet stay within $5.9$ pp of their uncompressed models after adaptation (\Cref{tab:domain-results}). See \Cref{app:similarity} for details on how both quantities are measured.

\begin{table}[h]
\centering
\small
\begin{tabular}{lrrrrrrrrr}
\toprule
Model & $L$ & \multicolumn{4}{c}{Cosine similarity (\%)}
& \multicolumn{4}{c}{Predictability (\%)} \\
\cmidrule(lr){3-6}\cmidrule(lr){7-10}
& & $k=1$ & $4$ & $8$ & $16$ & $k=1$ & $4$ & $8$ & $16$ \\
\midrule
Qwen3-235B-A22B                 & 94 & 94.9 & 82.8 & 70.9 & 56.6 & 90.2 & 69.4 & 52.1 & 35.1 \\
DeepSeek-V3                     & 58 & 93.7 & 77.5 & 61.5 & 39.7 & 88.1 & 62.2 & 40.3 & 17.5 \\
GLM-5.3                         & 75 & 93.4 & 76.7 & 60.0 & 38.0 & 87.3 & 60.1 & 38.2 & 16.6 \\
Qwen3-30B-A3B     &  48 & 91.6 & 74.9 & 62.0 & 47.0 & 83.2 & 56.0 & 40.1 & 25.1 \\
Kanana-2-30B-A3B                & 47 & 90.9 & 68.5 & 46.7 & 22.5 & 82.9 & 52.0 & 28.9 & 8.9 \\
\midrule
Moonlight-16B-A3B               & 26 & 86.1 & 58.4 & 35.5 & 18.2 & 72.7 & 35.4 & 14.7 & 4.3 \\
Nemotron-3.5-Lightning-30B-A3B  & 23 & 79.3 & 49.0 & 28.3 & 11.0 & 64.5 & 27.8 & 12.2 & 3.8 \\
Gemma 4 26B-A4B                 & 30 & 72.8 & 39.6 & 20.6 & 8.3 & 59.7 & 22.6 & 9.0 & 2.9 \\
\bottomrule
\end{tabular}
\caption{Similarity and predictability of MoE inputs across layer separations $k$. Values are averaged over all pairs of MoE layers separated by $k$. Predictability measures the fraction of variance in the later-layer input explained by a per-dimension affine predictor fitted on the earlier-layer input.}
\label{tab:moe-input-similarity}
\end{table}

\section{EchoMoE}
\label{sec:method}

\begin{figure}[h]
\begin{center}
\includegraphics[width=\textwidth]{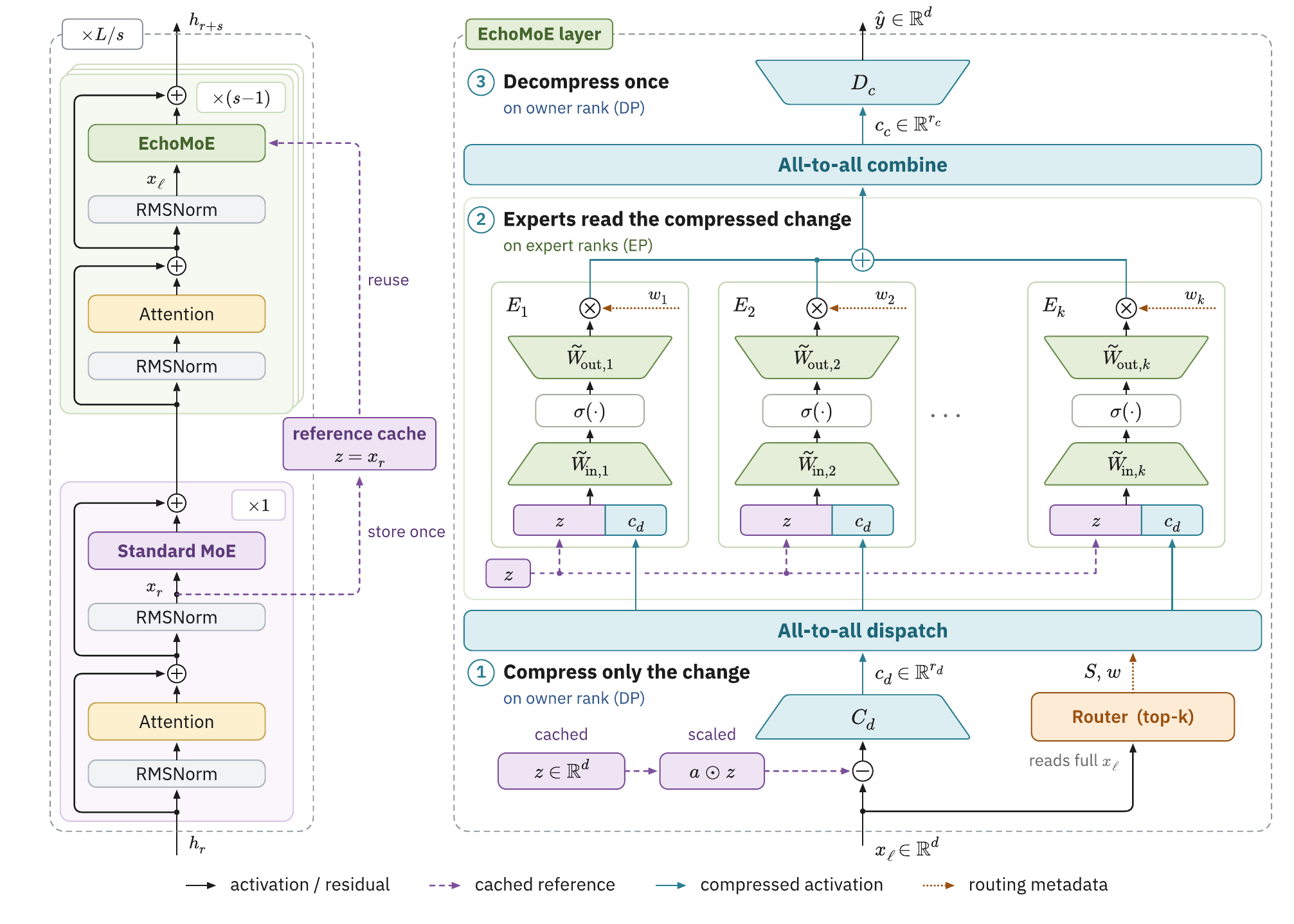}
\end{center}
\caption{\textbf{EchoMoE.} \textbf{Left:}
The $L$ MoE layers are partitioned into $L/s$ groups of $s$ consecutive layers. The first layer in each group serves as a reference layer, broadcasting its full-width state $x_\ell$, which every rank caches as $z$. The following $s-1$ layers in each group reuse this reference $z$.
\textbf{Right:} At each non-reference layer, the owner rank compresses and dispatches only the difference $x_\ell - a \odot z$ as $c_d \in \mathbb{R}^{r_d}$. Each expert applies
$\widetilde{W}_{\mathrm{in},i}$ to $\left[\begin{smallmatrix}z\\c_d\end{smallmatrix}\right]$ and $\widetilde{W}_{\mathrm{out},i}$, without explicit full-width input reconstruction or output materialization. Each node returns a single compressed state of width $r_c$, which is combined before the owner rank decompresses with $D_c$. The router reads the full $x_\ell$.}
\label{fig:arch}
\end{figure}

EchoMoE exploits the cross-layer similarity and predictability described in \Cref{sec:redundancy} by caching an earlier activation for each token on every rank and reusing it as a reference across several subsequent MoE layers. Between reference layers, only the change relative to its cached state is compressed and dispatched, and combine returns a compressed representation instead of full expert outputs.

Let $x_\ell\in\mathbb{R}^{d}$ denote a token's input to MoE layer $\ell$. EchoMoE partitions the $L$ MoE layers into groups of $s$ consecutive layers, where $s$ is the \emph{stride}. The first layer of each group is a reference layer. It broadcasts $x_\ell$ at the full width $d$ to every other rank, where every rank caches it as $z\in\mathbb{R}^{d}$. The $s-1$ layers that follow reuse $z$, and only need to communicate the compressed change. Full width hidden states are transmitted once per layer group rather than at every MoE layer.

\subsection{Compressing Dispatch}
\label{subsec:echo-dispatch}

For each non-reference layer, let $C_d\in\mathbb{R}^{r_d\times d}$ and $D_d\in\mathbb{R}^{d\times r_d}$ denote the linear compressor and decompressor for dispatch. Since every rank already holds $z$ from the latest reference layer, the owner rank compresses and dispatches only the component of $x_\ell$ not predicted from $z$. Specifically, a learned gain $a\in\mathbb{R}^{d}$ rescales $z$, and the compressor encodes the difference $x_\ell-a\odot z$ as:
\begin{equation}
c_d=C_d\left(x_\ell-a\odot z\right)
\label{eq:echo-dispatch}
\end{equation}
The owner dispatches $c_d\in\mathbb{R}^{r_d}$ to the ranks hosting the selected experts. At reference layers, dispatch is uncompressed. The owner first sends $x_\ell$ to the selected expert ranks, then asynchronously broadcasts $x_\ell$ to all ranks to cache as the reference state. Although this broadcast incurs additional traffic, it does not block the forward pass or introduce additional synchronization latency.

Each expert rank reconstructs $\hat x_\ell \in\mathbb{R}^{d}$ from the cached reference $z$ and dispatched code $c_d$:
\begin{equation}
\hat x_\ell=D_dc_d+Gz
\label{eq:echo-reconstruct}
\end{equation}
where $G\in\mathbb{R}^{d\times d}$ is a learned gain matrix. We learn $G$ separately from $a$, allowing the reconstruction to compensate for information lost when the difference $x_\ell-a\odot z$ is compressed. The rank then executes its selected experts on $\hat x_\ell$. For the plain compression baseline without the cached reference, dispatch compresses $x_\ell$ directly as $c_d=C_dx_\ell$ and a receiving rank recovers it as $\hat x_\ell=D_dc_d$.

\paragraph{Folding.} The decompressor $D_d$, the reconstruction gain $G$, and the gate and up projections $W_{\mathrm{gate},i}$ and $W_{\mathrm{up},i}$ of expert $i$ are all linear, so they fold into a single expert-specific projection:
\begin{equation}
\widetilde W_{\mathrm{in},i}=\left[\begin{smallmatrix}W_{\mathrm{gate},i}\\W_{\mathrm{up},i}\end{smallmatrix}\right]\begin{bmatrix}G&D_d\end{bmatrix}\in\mathbb{R}^{2f\times(d+r_d)}
\label{eq:echo-fold-in}
\end{equation}
The expert applies $\widetilde W_{\mathrm{in},i}$ directly to  $\left[\begin{smallmatrix}z\\c_d\end{smallmatrix}\right]$, avoiding explicit reconstruction of $\hat x_\ell$. Folding absorbs both $D_d$ and $G$ and increases the width of the expert's input projection from $d$ to $d+r_d$. A dense $G$ therefore incurs no additional serving-time operation. This expert-specific folding also makes $D_d$ and $G$ expert-specific, where $C_d$ and $a$ remain shared across experts within each layer.

\subsection{Compressing Combine}
\label{subsec:echo-combine}

Recent MoE models use fine-grained experts whose intermediate dimension is smaller than the hidden dimension, while activating more experts per token.
This suggests that expert outputs, which are updates to the residual stream, might be represented in a lower-dimensional space.

Let $C_c\in\mathbb{R}^{r_c\times d}$ and $D_c\in\mathbb{R}^{d\times r_c}$ denote the linear compressor and decompressor for combine, and let $S_j$ index the selected experts hosted by node $j$. Because combine and compression are linear, we apply both router weight $w_i$ and compressor $C_c$ locally, so node $j$ forms $c_c^{(j)}\in\mathbb{R}^{r_c}$:
\begin{equation}
c_c^{(j)}=\sum_{i\in S_j}w_i\,C_cE_i(\hat x_\ell)=C_c\sum_{i\in S_j}w_i\,E_i(\hat x_\ell)
\label{eq:echo-combine}
\end{equation}
The codes $c_c^{(j)}$ are combined at width $r_c$. The owner rank decompresses the combine sum as:
\begin{equation}
\hat y=D_c\sum_j c_c^{(j)}\in\mathbb{R}^{d}
\label{eq:echo-decombine}
\end{equation}
Combine compression is applied to every MoE layer. At a reference layer, the experts run on $x_\ell$.

\paragraph{Folding.} Similar to \Cref{eq:echo-fold-in}, the compressor $C_c$ and the down projection $W_{\mathrm{down},i}$ of expert $i$ are both linear so they compose into one expert-specific matrix:
\begin{equation}
\widetilde W_{\mathrm{out},i}=C_cW_{\mathrm{down},i}\in\mathbb{R}^{r_c\times f}
\label{eq:echo-fold-out}
\end{equation}
The expert applies $\widetilde W_{\mathrm{out},i}$ to its hidden state without materializing an output at width $d$. This allows the implicit $C_c$ to be expert-specific, while $D_c$ is shared across experts per layer.

\subsection{Communication Cost}
\label{subsec:echo-cost}

Let $m_{\max}$ denote the number of remote nodes in EP. A reference reaches all $m_{\max}$ nodes while a dispatch or combine reaches only the $\bar m$ that host selected experts, so $\bar m\le m_{\max}$. A group of $s$ layers sends one reference, $s-1$ dispatch and $s$ combine. Following \Cref{eq:baseline-communication}, the traffic is
\begin{equation}
C_{\mathrm{Echo}}(T)
=\underbrace{b\,d\,\tfrac{L}{s}\,m_{\max}\,T}_{\text{reference states}}
+\underbrace{b\,r_d\,\tfrac{s-1}{s}L\,\bar m\,T}_{\text{dispatch compressed}}
+\underbrace{b\,r_c\,L\,\bar m\,T}_{\text{combine compressed}}
\label{eq:echo-traffic}
\end{equation}
Reduced precision lowers $b$ in every term so it can be composed with EchoMoE. Caching $z$ costs negligible memory compared with the KV cache.

\subsection{Retrofitting an Existing Model}
\label{subsec:echo-adapt}

To retrofit an existing model without training from scratch, EchoMoE preserve its original behavior by first minimizing activation reconstruction error, then training end-to-end.

\paragraph{Initialization.} We run the uncompressed model over a set of documents and record the input activation $x_\ell$ and the combined expert output $y$ at every MoE layer. For combine, we reconstruct $y$.
For dispatch, reconstructing $x_\ell$ is sub-optimal because each expert uses $\hat x_\ell$ only through its gate and up projections. We therefore minimize reconstruction error after these projections. We define:
\begin{equation}R=\sum_{i}\rho_i\left(W_{\mathrm{gate},i}^{\top}W_{\mathrm{gate},i}+W_{\mathrm{up},i}^{\top}W_{\mathrm{up},i}\right)
\end{equation}
where $\rho_i$ denotes the fraction of tokens routed to expert $i$. Then $\lVert R^{1/2}v\rVert^2$ is the projection error induced by an activation reconstruction error $v$, averaged over router assignments. We then initialize $a$, $C_d$, $D_d$, $C_c$, and $D_c$ with closed-form solutions that minimize reconstruction error:
\begin{equation}
\min_{C_d,\,D_d}\ \mathbb{E}\left\lVert R^{1/2}\left(I-D_dC_d\right)\left(x_\ell-a\odot z\right)\right\rVert^2
\qquad\text{and}\qquad
\min_{C_c,\,D_c}\ \mathbb{E}\left\lVert\left(I-D_cC_c\right)y\right\rVert^2
\label{eq:echo-fit}
\end{equation}

\paragraph{End-to-End Training.} We perform the folding as described in \Cref{eq:echo-fold-in} and \Cref{eq:echo-fold-out} to obtain the expert-specific initialization for $\widetilde W_{\mathrm{in},i}$ and $\widetilde W_{\mathrm{out},i}$. Because this initialization minimizes reconstruction error independently at each layer, the resulting errors can compound across the $L$ layers. We therefore adapt the compressed model end-to-end using only a tiny amount of data, minimizing the KL divergence from the uncompressed model's output distribution and a small next-token cross-entropy loss.
For faster, lower-cost experimentation, we also use a development setting (dev mode) that skips folding and updates only the compression parameters (backbone frozen).

\section{Experiments}
\label{sec:exp}

\subsection{Experimental Setup}
\label{subsec:exp-setup}

\textbf{Model.} We fine-tune Qwen3-30B-A3B-Base \citep{qwen3technicalreport} on filtered Dolci-Instruct \citep{olmo2025olmo} with tool-call samples excluded. The resulting checkpoint is the uncompressed model, the starting point for retrofitting and the distillation teacher. To test other architectures, we also retrofit Gemma 4-26B-A4B \citep{gemmateam2026gemma4}, Kanana-2-30B-A3B \citep{kanana2}, Nemotron-3.5-Lightning-30B-A3B \citep{nvidia2026nemotron} and Moonlight-16B-A3B \citep{liu2025muonscalablellmtraining}, following the same procedure as for Qwen3-30B-A3B.

\textbf{Configurations and baselines.} A configuration $(s,r_d,r_c)$ specifies the reference stride and the dispatch and combine code widths. We consider $r_d,r_c\in\{256,512,768,1024,1280\}$ and $s\in\{4,6,8,12\}$. For analysis, we consider plain compression ($s = 0$; \Cref{subsec:echo-dispatch}) and dispatch-only and combine-only variants, which compress only dispatch or combine, respectively. We also compare with LSH-MoE \citep{nie2024lshmoecommunicationefficientmoetraining}, which communicates centroids of similar tokens within a layer. Its residual compensation adds offsets several times larger than the expert outputs they correct, so we scale the compensation by a learned factor initialized at $0$ and adapt it using the same training recipe as EchoMoE (\Cref{app:lsh}).

\textbf{Retrofitting.} We follow \Cref{subsec:echo-adapt} using a mixture of Dolmino \citep{olmo2025olmo} and Dolci-Instruct \citep{olmo2025olmo}. We use 4.32B tokens for adaptation and 1.04B in dev mode, approximately $8{,}500\times$ and $34{,}000\times$ fewer than Qwen3 pretraining respectively \citep{qwen3technicalreport}.
More details about the retrofitting pipeline can be found in \Cref{app:recipes}.

\textbf{Evaluation.} We evaluate task performance across $18$ benchmarks spanning knowledge and reasoning, mathematics, coding and instruction following.
We use MMLU \citep{hendryckstest2021}, HellaSwag \citep{zellers2019hellaswag} and GSM8K \citep{cobbe2021gsm8k} in zero-shot chat-CoT format as the development suite. \Cref{app:tasks} describes the full evaluation details.

\textbf{Hardware and communication.} We benchmark throughput on 4 DGX H100 nodes, each with 8 H100 SXM5 80GB GPUs and 8 ConnectX-7 400 Gb/s NICs.
\Cref{app:rdma-bytes} provide further details on the RDMA reduction calculation and analyze these reductions across different EP topologies.

\subsection{Task Performance and RDMA Reduction}
\label{subsec:exp-quality}

\begin{table}[t]
\centering\tablestyle[3pt]
\begin{tabular}{@{}ccccrrrr@{\hspace{10pt}}l@{}}
\toprule
\multicolumn{3}{c}{Configuration} &  & \multicolumn{5}{c}{Task Performance (\%) $\uparrow$} \\
\cmidrule(lr){1-3}\cmidrule(lr){5-9}
\makecell[b]{Stride\\$s$} & \makecell[b]{Dispatch\\$r_d$} & \makecell[b]{Combine\\$r_c$} & \makecell[b]{RDMA\\Reduction} & \makecell[b]{Knowledge\\\&\ Reasoning} & Math & Code & \makecell[b]{Instruction\\Following} & \makecell[b]{Overall\\\textcolor{tabchange}{($\Delta$)}} \\
\midrule
\multicolumn{9}{c}{\textbf{\textit{Qwen3-30B-A3B}}} \\
\multicolumn{3}{@{}l}{\textbf{Uncompressed} ($d=2048$)} & $1.00\times$ & 77.6 & 50.7 & 59.2 & 55.6 & 66.9 \\
\midrule
4 & 1280 & 1280 & $1.46\times$ & 76.2 & 49.1 & 58.1 & 54.5 & 65.5\,\scorechange{$-$1.4} \\
4 & 512 & 1280 & $1.84\times$ & 75.3 & 49.3 & 56.8 & 53.7 & 64.9\,\scorechange{$-$2.0} \\
4 & 512 & 1024 & $2.08\times$ & 74.3 & 47.8 & 55.2 & 53.5 & 63.8\,\scorechange{$-$3.1} \\
4 & 512 & 768 & $2.39\times$ & 73.6 & 47.2 & 54.4 & 53.3 & 63.1\,\scorechange{$-$3.8} \\
6 & 256 & 768 & $3.02\times$ & 68.3 & 42.5 & 50.4 & 50.1 & 58.3\,\scorechange{$-$8.6} \\
8 & 256 & 512 & $4.03\times$ & 64.2 & 40.9 & 50.7 & 51.0 & 55.7\,\scorechange{$-$11.2} \\
\addlinespace[2pt]
\midrule[\heavyrulewidth]
\addlinespace[4pt]
\multicolumn{9}{c}{\textbf{\textit{Gemma 4 26B-A4B}}} \\
\multicolumn{3}{@{}l}{\textbf{Uncompressed} ($d=2816$)} & $1.00\times$ & 72.5 & 41.4 & 53.9 & 57.3 & 61.2 \\
\midrule
4 & 768 & 1408 & $2.01\times$ & 67.7 & 37.1 & 50.7 & 54.1 & 56.8\,\scorechange{$-$4.4} \\
\midrule[\heavyrulewidth]
\multicolumn{9}{c}{\textbf{\textit{Kanana-2-30B-A3B}}} \\
\multicolumn{3}{@{}l}{\textbf{Uncompressed} ($d=2048$)} & $1.00\times$ & 68.2 & 41.5 & 53.4 & 53.0 & 58.5 \\
\midrule
4 & 512 & 1024 & $2.01\times$ & 66.5 & 37.9 & 49.5 & 52.5 & 56.1\,\scorechange{$-$2.4} \\
\midrule[\heavyrulewidth]
\multicolumn{9}{c}{\textbf{\textit{Nemotron-3.5-Lightning-30B-A3B}}} \\
\multicolumn{3}{@{}l}{\textbf{Uncompressed} ($d=2688$)} & $1.00\times$ & 71.6 & 50.5 & 53.0 & 55.0 & 62.8 \\
\midrule
4 & 768 & 1408 & $1.91\times$ & 67.9 & 41.7 & 51.4 & 54.1 & 58.3\,\scorechange{$-$4.6} \\
\midrule[\heavyrulewidth]
\multicolumn{9}{c}{\textbf{\textit{Moonlight-16B-A3B}}} \\
\multicolumn{3}{@{}l}{\textbf{Uncompressed} ($d=2048$)} & $1.00\times$ & 62.9 & 34.7 & 47.0 & 51.3 & 53.0 \\
\midrule
4 & 512 & 1024 & $1.99\times$ & 56.0 & 31.4 & 39.6 & 47.0 & 47.1\,\scorechange{$-$5.9} \\
\bottomrule
\end{tabular}
\caption{\textbf{Task performance and RDMA reduction after adaptation.} EchoMoE configurations with the backbone adapted: six for Qwen3-30B-A3B and one for each of the other four models. RDMA reduction compares estimated BF16 bytes at EP32 over four nodes using each model's measured routing fan-out. $\Delta$ shows the change in overall score relative to the original uncompressed version of each model, in percentage points. Results demonstrate EchoMoE's applicability across models and architectures. See \Cref{tab:allgroups} for frozen-backbone results and \Cref{tab:fp8} for FP8 results.}
\label{tab:domain-results}
\end{table}

\textbf{EchoMoE reduces RDMA traffic by $58\%$ with only $3.8$ pp drop in task performance.} The retrofitted $(4,512,768)$ configuration achieves a $2.39\times$ RDMA reduction (\Cref{tab:domain-results}). Milder compression reaches $1.84\times$ with a $2.0$ pp drop, and more aggressive compression reaches $4.03\times$ with an $11.2$ pp drop.
Per-configuration results are provided in \Cref{app:pertask}.

\textbf{EchoMoE outperforms baselines at comparable RDMA reductions.} At $2.08\times$ reduction, EchoMoE outperforms plain compression at $2.00\times$ by $5.0$ pp with the backbone frozen and $2.2$ pp after adaptation. At roughly $4\times$ RDMA reduction, adapted EchoMoE outperforms plain compression by $4.3$ pp. Across all matched comparisons, the frontier EchoMoE configuration achieves better performance with the same or less RDMA traffic (\Cref{app:matched}). Adapted with the same recipe, LSH-MoE loses $9.8$ pp at $1.59\times$ reduction, against $3.1$ pp for EchoMoE at $2.08\times$ (\Cref{app:lsh}).
Full-model adaptation consistently outperforms freezing the backbone, while FP8 transfer of the compressed representations incurs no measurable performance degradation (\Cref{app:recovery}, \Cref{tab:fp8}).

\subsection{Ablation Studies of Each Compression Path}
\label{subsec:exp-reference}

\begin{figure}[t]
\centering
\setlength{\abovecaptionskip}{4pt}
\includegraphics[width=\linewidth]{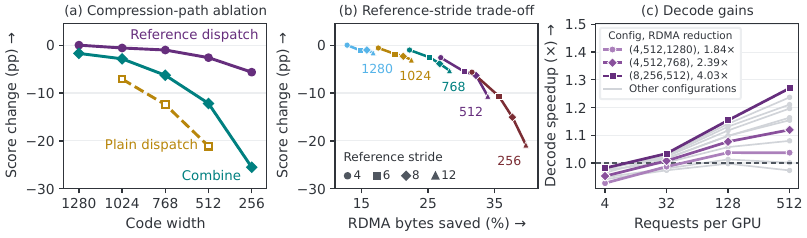}
\caption{
\textbf{(a)} Reference states substantially improve dispatch compression, consistently outperforming both plain dispatch and combine at matched code widths. Combine still outperforms plain dispatch, consistent with expert outputs being lower-dimensional, but requires larger code width. All results are single-path ablations, with the other communication path left uncompressed.
\textbf{(b)} At a fixed RDMA-byte reduction, smaller reference strides with narrower code widths preserve higher task performance.
\textbf{(c)} Decode gains increase across workloads with more requests per GPU. Input lengths are $128K$, $16K$, $4K$ and $1K$ tokens, respectively. Speedup uses time per output token (TPOT) for the $10\%$ of requests prefilled last. The dashed line marks the uncompressed model.
}
\label{fig:results-overview}
\end{figure}

\textbf{Reference states substantially improve dispatch compression.} At width $512$, plain dispatch loses $21.1$ pp whereas reference dispatch at stride $4$ loses only $2.6$ pp (\Cref{fig:results-overview}a). The reference removes the input's dominant directions, so a code on the residual drops $3.4\times$ less of the input's energy (\Cref{app:residual-spectrum}). This consistent advantage translates into a better RDMA–performance frontier for EchoMoE as shown in \Cref{fig:pareto}.
\Cref{tab:dispatch-grid,tab:correction} provide the ablation details.

\textbf{Expert outputs are more compressible than dispatch activations.} At the same code width, combine incurs substantially less degradation than plain dispatch, consistent with expert outputs having a lower-dimensional structure. Combine compression does require larger widths than reference-based dispatch to achieve comparable performance. We also explored augmenting combine with additional side information analogous to the dispatch reference state but observed similar performance (\Cref{app:combine-correction}). We therefore use direct combine compression throughout.

\textbf{Smaller strides with narrower codes outperform larger strides with wider codes.} The benefit of frequent reference updates grows as the dispatch code narrows (\Cref{fig:results-overview}b). At comparable RDMA savings, $(s,r_d)=(4,512)$ consistently outperforms $(8,768)$ across combine widths (\Cref{tab:refresh-budget}), showing that communication is better spent on more frequent references than wider dispatch codes.

\textbf{Efficient search for high-performing configurations.}
The isolated dispatch and combine performance drops are approximately additive (\Cref{fig:additivity}), allowing us to estimate joint performance by summing their dev mode drops (\Cref{app:search}).
We retrospectively evaluate the estimation by selecting its highest-scoring configuration at each RDMA target, and comparing the selection with the the best tested configuration.
The selected configuration is within $0.5$ pp of the best tested configuration for $94\%$ of RDMA targets, with a maximum gap of $0.7$ pp.
This enables compute-efficient search for high-performing joint
configurations without exhaustively training in the full Cartesian product.

\subsection{Throughput, Latency and Inference Speedup}
\label{subsec:exp-speed}

\begin{figure}[t]
\centering
\includegraphics[width=0.9\linewidth]{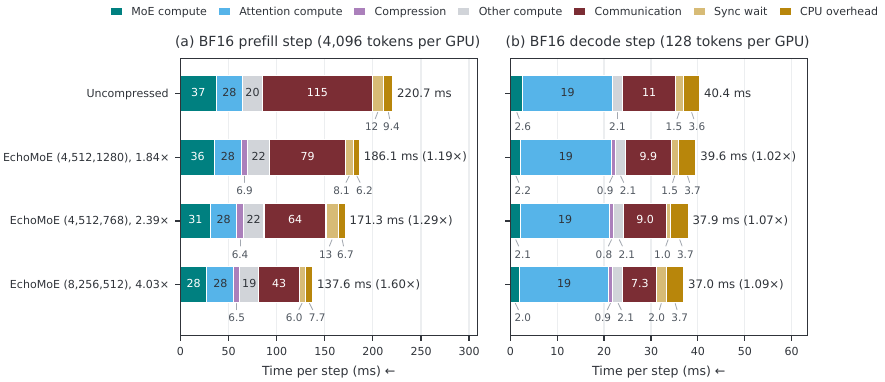}
\caption{\textbf{EchoMoE shortens prefill mainly by reducing communication.} Time per step at $4096/256$ with $128$ requests per GPU, BF16 communication at EP32 with balanced expert assignment. Communication includes the reference collective. Labels give each component's time in ms and each step's speedup over the uncompressed model. \Cref{app:serving-profile} profiles other workloads.}
\label{fig:breakdown-main}
\end{figure}

\textbf{Prefill savings translate into higher throughput.} With $4096$ input tokens, $256$ output tokens and $128$ requests per GPU, EchoMoE improves throughput by $22.6\%$ at $2.39\times$ RDMA reduction and by $42.9\%$ at $4.03\times$ (\Cref{fig:pareto} bottom right). Time to first token (TTFT) falls by $23.0\%$ and $36.9\%$, respectively. Profiling attributes these gains to reduced communication and expert computation costs: for $(4,512,768)$, prefill communication drops from $115$ to $64$ ms and expert computation from $37$ to $31$ ms (\Cref{fig:breakdown-main}a). The gains compose with FP8 communication, reach $2.38\times$ over TCP and hold under online serving of real prompts (\Cref{app:fp8-serving,app:serving-network,app:serving-online}). At about $2\times$ RDMA reduction, the adapted Kanana-2 and Gemma 4 improve throughput by $15\%$ and $14\%$ (\Cref{app:serving-adapted}).

\textbf{Decode gains depend on the workload.} Decode speedups grow with requests per GPU rather than context length, reaching $1.27\times$ at $512$ requests per GPU for $(8,256,512)$, while lower-concurrency workloads remain close to the uncompressed model (\Cref{fig:results-overview}c). At $128$ requests per GPU, communication takes only $11$ of the $40$ ms decode step (\Cref{fig:breakdown-main}b). As the number of tokens per step increases, communication constitutes a larger fraction of decode time, allowing reduced RDMA traffic to translate more directly into latency savings (\Cref{app:deepep-scaling}). \Cref{app:serving-setup,app:serving-workloads,app:serving-load} describe the throughput benchmarking protocol and workload comparisons.

\textbf{The implementation and throughput gains extend to larger MoE architectures.} With $s=4$, $r_d=d/4$ and $r_c=d/2$, throughput improves by $1.24\times$ on Qwen3-235B-A22B, $1.15\times$ on DeepSeek-V3 and $1.09\times$ on Kimi-K2.6 (\Cref{app:serving-large}). Due to limited compute for retrofitting, these tests use full-sized models with random weights and report throughput improvements only.

\section{Related Work}

\textbf{Communication-efficient MoE architectures.} Several works reduce MoE communication by modifying the model architecture. LatentMoE \citep{elango2026latentmoeoptimalaccuracyflop} and BigMac \citep{jin2025bigmaccommunicationefficientmixtureofexpertsmodel} project routed activations into a lower-dimensional expert space, reducing dispatch and combine width. Multi-Head LatentMoE \citep{cui2026multihead} and Federation of Experts \citep{abdurrahman2026federationexpertscommunicationefficient} instead restructure expert partitioning and communication, while ScMoE \citep{cai2025shortcutconnectedexpertparallelismaccelerating} uses earlier activations to overlap expert communication with computation. These methods are designed for training new architectures from scratch. In contrast, EchoMoE retrofits existing MoEs, using earlier activations as references to compress communication while reconstructing current expert inputs.

\textbf{Efficient expert parallelism.} DeepSpeed-MoE \citep{rajbhandari2022deepspeedmoeadvancingmixtureofexpertsinference} uses hierarchical all-to-all communication, while Tutel \citep{hwang2023tuteladaptivemixtureofexpertsscale} combines adaptive parallelism with communication pipelining to improve interconnect efficiency. At the transport level, DeepEP \citep{deepep2025} provides specialized dispatch and combine kernels with low-precision communication. These approaches optimize the movement of expert-parallel traffic. EchoMoE is complementary to these systems: it reduces the communication traffic and is integrated with DeepEP, while transport precision further reduces the cost per value. \cite{fang2025fatefastedgeinference, eliseev2023fastinferencemixtureofexpertslanguage, Yu_2026} exploits the cross-layer similarity in MoE models to accelerate expert fetching or offloading.

\textbf{Activation and residual compression.} Learned bottlenecks exploit low-dimensional activation structure; ResBM applies this idea to pipeline communication while preserving a low-rank identity path \citep{aboudib2026resbm}. Other methods exploit shared information: DIANA compresses gradients relative to estimates maintained across optimization steps \citep{mishchenko2019diana}, while LSH-MoE communicates centroids of similar tokens within a layer \citep{nie2024lshmoecommunicationefficientmoetraining}. EchoMoE instead combines a learned bottleneck with an earlier-layer reference for the same token, exploiting cross-layer predictability within a forward pass. Unlike LSH-MoE, which adds token--centroid residuals to expert outputs, EchoMoE uses compressed residuals to reconstruct current expert inputs.

\section{Conclusion and Future Work}

We introduced EchoMoE, a lightweight architectural retrofit for reducing EP communication in existing MoE models. EchoMoE exploits the cross-layer predictability of MoE inputs by periodically caching a full-width activation from an earlier layer as a reference, communicating only a compressed change relative to this reference on the dispatch path, and compressing expert outputs on the combine path. On Qwen3-30B-A3B, EchoMoE reduces inter-node RDMA traffic by $58\%$ with a $3.8$ pp drop in task performance and improves inference throughput by $22.6\%$, while requiring approximately $8{,}500\times$ fewer training tokens than pre-training.
Overall, EchoMoE establishes cross-layer predictability as a practical mechanism for substantially reducing EP communication in existing MoE models. While this work focuses on retrofitting pretrained models, EchoMoE could also be incorporated as a novel architecture for pre-training, where reducing representational redundancy may enable stronger performance. Its communication savings should also extend naturally to training.

\subsection*{AI use statement}

In this work, we used generative AI tools to polish the writing, improve the style of plots, and help with the implementation details of our method. We did not use these tools to generate ideas or propose experiments. We have reviewed all AI-assisted work. LLM-generated code was verified and tested for correctness by 2 authors. We take responsibility for the final content of this work, including all text, claims, and artifacts produced with the aid of generative AI.

\subsubsection*{Acknowledgments}
We acknowledge the support of the 2026 Google TPU Research \& Education Awards (TPU Builders Research Program). Siva Reddy is supported by the Canada CIFAR AI Chairs program. We thank the Mila IDT team and the Digital Research Alliance of Canada for providing the computational resources used in this work. We also thank members of McGill University, Mila, and ServiceNow Research, especially Tom\'as Vergara Browne and Spandana Gella.

\bibliographystyle{preprint}
\bibliography{references}

\crefalias{section}{appendix}
\appendix

\section{Cross-Layer Similarity of MoE-Layer Inputs}
\label{app:similarity}

\Cref{tab:moe-input-similarity} compares a token's input to one MoE layer with its input to an earlier MoE layer. The activation we compare is the normalized input to the MoE layer, which is what dispatch sends at the full width $d$. Each model runs one forward pass over $256$ documents from the Dolci Instruct \citep{olmo2025olmo}, tokenized by its own tokenizer and truncated to $2048$ positions.
Most documents are shorter than $2048$ tokens, so each model contributes about $137$K token positions. We split the documents into two subsets: $74$K tokens for fitting the predictor defined below and $63$K held-out tokens for evaluating it. The first position of every document is excluded because it acts as an attention sink \citep{xiao2024efficient,gu2025attentionsink} with a hidden-state norm orders of magnitude larger than at any other position.

Let $x_{\ell}\in\mathbb{R}^{d}$ denote a token's input to MoE layer $\ell$ and let $\ell-k$ be an earlier MoE layer. For each such pair, we compute a cosine similarity and a predictability on the held-out tokens, then average each over all pairs at separation $k$. The cosine similarity between the two inputs is
\begin{equation}
\cos\left(x_{\ell-k},x_{\ell}\right)=\frac{\langle x_{\ell-k},x_{\ell}\rangle}{\lVert x_{\ell-k}\rVert\,\lVert x_{\ell}\rVert}
\label{eq:similarity-cos}
\end{equation}
Cosine similarity measures only direction, while reconstructing $x_{\ell}$ from $x_{\ell-k}$ also requires matching its scale. We therefore also measure predictability, the fraction of variance in $x_{\ell}$ explained by a per-dimension affine predictor applied to $x_{\ell-k}$. We fit one scale $a_i$ and one shift $b_i$ per dimension $i$ by least squares on the fitting tokens and evaluate the predictor on the held-out tokens as
\begin{equation}
R^2 = 1-\frac{\sum_{t,i}\left(x^{(t)}_{\ell,i}-a_i x^{(t)}_{\ell-k,i}-b_i\right)^2}{\sum_{t,i}\left(x^{(t)}_{\ell,i}-\bar x_{\ell,i}\right)^2}
\label{eq:similarity-r2}
\end{equation}
where $t$ indexes the held-out tokens and $\bar x_{\ell,i}$ is the mean of dimension $i$ over the fitting tokens. This mean acts as a baseline and requires no information from $x_{\ell-k}$.

\subsection{Spectrum of the Residual After the Reference}
\label{app:residual-spectrum}

\begin{figure}[b]
\centering
\includegraphics[width=\linewidth]{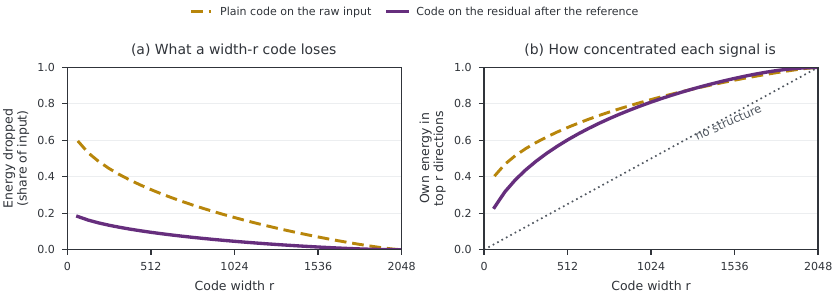}
\caption{\textbf{Compressing the residual loses much less of the input than compressing the raw input.} \textbf{(a)} Energy a width-$r$ code drops, as a share of the input. \textbf{(b)} Each signal's own energy in its top $r$ directions. Qwen3-30B-A3B teacher, all $48$ MoE layers, residuals one to three layers after a stride-$4$ reference.}
\label{fig:residual-spectrum}
\end{figure}

\textbf{Predictability sets how much remains to be sent.} On Qwen3-30B-A3B at stride $4$, the residual $x_\ell-a\odot z$ holds $15.4\%$, $25.9\%$ and $33.3\%$ of the input's energy at the first, second and third layer after the reference. We measure both signals on the teacher's activations over all $48$ MoE layers, using second-moment spectra fitted on the activations cached at the $512$K token positions used for initialization.

\textbf{The residual keeps a decaying spectrum.} Its top $512$ of $2048$ directions hold $60\%$ of its energy, against $25\%$ for a signal without structure, and its top $1024$ hold $81\%$ (\Cref{fig:residual-spectrum}b). The raw input is more concentrated than the residual, with a few dominant directions holding much of its energy (\Cref{tab:residual-spectrum}). The reference largely removes these directions, on which a plain code otherwise spends its width.

\textbf{A code on the residual loses $3.3$--$3.8\times$ less of the input.} A width-$512$ code on the residual drops $9.6\%$ of the input's energy, against $33.0\%$ for the same width on the raw input. At width $256$, the code drops $13.4\%$ against $44.5\%$ (\Cref{tab:residual-spectrum}). This is consistent with \Cref{fig:results-overview}a, where reference dispatch at width $512$ loses $2.6$ pp and plain dispatch loses $21.1$ pp.

\begin{table}[t]
\centering\tablestyle[6pt]
\begin{tabular}{@{}rrrrr@{}}
\toprule
\multirow{2}{*}{\makecell{Code width\\$r$}} & \multicolumn{2}{c}{Own energy in top $r$} & \multicolumn{2}{c}{Input energy dropped} \\
\cmidrule(lr){2-3}\cmidrule(lr){4-5}
 & Raw input & Residual & Plain code & Residual code \\
\midrule
128 & $0.47$ & $0.32$ & $0.531$ & $0.162$ \\
256 & $0.55$ & $0.44$ & $0.445$ & $0.134$ \\
512 & $0.67$ & $0.60$ & $0.330$ & $0.096$ \\
1024 & $0.82$ & $0.81$ & $0.177$ & $0.046$ \\
\bottomrule
\end{tabular}
\caption{\textbf{Spectra of the raw input and of the residual after the reference.} Own energy is the share of each signal's energy in its top $r$ directions. Input energy dropped is the energy a width-$r$ code loses, as a share of the input. Spectra are fitted and evaluated on the same activations cached for initialization, so the top-$r$ shares are slightly optimistic for both signals alike.}
\label{tab:residual-spectrum}
\end{table}

\section{Training and Evaluation Details}
\label{app:recipes}

We use Megatron-LM \citep{megatron-lm} for fine-tuning and compression training, and vLLM \citep{kwon2023efficientmemorymanagementlarge} for task evaluation. All compressed checkpoints are evaluated after folding the learned projections into the experts as described in \Cref{subsec:echo-adapt}.

\subsection{Model and Compression Configurations}

\textbf{Uncompressed models.} We fine-tune Qwen3-30B-A3B-Base \citep{qwen3technicalreport} on Dolci-Instruct \citep{olmo2025olmo} with tool-call samples excluded, using the Qwen3 Instruct chat template and tokenizer. The filtered dataset contains $1.9$M conversations and $1.16$B tokens. The resulting checkpoint provides the initial model weights and distillation targets for compression. Similarly, we fine-tune the base checkpoints of Gemma 4 26B-A4B \citep{gemmateam2026gemma4}, Kanana-2-30B-A3B \citep{kanana2}, Nemotron-3.5-Lightning-30B-A3B \citep{nvidia2026nemotron} and Moonlight-16B-A3B \citep{liu2025muonscalablellmtraining}, each with its own chat template and tokenizer. \Cref{tab:sft-settings} lists the fine-tuning hyperparameters.

\begin{table}[H]
\centering\tablestyle[8pt]
\begin{tabular}{@{}ll@{}}
\toprule
Setting & Value \\
\midrule
Loss function & Cross-entropy \\
Training epochs & $2$ \\
Training steps & $2230$ \\
Global batch size & $128$ packed sequences \\
Sequence length & $8192$ tokens \\
Training tokens & $2.32$B \\
Supervised tokens & $1.45$B \\
\midrule
Learning rate (peak $\to$ final) & $2\times10^{-5} \to 2\times10^{-6}$ \\
Warmup steps & $200$ \\
Weight decay & $0.1$ \\
Gradient clipping & $1.0$ \\
\midrule
Precision & BF16 \\
Random seed & $42$ \\
\bottomrule
\end{tabular}
\caption{\textbf{Fine-tuning hyperparameters for the uncompressed models.} Only response tokens are supervised. Step and token counts cover both epochs are for Qwen3-30B-A3B.}
\label{tab:sft-settings}
\end{table}

\textbf{Compression configurations.} We consider reference strides $s\in\{0,4,6,8,12\}$ and code widths $r_d,r_c\in\{256,512,768,1024,1280\}$, resulting in $125$ candidate joint configurations. Stride $s=0$ denotes plain compression, which encodes dispatch activations directly. We train $40$ joint configurations in dev mode and adapt the full model for $19$ of these configurations (\Cref{tab:sweep-recipes}). Dispatch-only and combine-only ablations leave the other transfer at full width. The combine ablations cover $5$ widths and $3$ decoder variants (\Cref{app:combine-correction}). All joint configurations use direct combine decoding. For all models except Qwen3-30B-A3B, we adapt only one configuration at stride $4$ with about $2\times$ RDMA reduction: $(4,768,1408)$ for Gemma 4 and Nemotron-3.5-Lightning, and $(4,512,1024)$ for Kanana-2 and Moonlight (\Cref{tab:domain-results}).

\begin{table}[t]
\centering\tablestyle[8pt]
\begin{tabular}{@{}lrr@{}}
\toprule
Compression setting & Dev mode & Adaptation \\
\midrule
Dispatch-only & $23$ & -- \\
Combine-only & $15$ & -- \\
Joint & $40$ & $19$ \\
\bottomrule
\end{tabular}
\caption{\textbf{Number of trained configurations for Qwen3-30B-A3B.} The $19$ adapted configurations are a subset of the $40$ joint configurations trained in dev mode. Dashes indicate settings not trained.}
\label{tab:sweep-recipes}
\end{table}

\subsection{Initialization and Adaptation}

\textbf{Initialization.} We fit the compression parameters on $512$K token positions from $64$K Dolci-Instruct documents (\Cref{subsec:echo-adapt}). We evaluate initialization KL on held-out Dolci-Instruct documents using the training distillation loss. Weighting each reconstruction error by the response of the experts' gate and up projections lowers this KL by $12$--$18\%$ compared with plain reconstruction error, at every stride and for plain dispatch.

\textbf{Training settings.} \Cref{tab:adaptation-settings} lists the hyperparameters and training data budgets for dev mode and full-model adaptation. The larger training budget for full-model adaptation makes it difficult to isolate the effect of updating the backbone.

\begin{table}[htbp]
\centering\tablestyle[5pt]
\begin{tabular}{@{}lcc@{}}
\toprule
Setting & Dev mode & Adaptation \\
\midrule
Trainable parameters & Compression parameters & All model parameters \\
Folding & After training & Before training \\
Training steps & $1000$ & $4152$ \\
Warmup steps & $50$ & $208$ \\
Training tokens & $1.04$B & $4.32$B \\
\quad Dolmino & $481$M & $2.00$B \\
\quad Dolci-Instruct & $560$M & $2.32$B \\
Initial Dolci-Instruct share & $46.50\%$ & $46.45\%$ \\
\midrule
\multicolumn{3}{@{}l}{\emph{Learning rate (peak $\to$ final)}} \\
Dense backbone & -- & $2\times10^{-5} \to 2\times10^{-6}$ \\
Router & -- & $1\times10^{-5} \to 1\times10^{-6}$ \\
Experts and folded factors & -- & $3\times10^{-5} \to 3\times10^{-6}$ \\
\bottomrule
\end{tabular}
\caption{\textbf{Training settings for dev mode and adaptation.} Token counts are per configuration and use the Qwen3 tokenizer. Dolci-Instruct share is the fraction of packed sequences drawn from that dataset. Dashes indicate frozen parameter groups.}
\label{tab:adaptation-settings}
\end{table}

\textbf{Training data and objective.} We mix Dolmino with Dolci-Instruct \citep{olmo2025olmo}. The Dolci-Instruct share is constant for the first $2/3$ of training and then increases linearly to $90\%$. To avoid keeping a separate copy of the model in memory as a teacher during training, we precompute and cache its top-$1024$ logits and the remaining probability mass. Distillation compares teacher and student probabilities for these tokens and a tail bucket containing the remaining mass. The objective combines this KL loss with cross-entropy on Dolci-Instruct responses. \Cref{tab:adaptation-optimization} lists the hyperparameters common to dev mode and full-model adaptation.

\begin{table}[htbp]
\centering\tablestyle[8pt]
\begin{tabular}{@{}ll@{}}
\toprule
Setting & Value \\
\midrule
Global batch size & $128$ packed sequences \\
Sequence length & $8192$ tokens \\
Distillation temperature & $1$ \\
KL loss weight & $1.0$ \\
Cross-entropy loss weight & $0.01$ \\
\midrule
Optimizer & Adam \\
Adam $(\beta_1,\beta_2)$ & $(0.9,0.999)$ \\
Adam $\epsilon$ & $10^{-8}$ \\
Learning rate (compression) & $5\times10^{-5} \to 5\times10^{-6}$ \\
Learning-rate schedule & Linear warmup and cosine decay \\
Warmup fraction & $5\%$ \\
Weight decay & $0$ \\
Gradient clipping & $1.0$ \\
\midrule
Precision & BF16 \\
Random seed & $42$ \\
\bottomrule
\end{tabular}
\caption{\textbf{Hyperparameters common to dev mode and full-model adaptation.} The compression learning rate (peak $\to$ final) applies to all compression parameters in both the dispatch and combine paths.}
\label{tab:adaptation-optimization}
\end{table}

\subsection{Adaptation Cost}
\label{app:adaptation-cost}

\textbf{One full adaptation costs about $252$ H100 GPU-hours.} \Cref{tab:adaptation-cost} reports the measured cost of each stage for Qwen3-30B-A3B, averaged over the $19$ full-model adaptation and $40$ dev mode runs, each on $64$ H100 GPUs. Initialization statistics take about $0.1$ GPU-hours, and the teacher targets are cached once and reused by every configuration. Dev mode runs serve the configuration search, which is a one-time study. Its conclusion transfers to other models without repeating the search: we adapt the other four models directly at stride $4$ with $r_d\approx d/4$ and $r_c\approx d/2$ (\Cref{tab:domain-results}).

\textbf{Adaptation pays for itself within days of serving.} At the measured BF16 speedups, one EP32 deployment over $4$ nodes at full load recovers an average adaptation run in $1.1$ days for $(8,256,512)$ and $3.2$ days for $(4,512,1280)$ (\Cref{tab:break-even}).

\textbf{Nearly the whole token budget goes to end-to-end training.} Initialization is a closed-form fit on $512$K token positions. End-to-end training matters more as compression grows: moving from dev mode on $1.04$B tokens to full-model adaptation on $4.32$B tokens adds $1.5$ pp at $(4,512,1280)$, $4.7$ pp at $(4,512,768)$ and $10.4$ pp at $(8,256,512)$ (\Cref{tab:alladapted}).

\begin{table}[htbp]
\centering\tablestyle[8pt]
\begin{tabular}{@{}lrr@{}}
\toprule
Stage & Tokens & H100 GPU-hours per run \\
\midrule
Full-model adaptation & $4.32$B & $252$ \\
Dev mode (configuration search) & $1.04$B & $67$ \\
Initialization statistics & $512$K positions & $0.1$ \\
\bottomrule
\end{tabular}
\caption{\textbf{Adaptation cost for Qwen3-30B-A3B.} Averages over $19$ full-model adaptation and $40$ dev mode runs, from recorded runtimes and GPU counts. Caching the teacher targets and capturing the initialization activations are one-time stages shared by every configuration and are not included.}
\label{tab:adaptation-cost}
\end{table}

\begin{table}[htbp]
\centering\tablestyle[6pt]
\begin{tabular}{@{}cccrrr@{}}
\toprule
\multicolumn{3}{c}{Configuration} & & & \\
\cmidrule(lr){1-3}
\makecell[b]{Stride\\$s$} & \makecell[b]{Dispatch\\$r_d$} & \makecell[b]{Combine\\$r_c$} & \makecell[b]{Throughput\\speedup} & \makecell[b]{GPU-hours saved\\per hour} & Break-even \\
\midrule
4 & 512 & 1280 & $1.11\times$ & $3.3$ & $77$ h ($3.2$ days) \\
4 & 512 & 1024 & $1.17\times$ & $4.6$ & $55$ h ($2.3$ days) \\
4 & 512 & 768 & $1.23\times$ & $5.9$ & $43$ h ($1.8$ days) \\
8 & 256 & 512 & $1.43\times$ & $9.6$ & $26$ h ($1.1$ days) \\
\bottomrule
\end{tabular}
\caption{\textbf{Serving time that repays an adaptation run.} One EP32 deployment over $4$ nodes ($32$ H100 GPUs) serving $4096$ input and $256$ output tokens at $128$ requests per GPU with BF16 communication. GPU-hours saved per hour follow from the measured throughput speedup, and break-even is the serving time that repays an average adaptation run of $252$ GPU-hours.}
\label{tab:break-even}
\end{table}

\subsection{Task Evaluation}
\label{app:tasks}

\textbf{Evaluation setup.} Few-shot tasks, zero-shot BBH and IFEval use the built-in tasks of \texttt{lm-eval} 0.4.12 with their default prompts and answer extraction. The other zero-shot chat tasks run in the same harness with our chain-of-thought prompts, extracting the option letter for multiple-choice tasks and the final answer for mathematics. Code and IFBench use their official harnesses: EvalPlus 0.3.1 for HumanEval+ and MBPP+, LiveCodeBench release v6 and IFBench 0.2.0. Every model and configuration uses the same suite, settings and seed, with vLLM 0.29.0 \citep{kwon2023efficientmemorymanagementlarge} as the inference backend. All evaluations run on $8$ GPUs with EP8 and DP8 and no tensor parallelism, a context limit of $32768$ tokens and seed $42$. Each configuration is trained and evaluated with a single seed.

\textbf{Uncompressed references.} The uncompressed reference for each model is our checkpoint fine-tuned on Dolci-Instruct (\Cref{subsec:exp-setup}), evaluated with thinking disabled and the same prompts as every EchoMoE configuration. Absolute scores therefore differ from those reported for the official releases, and we compare each EchoMoE model only with its own uncompressed reference.

\textbf{Evaluation suite.} We evaluate all trained configurations on $18$ benchmarks using zero-shot chat and few-shot non-chat prompts, with $9$ benchmarks evaluated in both formats. \Cref{tab:tasks} lists all benchmarks grouped by domain including their prompt formats.

\textbf{Development suite.} We use zero-shot chat-CoT evaluations of MMLU, HellaSwag and GSM8K for development. Their mean score for dispatch-only and combine-only configurations guides joint configuration selection (\Cref{app:search}).

\begin{table}[htbp]
\centering\tablestyle[6pt]
\begin{tabular}{@{}lcc@{}}
\toprule
Benchmark & Non-chat prompt & Chat prompt \\
\midrule
\multicolumn{3}{@{}l}{\emph{Knowledge and reasoning}} \\
MMLU \citep{hendryckstest2021} & 5-shot & 0-shot, CoT \\
MMLU-Pro \citep{wang2024mmlupro} & 5-shot, CoT & 0-shot, CoT \\
MMLU-Redux \citep{gema2024mmluredux} & 5-shot & 0-shot, CoT \\
ARC-Easy \citep{clark2018arc} & 5-shot & 0-shot, CoT \\
ARC-Challenge \citep{clark2018arc} & 5-shot & 0-shot, CoT \\
HellaSwag \citep{zellers2019hellaswag} & 5-shot & 0-shot, CoT \\
BBH \citep{suzgun2022bbh} & 3-shot, CoT & 0-shot, CoT \\
GPQA-Diamond \citep{rein2023gpqa} & -- & 0-shot, CoT \\
\midrule
\multicolumn{3}{@{}l}{\emph{Mathematics}} \\
GSM8K \citep{cobbe2021gsm8k} & 5-shot, CoT & 0-shot, CoT \\
MATH-500 \citep{lightman2023lets} & 4-shot, CoT & 0-shot, CoT \\
AIME 2024 \citep{maa2024aime} & -- & 0-shot, CoT \\
AIME 2025 \citep{maa2025aime} & -- & 0-shot, CoT \\
HMMT Feb.\ 2025 \citep{hmmt2025february} & -- & 0-shot, CoT \\
\midrule
\multicolumn{3}{@{}l}{\emph{Coding}} \\
HumanEval+ \citep{liu2023evalplus} & -- & 0-shot \\
MBPP+ \citep{liu2023evalplus} & -- & 0-shot \\
LiveCodeBench \citep{jain2024livecodebench} & -- & 0-shot \\
\midrule
\multicolumn{3}{@{}l}{\emph{Instruction following}} \\
IFEval \citep{zhou2023ifeval} & -- & 0-shot \\
IFBench \citep{pyatkin2025ifbench} & -- & 0-shot \\
\bottomrule
\end{tabular}
\caption{\textbf{Benchmarks and prompt formats used for evaluation grouped by domain.} The non-chat and chat columns specify the number of in-context examples and whether chain-of-thought (CoT) prompting is used.}
\label{tab:tasks}
\end{table}

\textbf{Chat evaluation.} We use the each model's chat template with thinking mode disabled wherever supported and decode with temperature $0.7$, top-$p=0.8$, top-$k=20$ and presence penalty $1.5$. We generate $1$ sample per question with a maximum of $8192$ new tokens. For AIME and HMMT, we increase the sample count to $64$ and the token limit to $24576$. For GPQA-Diamond, we increase the sample count to $10$. For LiveCodeBench, we generate $10$ samples per problem, lower the temperature to $0.2$, set top-$p=0.95$, disable top-$k$ filtering and increase the token limit to $16384$. For IFEval and IFBench, we use greedy decoding with token limits of $1280$ and $2048$, respectively.

\textbf{Few-shot evaluation.} We use prompts with in-context examples and score answer-option log-likelihoods for MMLU, ARC-Easy, ARC-Challenge and HellaSwag. For the remaining few-shot benchmarks, we use greedy decoding with a maximum of $2048$ new tokens for MMLU-Pro, GSM8K and MATH-500, $1024$ for BBH and $32$ for MMLU-Redux. We select in-context examples using seed $42$.

\textbf{Scoring.} For tasks with multiple samples per question, we score each question by the fraction of correct samples and report the average across questions. We verify boxed answers for AIME and HMMT using \texttt{math\_verify}. We report pass@$1$ for code benchmarks, with $164$ HumanEval+ and $378$ MBPP+ problems. We use strict prompt-level accuracy for IFEval and accuracy for the remaining benchmarks.

\textbf{Uncompressed reference.} We report score changes for Qwen3-30B-A3B relative to the uncompressed model's score of $66.92\%$ with the standard vLLM inference engine, and for the other models relative to their own uncompressed variants (\Cref{tab:domain-results}). For the BF16--FP8 comparison, we evaluate all models with our vLLM integration, which supports a custom DeepEP-based FP8 combine kernel. The uncompressed model scores $66.63\%$ in BF16 (\Cref{tab:fp8}).

\section{Adaptation and Baselines Comparisons}
\label{app:performance}

\subsection{Dev Mode and Full Model Adaptation}
\label{app:recovery}

\begin{figure}[htbp]
\centering
\includegraphics[width=\linewidth]{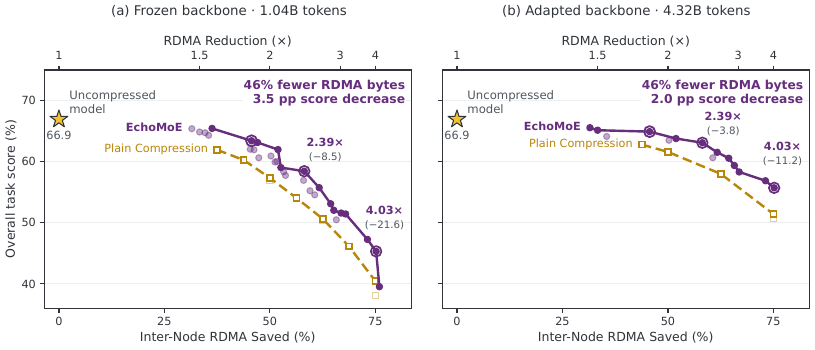}

\caption{\textbf{Dev mode reveals the benefits of compression, while full model adaptation further improves the quality--communication frontier.} EchoMoE outperforms plain compression in both \textbf{(a)} dev mode and \textbf{(b)} full-model adaptation. Lines show the empirical Pareto frontiers over $40$ dev mode and $19$ adapted joint configurations, with selected configurations from \Cref{fig:pareto} highlighted. BF16 RDMA savings are calculated for a $4$-node EP$32$ group, including reference transfers.}
\label{fig:pareto-frozen}
\end{figure}

\textbf{Dev mode predicts the benefits of full-model adaptation, which further improves the quality--communication trade-off.} Dev mode already demonstrates EchoMoE's advantage over plain compression (\Cref{fig:pareto-frozen}), providing a useful signal for the performance achievable after adaptation. Full-model adaptation further improves task performance in $18$ of the $19$ joint configurations, with a gain of $10.4$ pp at $4.03\times$ RDMA reduction. For example, adapted $(4,512,768)$ achieves $63.08\%$ overall task performance at $2.39\times$ RDMA reduction, comparable to dev mode $(4,768,1024)$ at $63.12\%$ and $1.89\times$, while sending $21\%$ fewer RDMA bytes. Thus, dev mode guides configuration selection, while full-model adaptation unlocks better quality--communication trade-offs. Complete results are reported in \Cref{tab:allgroups,tab:alladapted}.

\begin{figure}[H]
\centering
\includegraphics[width=0.73\linewidth]{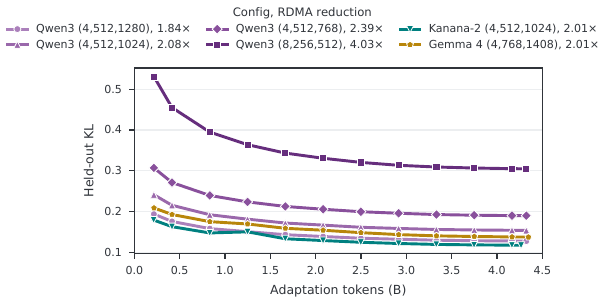}
\caption{\textbf{Teacher--student KL keeps falling throughout adaptation.} Held-out KL for four Qwen3-30B-A3B configurations and for Kanana-2 and Gemma 4 at about $2\times$ RDMA reduction, averaged over four general held-out sets (Dolci-Instruct, Dolmino, HellaSwag and MMLU), against adaptation tokens. Each model's KL is measured against its own uncompressed model.}
\label{fig:training-curves}
\end{figure}

\textbf{Additional adaptation may further narrow the gap to the uncompressed model.} We adapt Qwen3-30B-A3B, Kanana-2, and Gemma 4 with similar budgets of approximately $4.3$B tokens ($4.27$--$4.35$B). For Kanana-2 and Gemma 4, task performance improves during adaptation, narrowing the gap to their uncompressed counterparts by $2.2$ pp from update $1000$ to $3600$. Improvements taper off over the final $13\%$ of updates, as the learning rate decays to $10$--$14\%$ of its peak. Nevertheless, held-out KL to the uncompressed teacher continues to decrease by $0.6$--$1.0\%$ across all three models during this period (\Cref{fig:training-curves}). These trends suggest that longer adaptation schedules, potentially with additional training data, could further close the performance gap.

\subsection{Comparison with Plain Compression}
\label{app:matched}

\textbf{EchoMoE improves task performance within the same RDMA budget.} EchoMoE outperforms plain compression in all $10$ dev mode and $5$ adapted comparisons, with improvements of up to $7.2$ and $5.0$ pp, respectively as shown in \Cref{tab:matchedall}. Each comparison allows the EchoMoE stride and both code widths to vary while restricting RDMA bytes to the plain-compression budget. EchoMoE also preserves routing better than plain compression at matched bytes (\Cref{app:routing}).

\begin{table}[t]
\centering\tablestyle[3pt]
\begin{tabular}{@{}ccrrcccrrr@{}}
\toprule
\multicolumn{4}{c}{Plain compression} & \multicolumn{6}{c}{EchoMoE} \\
\cmidrule(lr){1-4}\cmidrule(lr){5-10}
\multicolumn{2}{c}{Configuration} & & & \multicolumn{3}{c}{Configuration} & & & \\
\cmidrule(lr){1-2}\cmidrule(lr){5-7}
\makecell[b]{Dispatch\\$r_d$} & \makecell[b]{Combine\\$r_c$} & \makecell[b]{RDMA\\Reduction} & \makecell[b]{Overall\\(\%)} & \makecell[b]{Stride\\$s$} & \makecell[b]{Dispatch\\$r_d$} & \makecell[b]{Combine\\$r_c$} & \makecell[b]{RDMA\\Reduction} & \makecell[b]{Overall\\(\%)} & \makecell[b]{$\Delta$\\(pp)} \\
\midrule
\multicolumn{10}{@{}l}{\emph{Dev mode}} \\
1280 & 1280 & $1.60\times$ & 61.9 & 4 & 512 & 1280 & $1.84\times$ & 63.4 & $+1.5$ \\
1280 & 1024 & $1.78\times$ & 60.2 & 4 & 512 & 1280 & $1.84\times$ & 63.4 & $+3.2$ \\
1024 & 1024 & $2.00\times$ & 56.9 & 4 & 512 & 1024 & $2.08\times$ & 62.0 & $+5.0$ \\
1280 & 768 & $2.00\times$ & 57.3 & 4 & 512 & 1024 & $2.08\times$ & 62.0 & $+4.6$ \\
1024 & 768 & $2.29\times$ & 54.0 & 4 & 512 & 768 & $2.39\times$ & 58.4 & $+4.4$ \\
1024 & 512 & $2.67\times$ & 50.6 & 4 & 512 & 512 & $2.81\times$ & 53.1 & $+2.5$ \\
768 & 768 & $2.67\times$ & 50.3 & 4 & 512 & 512 & $2.81\times$ & 53.1 & $+2.8$ \\
768 & 512 & $3.20\times$ & 46.1 & 6 & 256 & 512 & $3.72\times$ & 47.3 & $+1.1$ \\
512 & 512 & $4.00\times$ & 40.4 & 8 & 256 & 512 & $4.03\times$ & 45.3 & $+4.9$ \\
768 & 256 & $4.00\times$ & 38.1 & 8 & 256 & 512 & $4.03\times$ & 45.3 & $+7.2$ \\
\midrule
\multicolumn{10}{@{}l}{\emph{Full model adaptation}} \\
1280 & 1024 & $1.78\times$ & 62.8 & 4 & 512 & 1280 & $1.84\times$ & 64.9 & $+2.1$ \\
1024 & 1024 & $2.00\times$ & 61.6 & 4 & 512 & 1024 & $2.08\times$ & 63.8 & $+2.2$ \\
768 & 768 & $2.67\times$ & 58.0 & 4 & 512 & 512 & $2.81\times$ & 60.6 & $+2.6$ \\
512 & 512 & $4.00\times$ & 51.4 & 8 & 256 & 512 & $4.03\times$ & 55.7 & $+4.3$ \\
768 & 256 & $4.00\times$ & 50.7 & 8 & 256 & 512 & $4.03\times$ & 55.7 & $+5.0$ \\
\bottomrule
\end{tabular}
\caption{\textbf{EchoMoE consistently outperforms plain compression at matched RDMA budgets.} Each plain-compression configuration is paired with the highest-scoring tested EchoMoE configuration that sends no more BF16 RDMA bytes at EP$=32$ over $4$ nodes, including reference transfers. Overall task performance is reported in percent, with $\Delta$ denoting the EchoMoE improvement in pp.}
\label{tab:matchedall}
\end{table}

\textbf{Plain compression reduces local computation but incurs larger quality losses.} Because its experts operate directly on compressed codes without reference reconstruction, plain compression requires $35$--$65\%$ fewer matrix MACs at matched RDMA budgets, with no reference buffer or collective (\Cref{tab:matched-resources}). However, after adaptation, it loses an additional $2.1$--$5.1$ pp relative to EchoMoE, from $5.3$ versus $3.1$ pp at about $2\times$ RDMA reduction to $15.5$ versus $11.2$ pp at about $4\times$. Conversely, at comparable task performance, EchoMoE sends $26\%$ fewer RDMA bytes (\Cref{fig:pareto}). For example, EchoMoE $(4,512,768)$ at $2.39\times$ RDMA reduction outperforms even the least-compressed adapted plain configuration, $(1280,1024)$ at $1.78\times$. Given this consistent quality advantage, we evaluate serving performance against quality-preserving deployment alternatives: the uncompressed model and FP8 communication (\Cref{app:serving}).

\begin{table}[t]
\centering\tablestyle[5pt]
\begin{tabular}{@{}lrrr@{}}
\toprule
Configuration & \makecell[b]{RDMA\\Reduction} & \makecell[b]{Overall\\(\%)} & \makecell[b]{Matrix\\MACs} \\
\midrule
\emph{Qwen3-30B-A3B ($d=20248$)} & $1.00\times$ & 66.9 & $1.00\times$ \\
\midrule
\multicolumn{4}{@{}l}{\emph{About $1.8\times$}} \\
EchoMoE $(4,512,1280)$ & $1.84\times$ & 64.9\,\scorechange{$-$2.0} & $1.09\times$ \\
Plain $(1280,1024)$ & $1.78\times$ & 62.8\,\scorechange{$-$4.1} & $0.71\times$ \\
\midrule
\multicolumn{4}{@{}l}{\emph{About $2\times$}} \\
EchoMoE $(4,512,1024)$ & $2.08\times$ & 63.8\,\scorechange{$-$3.1} & $1.03\times$ \\
Plain $(1024,1024)$ & $2.00\times$ & 61.6\,\scorechange{$-$5.3} & $0.61\times$ \\
\midrule
\multicolumn{4}{@{}l}{\emph{About $2.5\times$}} \\
EchoMoE $(4,512,768)$ & $2.39\times$ & 63.1\,\scorechange{$-$3.8} & $0.98\times$ \\
Plain $(768,768)$ & $2.67\times$ & 58.0\,\scorechange{$-$8.9} & $0.46\times$ \\
\midrule
\multicolumn{4}{@{}l}{\emph{About $4\times$}} \\
EchoMoE $(8,256,512)$ & $4.03\times$ & 55.7\,\scorechange{$-$11.2} & $0.86\times$ \\
Plain $(512,512)$ & $4.00\times$ & 51.4\,\scorechange{$-$15.5} & $0.31\times$ \\
\bottomrule
\end{tabular}
\caption{\textbf{Quality and computation trade-offs at matched RDMA budgets.} Adapted Qwen3-30B-A3B configurations. Matrix MACs count expert and owner projections per token and MoE layer, normalized to the uncompressed model (\Cref{eq:local-macs}). Plain compression uses the same accounting without the reference.}
\label{tab:matched-resources}
\end{table}

\subsection{Routing Agreement with the Uncompressed Model}
\label{app:routing}

\textbf{Routing stays close to the uncompressed model's.} The router runs on the owner rank and reads the full $x_\ell$ (\Cref{fig:arch}), so compression changes routing only through errors that propagate to later layers. After adaptation, $(4,512,1280)$ selects $87.6$--$87.8\%$ of the uncompressed model's top-$8$ experts with a router KL of $0.020$--$0.022$, and $(8,256,512)$ selects $79.8$--$80.5\%$ (\Cref{tab:routing}). Adaptation raises agreement for every configuration.

\textbf{The reference preserves routing better at matched bytes.} At about $2\times$, EchoMoE keeps $86.4$--$86.6\%$ of the uncompressed model's experts against $84.7$--$84.8\%$ for plain compression. At about $4\times$, it keeps $79.8$--$80.5\%$ against $77.7$--$78.2\%$, following the same order as task performance.

\begin{table}[t]
\centering\tablestyle[4pt]
\begin{tabular}{@{}lrcccccc@{}}
\toprule
& & \multicolumn{4}{c}{Experts shared (\%)} & \multicolumn{2}{c}{Router KL} \\
\cmidrule(lr){3-6}\cmidrule(lr){7-8}
& & \multicolumn{2}{c}{Initialization} & \multicolumn{2}{c}{After adaptation} & \multicolumn{2}{c}{After adaptation} \\
\cmidrule(lr){3-4}\cmidrule(lr){5-6}\cmidrule(lr){7-8}
Configuration & RDMA reduction & Dolci & Dolmino & Dolci & Dolmino & Dolci & Dolmino \\
\midrule
EchoMoE $(4,512,1280)$ & $1.84\times$ & $84.8$ & $83.2$ & $87.6$ & $87.8$ & $0.020$ & $0.022$ \\
EchoMoE $(4,512,1024)$ & $2.08\times$ & $83.0$ & $81.1$ & $86.4$ & $86.6$ & $0.024$ & $0.028$ \\
Plain $(1024,1024)$ & $2.00\times$ & $80.2$ & $76.1$ & $84.8$ & $84.7$ & $0.033$ & $0.040$ \\
EchoMoE $(4,512,768)$ & $2.39\times$ & $80.3$ & $77.9$ & $84.7$ & $85.0$ & $0.032$ & $0.037$ \\
EchoMoE $(8,256,512)$ & $4.03\times$ & $69.4$ & $65.1$ & $79.8$ & $80.5$ & $0.066$ & $0.072$ \\
Plain $(512,512)$ & $4.00\times$ & $64.2$ & $57.0$ & $77.7$ & $78.2$ & $0.088$ & $0.100$ \\
\bottomrule
\end{tabular}
\caption{\textbf{Routing agreement with the uncompressed model.} Qwen3-30B-A3B on held-out Dolci-Instruct and Dolmino data, one set per source of the adaptation mixture. Experts shared is the fraction of the uncompressed model's top-$8$ experts that the compressed model also selects, averaged over positions and all $48$ MoE layers. Router KL is measured from the uncompressed model's routing distribution to the compressed model's.}
\label{tab:routing}
\end{table}

\subsection{Comparison with LSH-MoE}
\label{app:lsh}

LSH-MoE \citep{nie2024lshmoecommunicationefficientmoetraining} hashes tokens sent to each expert, communicates bucket centroids, and adds token--centroid residuals to expert outputs. EchoMoE instead predicts each token's expert input from its own earlier-layer reference.

\textbf{Cross-layer redundancy is stronger than cross-token redundancy.}
Using the per-dimension affine predictor from \Cref{tab:moe-input-similarity}, a token's input four layers earlier recovers $55.7\%$, $70.0\%$, and $62.2\%$ of its MoE input on Qwen3-30B-A3B, Qwen3-235B-A22B, and DeepSeek-V3, respectively. The best other token sent to the same expert recovers only $51.3\%$, $42.7\%$, and $36.2\%$, while LSH-MoE's actual representatives recover just $8.0\%$, $5.9\%$, and $5.8\%$ (\Cref{tab:lsh-predict}). Larger token pools do not close the gap: on Qwen3-30B-A3B, best-partner predictability is $16.7\%$, $36.8\%$, $51.2\%$, and $51.3\%$ for pools of $100$, $1$K, $10$K, and all captured tokens. Three in four best partners come from the same document, suggesting that cross-token redundancy requires many tokens from the same sequence in flight, whereas decoding processes only one token per sequence per step.

\textbf{EchoMoE preserves more task performance after adaptation.}
LSH-MoE's residual compensation outputs $E(c)+(x_i-c)$ for each cluster member. On Qwen3-30B-A3B, the residual $x_i-c$ has $6$--$8\times$ the norm of the expert output $E(c)$ at the first two MoE layers ($3.3$ versus $0.52$ and $9.8$ versus $1.18$), and including compensation yields an initial KL to the uncompressed model of $16.4$ nats. \citet{nie2024lshmoecommunicationefficientmoetraining} fine-tune pretrained GPT-MoE and Swin-MoE checkpoints but do not specify whether compensation is used. We therefore keep the compensation but learn its scale, $E(c)+\alpha(x_i-c)$, where each layer and expert has its own $\alpha$, initialized at $0$ and learned during adaptation. After adaptation, the per-layer means of $\alpha$ lie between $-0.02$ and $0.08$, far below the $\alpha=1$ of the published method. Adapted with our $4.32$B-token recipe, LSH-MoE reaches $57.1\%$ overall performance, versus $63.8\%$ for EchoMoE $(4,512,1024)$, despite sending $31\%$ more RDMA bytes during training. LSH-MoE underperforms EchoMoE on $26$ of $27$ tasks (\Cref{tab:lsh-trained}), with final held-out KL of $0.389$ and $0.264$ on Dolmino and Dolci-Instruct, versus $0.126$ and $0.067$ for EchoMoE. During evaluation, decode batches contain too few tokens per bucket for effective merging, and LSH-MoE sends one centroid per remote expert rather than one copy per remote node as in DeepEP. Consequently, it sends about twice DeepEP's bytes in decode, whereas EchoMoE's reduction is batch-independent.

\begin{table}[t]
\centering\tablestyle[3pt]
\begin{tabular}{@{}lrrrrr@{\hspace{10pt}}l@{}}
\toprule
& & \multicolumn{5}{c}{Task Performance (\%) $\uparrow$} \\
\cmidrule(lr){3-7}
Method & \makecell[b]{RDMA\\Reduction} & \makecell[b]{Knowledge\\\&\ Reasoning} & Math & Code & \makecell[b]{Instruction\\Following} & \makecell[b]{Overall\\\textcolor{tabchange}{($\Delta$)}} \\
\midrule
Uncompressed & $1.00\times$ & 77.6 & 50.7 & 59.2 & 55.5 & 66.9 \\
EchoMoE $(4,512,1024)$ & $2.08\times$ & 74.3 & 47.8 & 55.2 & 53.4 & 63.8\,\scorechange{$-$3.1} \\
LSH-MoE & $1.59\times$ & 65.3 & 44.2 & 53.1 & 46.7 & 57.1\,\scorechange{$-$9.8} \\
\bottomrule
\end{tabular}
\caption{\textbf{EchoMoE preserves more task performance than LSH-MoE after adaptation.} Qwen3-30B-A3B, both adapted with $4.32$B tokens and evaluated on the full task suite. LSH-MoE uses cross-polytope hashing with $5$ projection dimensions and $3$ hashes per source rank and expert, calibrated on LSH-MoE students' own states, and scales the residual $x_i-c$ by one learned $\alpha$ per layer and expert, trained from $0$. Its RDMA reduction is averaged over training; at evaluation, it sends $1.19\times$ DeepEP's bytes overall and about $2\times$ in decode. After out-of-memory errors, $11$ of its $27$ tasks were served with $0.85$ rather than $0.94$ of GPU memory.}
\label{tab:lsh-trained}
\end{table}

\begin{table}[t]
\centering\tablestyle[5pt]
\begin{tabular}{@{}lrrr@{}}
\toprule
Predicted from & Qwen3-30B-A3B & Qwen3-235B-A22B & DeepSeek-V3 \\
\midrule
Same token, $4$ layers earlier (EchoMoE) & $55.7\%$ & $70.0\%$ & $62.2\%$ \\
Best other token, same expert & $51.3\%$ & $42.7\%$ & $36.2\%$ \\
LSH-MoE representative, six hashes & $8.0\%$ & $5.9\%$ & $5.8\%$ \\
\bottomrule
\end{tabular}
\caption{\textbf{Cross-layer versus cross-token predictability.} Fraction of MoE input recovered by a per-dimension affine predictor on held-out tokens. The second row selects the closest token among all tokens sent to the same expert, an oracle unavailable to hashing. The third row includes only tokens compressed by LSH-MoE.}
\label{tab:lsh-predict}
\end{table}

\section{Compression Ablations}
\label{app:ablations}

All ablations reported in this section use Qwen3-30B-A3B trained in dev mode. Dispatch ablations keep combine at full width and combine ablations keep dispatch at full width.

\subsection{Reference Stride and Dispatch Width}
\label{app:reference-budget}

\textbf{Smaller dispatch widths benefit more from frequent reference updates.} Increasing the reference stride from $4$ to $12$ reduces task performance by $1.6$ pp at width $1280$, compared with $15.2$ pp at width $256$ (\Cref{tab:dispatch-grid}). Reference dispatch consistently outperforms plain dispatch at all $3$ compression widths. Across all $23$ configurations, we keep combine at full width and follow the dev mode training recipe.

\begin{table}[t]
\centering\tablestyle[5pt]
\begin{tabular}{@{}crrrrr@{}}
\toprule
 & \multicolumn{5}{c}{Overall drop (pp)} \\
\cmidrule(lr){2-6}
\multirow{2}{*}{\makecell{Dispatch\\$r_d$}} & \multicolumn{4}{c}{Reference stride $s$} & \multirow{2}{*}{\makecell{Plain\\dispatch}} \\
\cmidrule(lr){2-5}
 & 4 & 6 & 8 & 12 & \\
\midrule
1280 & 0.0 & 1.1 & 1.0 & 1.6 & -- \\
1024 & 0.6 & 1.8 & 2.3 & 3.0 & 7.1 \\
768  & 1.0 & 2.5 & 3.9 & 5.3 & 12.4 \\
512  & 2.6 & 5.5 & 6.3 & 10.6 & 21.1 \\
256  & 5.6 & 10.7 & 15.1 & 20.8 & -- \\
\bottomrule
\end{tabular}
\caption{\textbf{Dispatch compression across code widths and reference strides.} Overall score drops are relative to the uncompressed model ($66.92\%$), in percentage points. Reference dispatch is compared with plain dispatch at matched code widths. All configurations follow the dev mode training recipe with combine at full width.}
\label{tab:dispatch-grid}
\end{table}

\textbf{More frequent reference updates compensate for narrower dispatch codes.} At fixed combine width, $(s,r_d)=(4,512)$ outperforms $(8,768)$ by up to $2.0$ pp across all $3$ comparisons while sending slightly fewer RDMA bytes (see \Cref{tab:refresh-budget}). These results suggest that, at similar RDMA budgets, updating the reference more frequently can preserve task performance better than increasing dispatch width.

\begin{table}[b]
\centering\tablestyle[5pt]
\begin{tabular}{@{}cccrrr@{}}
\toprule
\multicolumn{3}{c}{Configuration} & & & \\
\cmidrule(lr){1-3}
\makecell[b]{Stride\\$s$} & \makecell[b]{Dispatch\\$r_d$} & \makecell[b]{Combine\\$r_c$} & \makecell[b]{RDMA\\Reduction} & \makecell[b]{Overall\\(\%)} & \makecell[b]{$\Delta$\\(pp)} \\
\midrule
8 & 768 & 1280 & $1.83\times$ & 62.1 & -- \\
4 & 512 & 1280 & $1.84\times$ & 63.4 & $+1.3$ \\
\midrule
8 & 768 & 1024 & $2.07\times$ & 60.0 & -- \\
4 & 512 & 1024 & $2.08\times$ & 62.0 & $+2.0$ \\
\midrule
8 & 768 & 768 & $2.38\times$ & 57.0 & -- \\
4 & 512 & 768 & $2.39\times$ & 58.4 & $+1.5$ \\
\bottomrule
\end{tabular}
\caption{\textbf{Reference stride and dispatch width at comparable RDMA reductions.} Both configurations in each comparison use the same combine width and follow the dev mode training recipe. RDMA reductions are calculated for BF16 dispatch, combine and reference transfers at EP$32$ group across $4$ nodes. $\Delta$ shows the improvement in overall task performance for the stride-$4$ configuration relative to its stride-$8$ counterpart, in percentage points.}
\label{tab:refresh-budget}
\end{table}

\textbf{Less frequent reference updates leave larger residuals.} At initialization, we measure the residual $\delta_\ell=x_\ell-a\odot z$. Averaged over the layers each stride compresses, its relative energy $\mathbb E\|\delta_\ell\|^2/\mathbb E\|x_\ell\|^2$ increases from $26.6\%$ at stride $4$ to $52.3\%$ at stride $12$. At each code width, the fraction of residual energy lost during compression, $\mathbb E\|\delta_\ell-D_dC_d\delta_\ell\|^2/\mathbb E\|\delta_\ell\|^2$, varies by less than $2$ pp across strides. This suggests that the larger strides increase reconstruction error mainly because the residual is larger.

\textbf{Non-uniform code widths give only a small gain.} In an offline study on the Qwen3-30B-A3B activations cached for initialization, the best per-layer allocation retains $83.7\%$ of the signal energy compared with $81.8\%$ for uniform widths. However, an allocation tuned for one set of model activations does not directly transfer to another model. We therefore continue with uniform widths given the small difference and no need for model-specific search.

\subsection{Combine Decoder Variants}
\label{app:combine-correction}

\textbf{Decoder variants.} Let $c=\sum_jc_c^{(j)}$ be the combined code, $w\in\mathbb{R}^N$ the routing weights with zeros for unselected experts and $x_\ell$ the input retained by the owner rank. Direct decoding returns $D_cc$ (\Cref{eq:echo-decombine}). The bias and gated-bias variants add trainable per-expert offsets $o_i\in\mathbb{R}^d$:
\begin{align}
\hat y_{\mathrm{bias}}&=D_cc+\sum_iw_io_i\label{eq:owner-bias}\\
\hat y_{\mathrm{gate}}&=D_cc+\sum_iw_i(1+\alpha_i)o_i\label{eq:owner-gate}\\
\alpha&=U\operatorname{SiLU}\!\left(W_x\tilde x_\ell+W_c c/\sigma_c+E_{\mathrm{exp}}^\top w\right)\nonumber
\end{align}
where $\tilde x_\ell$ is the RMS-normalized input and $\sigma_c$ is the code normalization scale.
The gate uses $W_x\in\mathbb{R}^{q\times d}$, $W_c\in\mathbb{R}^{q\times r_c}$, expert embeddings $E_{\mathrm{exp}}\in\mathbb{R}^{N\times q}$ and an output projection $U\in\mathbb{R}^{N\times q}$, producing $\alpha\in\mathbb{R}^N$, with $q=128$.

\textbf{Initialization.} We fit $\bar Y\in\mathbb{R}^{N\times d}$ so that $\bar Y^\top w$ predicts the combined expert output $y$, then initialize each offset from the reconstruction residual of its fitted vector,
\begin{equation}
\bar Y=\left(\mathbb{E}[ww^\top]+\epsilon I\right)^{-1}\mathbb{E}[wy^\top],
\qquad o_i^{(0)}=(I-D_cC_c)\bar y_i
\label{eq:owner-mean-table}
\end{equation}
where $\bar y_i^\top$ is row $i$ of $\bar Y$ and $\epsilon = 10^{-6}$. We initialize the code normalization scale as
\begin{equation}
\sigma_c^2=\operatorname{tr}(C_c\Sigma_yC_c^\top)/r_c,
\qquad \Sigma_y=\mathbb{E}[yy^\top]
\label{eq:owner-code-scale}
\end{equation}
We initialize $W_x$, $W_c$ and $E_{\mathrm{exp}}$ randomly with seed $42$ and set $U=0$, so the bias and gated-bias variants produce identical outputs at initialization.

\textbf{Bias corrections provide no consistent improvement over direct combine decoding.} Across $5$ combine widths, per-expert bias and gated-bias corrections change overall task performance by at most $0.4$ pp (see \Cref{tab:correction}). Both variants add parameters and computation without reducing RDMA bytes. We therefore use direct combine decoding in all joint configurations.

\begin{table}[htbp]
\centering\tablestyle[5pt]
\begin{tabular}{@{}crrr@{}}
\toprule
Configuration & Direct decoding & \multicolumn{2}{c}{Change from direct decoding} \\
\cmidrule(lr){1-1}\cmidrule(lr){2-2}\cmidrule(lr){3-4}
\makecell[b]{Combine\\$r_c$} & \makecell[b]{Overall\\(\%)} & \makecell[b]{Bias\\$\Delta$ (pp)} & \makecell[b]{Gated bias\\$\Delta$ (pp)} \\
\midrule
\multicolumn{1}{@{}l}{\emph{Qwen3-30B-A3B ($d=2048$)}} & 66.9 & -- & -- \\
\midrule
1280 & 65.2 & $+0.2$ & $+0.2$ \\
1024 & 64.1 & $-0.2$ & $-0.3$ \\
768 & 60.6 & $+0.2$ & $+0.4$ \\
512 & 54.7 & $+0.2$ & $+0.1$ \\
256 & 41.4 & $+0.1$ & $0.0$ \\
\bottomrule
\end{tabular}
\caption{\textbf{Combine decoder variants at fixed code width.} All configurations are trained in dev mode with dispatch at full width. Overall reports task performance with direct decoding. $\Delta$ shows the change in task performance relative to direct decoding at the same combine width, in percentage points.}
\label{tab:correction}
\end{table}

\section{Configuration Selection}
\label{app:search}

\textbf{Selection rule.} Let $Q_{\mathrm{dev}}$ denote mean task performance on the development task suite (\Cref{app:tasks}). We estimate joint performance by summing the drops from separate dispatch and combine ablations:
\begin{equation}
\widehat Q_{\mathrm{dev}}(s,r_d,r_c)
=Q_{\mathrm{dev}}^{\mathrm{d}}(s,r_d)+Q_{\mathrm{dev}}^{\mathrm{c}}(r_c)-Q_{\mathrm{dev}}^{0}
\label{eq:dev-selection-appendix}
\end{equation}
where $Q_{\mathrm{dev}}^{\mathrm{d}}$ and $Q_{\mathrm{dev}}^{\mathrm{c}}$ denote performance with only dispatch or combine compressed, respectively, and $Q_{\mathrm{dev}}^{0}$ is the uncompressed model's performance. Combine uses direct decoding and $s=0$ denotes plain dispatch.

\textbf{Development scores identify competitive joint configurations.} We evaluate selection retrospectively on $37$ of the $40$ dev mode joint configurations with corresponding dispatch and combine ablations. Their $35$ distinct RDMA reductions define the targets. For each target, we select the configuration with the highest predicted score among those that meet the reduction target, then compare its full-suite performance with the best tested configuration under the same constraint. Selected configurations are within $0.5$ pp of the best tested configurations at $33$ of $35$ targets ($94\%$), with a maximum gap of $0.70$ pp.

\textbf{The selection result holds without the development tasks.} Excluding the $3$ development evaluations from the full-suite average preserves the $33$ of $35$ targets within $0.5$ pp, with a maximum gap of $0.75$ pp. This check averages the rounded per-task results and keeps the few-shot evaluations of these benchmarks (MMLU, HellaSwag and GSM8K). These results assess selection within the trained subset, which was itself chosen using the separate ablations, and do not establish optimality over the full configuration grid.

\begin{figure}[H]
\centering
\includegraphics[width=\linewidth]{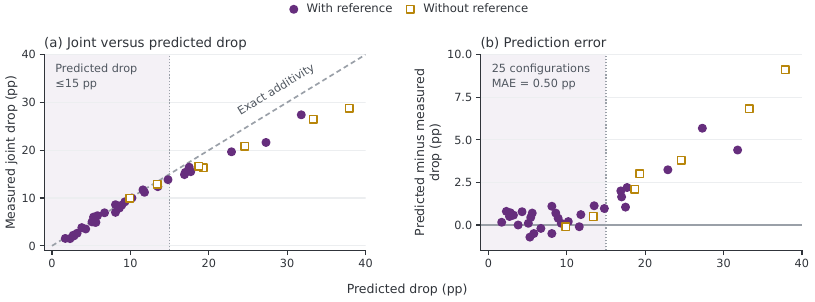}
\caption{\textbf{Dispatch and combine performance drops are approximately additive at moderate compression.} Predictions sum the separate drops on the full evaluation suite, with all configurations trained in dev mode. \textbf{(a)} Predicted and joint performance drops. \textbf{(b)} Prediction error, computed as predicted minus joint drop. Mean absolute error is $0.50$ pp for the $25$ configurations with predicted drops of at most $15$ pp.}
\label{fig:additivity}
\end{figure}

\textbf{Performance drops are approximately additive at moderate compression.} We assess additivity using dispatch, combine and joint performance drops on the full evaluation suite (\Cref{fig:additivity}). Across the $25$ configurations with predicted drops of at most $15$ pp, the mean absolute error is $0.50$ pp and the maximum is $1.13$ pp. The approximation becomes less accurate at higher compression, where the sum increasingly overestimates the joint drop.

\section{Communication and Model Costs}
\label{app:rdma-bytes}

\subsection{RDMA Reduction Across EP Topologies}

\textbf{RDMA calculation.} BF16 dispatch and combine transfer one row per token per destination node, with expert outputs combined within each node before transfer. Reference transfers are also counted per destination node. At EP32 over $4$ nodes, routed transfers reach $2.73$ remote nodes per token on average. These byte count estimates exclude protocol overhead and startup cost and depend only on the configuration, topology and routing trace.

\textbf{Reference transfers reduce savings at larger EP sizes.} Reference updates must reach more nodes as the EP group grows, while routed transfers reach only the nodes hosting selected experts. From EP16 to EP128, the RDMA reduction for $(4,512,768)$ on Qwen3-30B-A3B falls from $2.46\times$ to $1.82\times$ and from $2.12\times$ to $1.61\times$ for $(4,1792,3584)$ on DeepSeek-V3 (\Cref{tab:topology}). Combine compression does not use the reference transfers, so every configuration still reduces RDMA bytes by at least $1.23\times$ at EP128. Plain compression sends no references and keeps the same reduction at every EP degree.

\begin{table}[H]
\centering\tablestyle[5pt]
\begin{tabular}{@{}cccrrrr@{}}
\toprule
\multicolumn{3}{c}{Configuration} & \multicolumn{4}{c}{BF16 RDMA Reduction} \\
\cmidrule(lr){1-3}\cmidrule(lr){4-7}
\makecell[b]{Stride\\$s$} & \makecell[b]{Dispatch\\$r_d$} & \makecell[b]{Combine\\$r_c$} & \makecell[b]{EP16\\$2$ nodes} & \makecell[b]{EP32\\$4$ nodes} & \makecell[b]{EP64\\$8$ nodes} & \makecell[b]{EP128\\$16$ nodes} \\
\midrule
\multicolumn{7}{@{}l}{\emph{Qwen3-30B-A3B ($d=2048$, $128$ experts)}} \\
4 & 1280 & 1280 & $1.49\times$ & $1.46\times$ & $1.37\times$ & $1.23\times^{\dagger}$ \\
4 & 512 & 1280 & $1.88\times$ & $1.84\times$ & $1.69\times$ & $1.49\times^{\dagger}$ \\
4 & 512 & 1024 & $2.13\times$ & $2.08\times$ & $1.89\times$ & $1.64\times^{\dagger}$ \\
4 & 512 & 768 & $2.46\times$ & $2.39\times$ & $2.15\times$ & $1.82\times^{\dagger}$ \\
8 & 256 & 512 & $4.13\times$ & $4.03\times$ & $3.66\times$ & $3.04\times^{\dagger}$ \\
\midrule
\multicolumn{7}{@{}l}{\emph{DeepSeek-V3 ($d=7168$, $256$ experts)}} \\
4 & 1792 & 3584 & $2.12\times^{\dagger}$ & $2.06\times^{\dagger}$ & $1.87\times^{\dagger}$ & $1.61\times^{\dagger}$ \\
\bottomrule
\end{tabular}
\caption{\textbf{RDMA reduction across EP topologies.} Each configuration is compared with uncompressed BF16 transfers at the same topology, with $8$ GPUs per node. RDMA bytes depend on how many remote nodes does each token's selected experts are hosted on. Unmarked entries count these nodes from Qwen3-30B-A3B's measured routing. $^{\dagger}$ assumes each token selects its experts uniformly at random. The estimates differ by at most $0.01\times$ for Qwen3-30B-A3B from EP16 to EP64.}
\label{tab:topology}
\end{table}

At stride $4$, recorded reference destinations within each layer group average $3.0$ remote nodes at EP32 and $7.0$ remote nodes at EP64. Broadcasting references to all remote nodes, as assumed in \Cref{eq:echo-traffic}, would increase reference bytes at EP$64$ by $1.4\%$, with a smaller relative increase in total RDMA bytes.

\subsection{FP8 Communication}

\textbf{FP8 and BF16 achieve similar task performance.} Across the $19$ adapted configurations of Qwen3-30B-A3B, using FP8 for dispatch, reference and combine transfers changes overall task performance by at most $0.29$ pp (\Cref{tab:fp8}). Both evaluations use the same vLLM backend. For the uncompressed model, FP8 is applied to the full-width activations.

\begin{table}[t]
\centering\tablestyle[5pt]
\begin{tabular}{@{}cccrrr@{}}
\toprule
\multicolumn{3}{c}{Configuration} & \multicolumn{2}{c}{Overall (\%)} & \\
\cmidrule(lr){1-3}\cmidrule(lr){4-5}
\makecell[b]{Stride\\$s$} & \makecell[b]{Dispatch\\$r_d$} & \makecell[b]{Combine\\$r_c$} & BF16 & FP8 & \makecell[b]{$\Delta$\\(pp)} \\
\midrule
\multicolumn{3}{@{}l}{\emph{Qwen3-30B-A3B ($d=2048$, $128$ experts)}} & 66.63 & 66.83 & $+0.20$ \\
\midrule
\multicolumn{6}{@{}l}{\emph{EchoMoE}} \\
4 & 1280 & 1280 & 65.54 & 65.41 & $-0.13$ \\
6 & 1280 & 1280 & 65.13 & 65.17 & $+0.04$ \\
12 & 1280 & 1280 & 64.11 & 63.92 & $-0.19$ \\
4 & 512 & 1280 & 64.90 & 64.86 & $-0.04$ \\
6 & 768 & 1024 & 63.49 & 63.42 & $-0.07$ \\
4 & 512 & 1024 & 63.77 & 63.65 & $-0.13$ \\
4 & 512 & 768 & 63.08 & 62.99 & $-0.09$ \\
6 & 256 & 1024 & 60.62 & 60.71 & $+0.09$ \\
6 & 512 & 768 & 61.51 & 61.45 & $-0.06$ \\
4 & 512 & 512 & 60.55 & 60.76 & $+0.21$ \\
12 & 768 & 512 & 59.35 & 59.32 & $-0.03$ \\
6 & 256 & 768 & 58.30 & 58.48 & $+0.18$ \\
6 & 256 & 512 & 56.86 & 56.70 & $-0.16$ \\
8 & 256 & 512 & 55.70 & 55.56 & $-0.15$ \\
\midrule
\multicolumn{6}{@{}l}{\emph{Plain compression}} \\
-- & 1280 & 1024 & 62.78 & 63.03 & $+0.25$ \\
-- & 1024 & 1024 & 61.58 & 61.28 & $-0.29$ \\
-- & 768 & 768 & 57.98 & 57.91 & $-0.07$ \\
-- & 512 & 512 & 51.41 & 51.42 & $+0.01$ \\
-- & 768 & 256 & 50.73 & 50.63 & $-0.11$ \\
\bottomrule
\end{tabular}
\caption{\textbf{Task performance with BF16 and FP8 communication for Qwen3-30B-A3B.} Overall task performance of the uncompressed model and $19$ adapted configurations with BF16 or FP8 for dispatch, reference and combine transfers. $\Delta$ shows the change from BF16 to FP8 in percentage points, computed before rounding.}
\label{tab:fp8}
\end{table}

\textbf{FP8 reduces bytes per activation row.} With one FP32 scale per $128$ elements, FP8 values and scales occupy $33/64$ of the corresponding BF16 row, saving $48.4\%$. This ratio excludes routing metadata. FP8 dispatch sends one row per destination node but the custom combine kernel sends one row per destination rank and can add padding under CUDA graphs. The per-row saving therefore does not directly determine the total RDMA reduction.

\subsection{Parameter, Memory and Computation Costs}

\textbf{Folding can reduce the total parameter count.} Expert input width increases from $d$ to $d+r_d$ at non-reference layers, while output width decreases from $d$ to $r_c$ at every layer. Shared projections on the owner rank contribute additional parameters. The $(4,512,1280)$ configuration contains $30.70$B parameters, only $0.6\%$ more than the uncompressed model. Reducing the code widths brings the total down to $25.47$B for $(8,256,512)$ (\Cref{tab:local-costs}).

\begin{table}[b]
\centering\tablestyle[4pt]
\begin{tabular}{@{}cccrrrrr@{}}
\toprule
\multicolumn{3}{c}{Configuration} & & & \multicolumn{3}{c}{Relative Matrix MACs} \\
\cmidrule(lr){1-3}\cmidrule(lr){6-8}
\makecell[b]{Stride\\$s$} & \makecell[b]{Dispatch\\$r_d$} & \makecell[b]{Combine\\$r_c$} & \makecell[b]{BF16 RDMA\\Reduction} & \makecell[b]{Parameters\\(B)} & Expert & Owner & Total \\
\midrule
\multicolumn{3}{@{}l}{\emph{Qwen3-30B-A3B ($d=2048$, $128$ experts)}} & $1.00\times$ & 30.53 & $1.000$ & -- & $1.000$ \\
\midrule
4 & 512 & 1280 & $1.84\times$ & 30.70 & $1.000$ & $0.090$ & $1.090$ \\
4 & 512 & 768 & $2.39\times$ & 28.23 & $0.917$ & $0.062$ & $0.979$ \\
6 & 256 & 768 & $3.02\times$ & 26.60 & $0.861$ & $0.053$ & $0.914$ \\
8 & 256 & 512 & $4.03\times$ & 25.47 & $0.823$ & $0.040$ & $0.863$ \\
\bottomrule
\end{tabular}
\caption{\textbf{Parameter counts and matrix computation after folding.} Parameter counts cover the full model with shared owner projections. Expert and Owner report MACs relative to the uncompressed expert computation and Total reports their sum. RDMA reductions cover BF16 dispatch, combine and reference transfers at EP32 over $4$ nodes.}
\label{tab:local-costs}
\end{table}

\textbf{Reference-buffer memory grows with the EP degree.} Each rank reuses a single full-width reference buffer across layer groups. At $d=2048$, storing a BF16 reference requires $4096$ bytes per token. The buffer capacity is the engine's per-rank batched-token limit multiplied by the EP degree. With a limit of $8192$ tokens per rank, the buffer occupies $1.00$ GiB at EP32 and $2.00$ GiB at EP64 on each rank, plus additional memory for staging buffers. The smaller folded expert weights offset this memory, so the KV cache that vLLM allocates after loading changes by only $-0.8\%$ to $+2.3\%$ per GPU (\Cref{tab:fp8-memory}).

\textbf{Reducing the combine width can lower total matrix computation.} With expert intermediate width $f=768$ and $K=8$ active experts, the uncompressed model requires $3Kdf$ multiply-accumulates (MACs) per token per MoE layer for expert computation. For EchoMoE, the expert and owner matrix costs, averaged over reference and non-reference layers and normalized by this uncompressed cost, are
\begin{equation}
\eta_{\mathrm{expert}}
=\frac{2d+2r_d(1-1/s)+r_c}{3d},
\qquad
\eta_{\mathrm{owner}}
=\frac{r_d(1-1/s)+r_c}{3Kf}
\label{eq:local-macs}
\end{equation}
where $\eta_{\mathrm{expert}}$ denotes the relative cost of the folded expert input and output projections and $\eta_{\mathrm{owner}}$ denotes the dispatch encoding and combine decoding on the owner rank. At $(4,512,1280)$, the wider expert inputs and narrower expert outputs cancel, so $\eta_{\mathrm{expert}}=1$ and the owner projections add $9.0\%$ to the uncompressed expert computation (\Cref{tab:local-costs}). Reducing $r_c$ to $768$ lowers the total to $2.1\%$ below the uncompressed cost. These counts exclude attention, routing, nonlinearities, elementwise reference operations and communication. \Cref{app:serving} reports end-to-end inference performance.

\subsection{When Does Compression Speed Up Communication?}
\label{app:deepep-scaling}

\begin{figure}[htbp]
\centering
\begin{subfigure}{.5\textwidth}
  \centering
  \includegraphics[width=\linewidth]{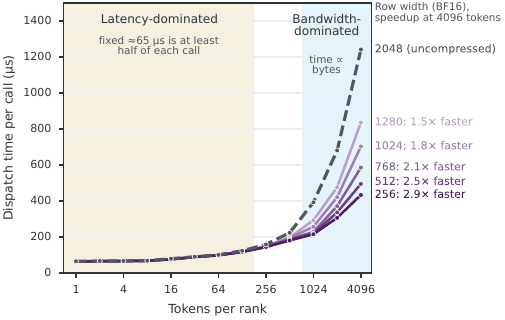}
  \caption{Dispatch, $d=2048$}
  \label{fig:deepep-scaling-dispatch}
\end{subfigure}%
\begin{subfigure}{.5\textwidth}
  \centering
  \includegraphics[width=\linewidth]{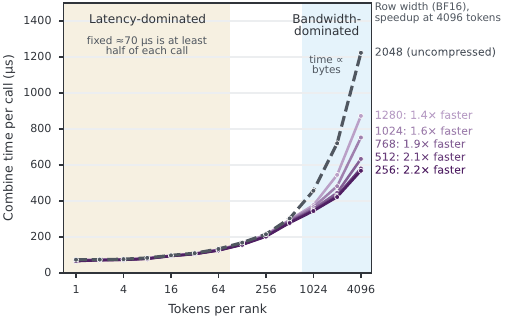}
  \caption{Combine, $d=2048$}
  \label{fig:deepep-scaling-combine}
\end{subfigure}
\par\medskip
\begin{subfigure}{.5\textwidth}
  \centering
  \includegraphics[width=\linewidth]{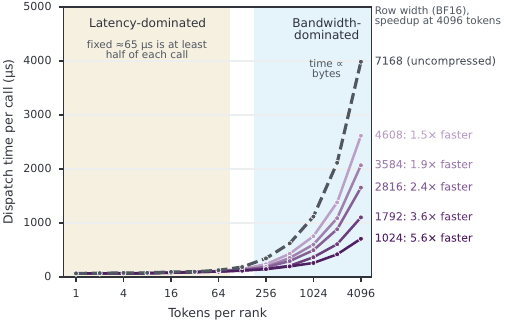}
  \caption{Dispatch, $d=7168$}
  \label{fig:deepep-scaling-dispatch-7168}
\end{subfigure}%
\begin{subfigure}{.5\textwidth}
  \centering
  \includegraphics[width=\linewidth]{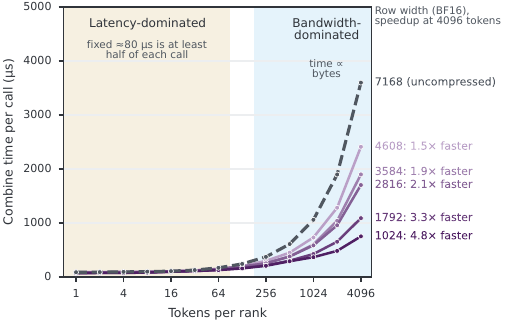}
  \caption{Combine, $d=7168$}
  \label{fig:deepep-scaling-combine-7168}
\end{subfigure}
\caption{\textbf{Compression speeds up dispatch and combine only once calls become bandwidth-bound, which happens at fewer tokens per rank for wider rows.} DeepEP V2 time per call at EP32 over $4$ nodes in BF16, with one line per width and the fastest SM count, at \textbf{(a, b)} $d=2048$ and \textbf{(c, d)} $d=7168$. At $d=2048$ and up to $512$ tokens per rank, code widths from $256$ to $1280$ speed up dispatch by at most $1.24\times$ and combine by at most $1.13\times$. At $d=7168$ and $512$ tokens per rank, a $d/4$-wide code already speeds up dispatch by $3.0\times$ and combine by $2.0\times$. Gold: the startup cost (${\approx}65~\mu$s for dispatch, ${\approx}70$--$80~\mu$s for combine) is at least half of each call. Blue: time grows with bytes. Line labels give each width's speedup over the uncompressed width at $4096$ tokens per rank.}
\label{fig:deepep-scaling}
\end{figure}

As described in \Cref{subsec:cost}, each dispatch or combine call incurs a fixed startup cost plus the time to transmit its payload. Compression reduces only the payload, so it helps only when transmission time dominates. \Cref{fig:deepep-scaling} shows where this happens across message sizes and how the boundary moves with the hidden size.

\textbf{Benchmark.} We time DeepEP V2 dispatch and combine in hybrid mode without model or expert computation on the hardware in \Cref{app:serving-setup} at EP32 over $4$ nodes with BF16 activations. Routing is uniformly at random top-$8$ over $128$ experts, as in Qwen3-30B-A3B. We sweep tokens per rank from $1$ to $4096$ in powers of two at the uncompressed width $d=2048$ and the code widths $\{1280, 1024, 768, 512, 256\}$. For each call, we keep the fastest SM count and report the median time of the slowest rank. We use CUDA graphs for calls of up to $512$ tokens per rank and larger calls run eagerly to match the serving setup (\Cref{tab:serving-settings}). To test how the regimes depend on the hidden size, we repeat the sweep at $d=7168$, the hidden size of DeepSeek-V3 and Kimi-K2.6, and keep everything else unchanged, including the $128$ experts. Its code widths $\{4608, 3584, 2816, 1792, 1024\}$ keep the fractions of $d$ used at $d=2048$, rounded up to the multiples of $256$ that DeepEP's hybrid combine requires. Repeating part of the $d=2048$ sweep on the same nodes (widths $256$ and $2048$ at $1$, $128$ and $4096$ tokens per rank) reproduces it within $4.5\%$.

\textbf{At $d=2048$, compression reduces call time only above about $512$ tokens per rank.} Up to $512$ tokens per rank, the code widths speed up dispatch by at most $1.24\times$ and combine by at most $1.13\times$ despite sending up to $8\times$ fewer bytes. Above $512$, the speedup grows with message size, reaching $1.5$--$2.9\times$ for dispatch and $1.4$--$2.2\times$ for combine at $4096$ tokens per rank. Calls fall into three regimes: dominated by a startup cost, a per-token cost and a per-byte cost.

\textbf{Small calls are dominated by the startup cost.} With one token per rank, a call takes $64$--$65~\mu$s for dispatch and $66$--$74~\mu$s for combine at every width. This startup cost remains at least half of each call up to $128$ and $64$ tokens per rank for dispatch and combine, respectively. Reducing the code width from $d=2048$ to $256$ in this startup-dominated regime makes dispatch at most $1.06\times$ faster and combine at most $1.13\times$ faster. Combine also sums each token's expert outputs and grows faster with token count ($157$--$169~\mu$s against $117$--$124~\mu$s for dispatch at $128$ tokens per rank), so its startup cost stops dominating earlier.

\textbf{Mid-sized calls are dominated by a per-token cost.} From there up to $512$ tokens per rank, time grows with token count but barely with width. At $512$ tokens per rank, reducing the width by $8\times$ makes dispatch only $1.24\times$ faster ($224 \rightarrow 180~\mu$s) and combine only $1.09\times$ faster ($303 \rightarrow 278~\mu$s). Call time therefore depends on the number of tokens and not only on the payload. The same $2$~MiB per rank takes $224~\mu$s as $512$ rows of width $2048$ but $434~\mu$s as $4096$ rows of width $256$.

\textbf{Large calls are dominated by a per-byte cost.} From $1024$ tokens per rank, dispatch becomes bandwidth-bound and its time grows nearly in proportion to bytes. At $d=2048$, dispatch takes $392$, $681$ and $1242~\mu$s at $1024$, $2048$ and $4096$ tokens per rank. Combine time grows more slowly ($457$, $721$ and $1223~\mu$s) and approaches proportional growth only near $4096$ tokens per rank. The largest dispatch call sends about $43.5$~MiB per GPU over RDMA ($4096$ rows of $4096$ bytes to $\bar m=2.72$ remote nodes under uniform routing) in $1.24$~ms. That is about $290$~Gb/s, below the $400$~Gb/s peak assumed in \Cref{tab:communication-cost}. At $1024$ tokens per rank, the code widths already speed up dispatch by $1.3$--$1.8\times$ and combine by $1.2$--$1.3\times$. These speedups stay below the $1.6$--$8\times$ byte reductions because the per-token cost remains. For combine, width $256$ is barely faster than width $512$ ($568$ against $581~\mu$s at $4096$ tokens per rank).

\textbf{Prefill calls are bandwidth-bound, and decode calls become bandwidth-bound at high concurrency.} With a $4096$-token step budget (\Cref{tab:serving-settings}), prefill chunks reach $4096$ tokens per rank and fall in the bandwidth-bound regime. A decode step processes one token per sequence, so its calls carry as many tokens per rank as there are requests per GPU and become bandwidth-bound as concurrency grows. At $d=2048$, this happens from about $1024$ requests per GPU. At $512$, the most in our workloads, the code widths already speed up dispatch by up to $1.24\times$ per call, and decode speedups grow with the number of requests (\Cref{app:serving-load}). A slower network raises the per-byte cost, so we expect the bandwidth-bound regime to begin at fewer tokens per rank, consistent with the larger gains on capped InfiniBand and TCP (\Cref{app:serving-network}).

\textbf{Larger models with wider rows become bandwidth-bound at fewer tokens per rank.} Larger MoE models such as DeepSeek-V3 and Kimi-K2.6 use $d=7168$, so each token carries $3.5\times$ more bytes than in Qwen3-30B-A3B, while the dispatch startup cost is unchanged ($64$--$66~\mu$s at one token per rank). Uncompressed dispatch time grows nearly in proportion to bytes from $256$ tokens per rank ($184$, $344$ and $622~\mu$s at $128$, $256$ and $512$), compared with $1024$ at $d=2048$. Both transitions occur at $3.5$--$4$~MiB per rank. Combine reaches proportional growth from $1024$ tokens per rank at $d=7168$ and near $4096$ at $d=2048$, again at a similar $14$--$16$~MiB per rank. The bytes per call, rather than the token count, therefore appear to set the threshold. As a result, a $d/4$-wide code speeds up dispatch by $2.3\times$ and $3.0\times$ at $256$ and $512$ tokens per rank, compared with $1.10\times$ and $1.21\times$ at $d=2048$, and combine by $1.7\times$ and $2.0\times$, compared with $1.05\times$ and $1.07\times$ (\Cref{fig:deepep-scaling-dispatch-7168,fig:deepep-scaling-combine-7168}). At $4096$ tokens per rank, the narrowest code ($1024$, $7\times$ fewer bytes) is $5.6\times$ faster for dispatch and $4.8\times$ faster for combine. At $d=2048$, width $256$ sends $8\times$ fewer bytes but is only $2.9\times$ and $2.2\times$ faster. The startup-dominated regime also ends earlier, at $64$ tokens per rank for both calls. For decode, these models therefore need about $4\times$ lower concurrency than Qwen3-30B-A3B before dispatch calls become bandwidth-bound ($256$ against $1024$ requests per GPU), so we expect EchoMoE's savings to reach decode across a wider range of serving loads on larger models.

\section{Inference Benchmarks}
\label{app:serving}

\subsection{Benchmark Setup}
\label{app:serving-setup}

\textbf{Hardware and runtime.} We use $4$ DGX H100 nodes, each with $8$ H100 SXM5 80GB GPUs and $8$ ConnectX-7 $400$ Gb/s NICs. Inference uses vLLM and DeepEP with EP32, TP1 and DP32. Each GPU serves its owns requests and stores $4$ experts per layer for Qwen3-30B-A3B. We benchmark the same adapted checkpoints used for task evaluation. EchoMoE and the uncompressed model use the same hardware allocation and engine settings (\Cref{tab:serving-settings}). In both models, a measured cost table chooses for each dispatch, combine and reference call on either the standard collective (DeepEP or NCCL) or our two-hop kernel, whichever is faster. The choice does not change outputs (\Cref{app:integration}).

\begin{table}[t]
\centering\tablestyle[8pt]
\begin{tabular}{@{}ll@{}}
\toprule
Setting & Value \\
\midrule
Prefill scheduling & Chunked prefill \\
Token budget per step & $4096$ \\
Activation transfers & BF16 \\
Collectives & DeepEP V2 hybrid mode or two-hop kernels, per call \\
Expert assignment & Balanced \\
DP synchronization & NCCL \\
CUDA graphs & Decode steps \\
\texttt{torch.compile} & Disabled \\
End-of-sequence stopping & Disabled \\
Warmup repetitions & $1$ \\
Timed repetitions & $2$ \\
\bottomrule
\end{tabular}
\caption{\textbf{Inference settings for Qwen3-30B-A3B.} EchoMoE and the uncompressed model use the same settings across the $5$ workloads in \Cref{tab:serving-workloads}. Settings for larger architectures are listed in \Cref{app:serving-large}.}
\label{tab:serving-settings}
\end{table}

\textbf{Balanced expert assignment.} Random-token prompts skew the model's own routing far more than real text. Across different EchoMoE configurations we measured, the busiest GPU carried $4.95$--$5.16\times$ the mean expert load on random-token prompts, compared with $2.21$--$2.23\times$ on evaluation text. We therefore give both models identical random-token prompts, seeded by rank and request, and use vLLM's routing simulator to assign experts uniformly at random. \Cref{app:serving-online} serves real prompts with each model's own routing.

\textbf{Throughput and latency.} We submit all requests together and generate a fixed number of output tokens. Throughput is the output-token count divided by batch completion time. Time to first token (TTFT) measures the interval from submission to the first generated token, including queueing. We average timings over $2$ repetitions after $1$ warmup. The reported repetitions agree within $0.5\%$, with settings affected by network contention repeated. These timings describe batch completion instead of steady-state request arrivals.

\textbf{Decode latency.} Time per output token (TPOT) divides the interval between the first and last generated tokens by the number of remaining output tokens. To reduce overlap with prefill, we report the mean TPOT of the $10\%$ of requests with the latest first token across all $32$ ranks, averaged over repetitions. These requests decode mostly after prefill ends, although concurrency falls as requests finish. Unlike the profiling step timings in \Cref{app:serving-profile}, this statistic includes host overhead.

\textbf{Reference transfers.} At reference layers, the Qwen3-30B-A3B implementation broadcasts full-width states with an all-gather and returns expert outputs with a reduce-scatter, each on NCCL or our two-hop kernels. Other layers dispatch and combine at the configured code widths. RDMA reductions are computed from tensor sizes and routing traces (\Cref{app:rdma-bytes}) and not measured from these collectives.

\subsection{Serving Stack Integration}
\label{app:integration}

\textbf{EchoMoE runs on a standard serving stack.} \Cref{tab:integration} lists what each part of an MoE serving stack needs. The compression projections fold into the expert weights (\Cref{subsec:echo-adapt}), so the existing fused MoE kernels run them as ordinary experts with different input and output widths. DeepEP carries the narrower rows unchanged and the reference uses standard NCCL collectives. Our two-hop dispatch and combine kernels are optional and exact. With them, all $768$ prompts generate the same tokens and log-probabilities as with DeepEP alone in both BF16 and FP8.

\begin{table}[htbp]
\centering\tablestyle[5pt]
\begin{tabular}{@{}>{\raggedright\arraybackslash}p{0.4\linewidth}>{\raggedright\arraybackslash}p{0.55\linewidth}@{}}
\toprule
Component & Role in EchoMoE \\
\midrule
\multicolumn{2}{@{}l}{\emph{Existing components}} \\
Expert kernels (fused MoE GEMMs) & Run the folded weights as ordinary expert weights with different input and output widths \\
Dispatch and combine kernels (DeepEP) & Carry the narrower rows, with an optional hybrid-mode connection patch that the uncompressed model also uses \\
Reference collectives (NCCL) & Broadcast full-width states at reference layers \\
\midrule
\multicolumn{2}{@{}l}{\emph{Added components}} \\
Owner-rank compression kernels & Encode dispatch codes and decode combine codes \\
Two-hop dispatch and combine kernels & Optionally replace DeepEP or NCCL calls that a measured cost table predicts to be slower, with identical outputs \\
\bottomrule
\end{tabular}
\caption{\textbf{Serving-stack integration of EchoMoE.} EchoMoE reuses the existing kernels and collectives of an MoE serving stack and adds owner-rank compression kernels and optional two-hop kernels.}
\label{tab:integration}
\end{table}

\subsection{Throughput and Latency Across Workloads}
\label{app:serving-workloads}

\Cref{tab:serving-workloads} reports all $11$ configurations across $5$ workloads. For the $128000$-token workload, both models use YaRN scaling with factor $4$ over $32768$ positions.

\textbf{Longer generation reduces the throughput benefit of faster prefill.} For $(4,512,768)$ at $512$ requests per GPU, TTFT improves by $1.31\times$ with $1024$ input and $128$ output tokens and by $1.30\times$ with $128$ input and $1024$ output tokens. Throughput improves by $1.26\times$ and $1.11\times$, respectively, because decode takes a larger share of completion time in the later workload.

\textbf{Decode improvements depend on the workload.} For $(4,512,1280)$, the decode speedup is $0.93\times$ at $128000$ input and $512$  output tokens with $4$ requests per GPU, compared to $1.04\times$ with $1024$ input and $128$ output tokens with $512$ requests. More aggressive compression $(8,256,512)$ reaches $1.27\times$ in the latter workload, both configurations with dispatch width $1280$ provides little decode benefit. Context length and request count varies together in these workloads whereas \Cref{app:serving-load} varies them independently.

\begin{table}[htbp]
\centering\tablestyle[4pt]
\begin{tabular}{@{}cccrrrrr@{}}
\toprule
\multicolumn{3}{c}{Configuration} & \multicolumn{5}{c}{Input / output tokens (requests per GPU)} \\
\cmidrule(lr){1-3}\cmidrule(lr){4-8}
\makecell[b]{Stride\\$s$} & \makecell[b]{Dispatch\\$r_d$} & \makecell[b]{Combine\\$r_c$} & \makecell[b]{1024/128\\(512)} & \makecell[b]{4096/256\\(128)} & \makecell[b]{16384/512\\(32)} & \makecell[b]{128000/512\\(4)} & \makecell[b]{128/1024\\(512)} \\
\midrule
\multicolumn{8}{@{}l}{\emph{Throughput speedup ($\times$)} } \\
4 & 1280 & 1280 & $1.024$ & $1.025$ & $1.013$ & $0.991$ & $0.974$ \\
6 & 1280 & 1280 & $1.058$ & $1.055$ & $1.031$ & $1.002$ & $0.991$ \\
4 & 512 & 1280 & $1.126$ & $1.115$ & $1.070$ & $1.012$ & $1.020$ \\
4 & 512 & 1024 & $1.192$ & $1.169$ & $1.116$ & $1.039$ & $1.054$ \\
4 & 512 & 768 & $1.261$ & $1.226$ & $1.154$ & $1.052$ & $1.105$ \\
6 & 512 & 768 & $1.323$ & $1.276$ & $1.186$ & $1.058$ & $1.141$ \\
4 & 512 & 512 & $1.333$ & $1.287$ & $1.194$ & $1.075$ & $1.151$ \\
12 & 768 & 512 & $1.407$ & $1.348$ & $1.230$ & $1.083$ & $1.179$ \\
6 & 256 & 768 & $1.383$ & $1.325$ & $1.218$ & $1.078$ & $1.183$ \\
6 & 256 & 512 & $1.468$ & $1.396$ & $1.261$ & $1.093$ & $1.230$ \\
8 & 256 & 512 & $1.515$ & $1.429$ & $1.280$ & $1.101$ & $1.256$ \\
\midrule
\multicolumn{8}{@{}l}{\emph{TTFT speedup ($\times$)} } \\
4 & 1280 & 1280 & $1.036$ & $1.037$ & $1.037$ & $1.002$ & $1.033$ \\
6 & 1280 & 1280 & $1.074$ & $1.073$ & $1.060$ & $1.015$ & $1.060$ \\
4 & 512 & 1280 & $1.152$ & $1.150$ & $1.124$ & $1.032$ & $1.153$ \\
4 & 512 & 1024 & $1.227$ & $1.221$ & $1.193$ & $1.054$ & $1.228$ \\
4 & 512 & 768 & $1.307$ & $1.300$ & $1.258$ & $1.076$ & $1.300$ \\
6 & 512 & 768 & $1.378$ & $1.367$ & $1.310$ & $1.085$ & $1.383$ \\
4 & 512 & 512 & $1.394$ & $1.384$ & $1.327$ & $1.097$ & $1.397$ \\
12 & 768 & 512 & $1.477$ & $1.463$ & $1.387$ & $1.110$ & $1.489$ \\
6 & 256 & 768 & $1.449$ & $1.434$ & $1.368$ & $1.105$ & $1.454$ \\
6 & 256 & 512 & $1.552$ & $1.536$ & $1.447$ & $1.120$ & $1.548$ \\
8 & 256 & 512 & $1.605$ & $1.584$ & $1.482$ & $1.131$ & $1.621$ \\
\midrule
\multicolumn{8}{@{}l}{\emph{Decode TPOT speedup ($\times$)} } \\
4 & 1280 & 1280 & $0.973$ & $0.998$ & $0.974$ & $0.941$ & $0.970$ \\
6 & 1280 & 1280 & $1.001$ & $1.013$ & $0.980$ & $0.948$ & $0.987$ \\
4 & 512 & 1280 & $1.037$ & $1.037$ & $0.987$ & $0.927$ & $1.013$ \\
4 & 512 & 1024 & $1.080$ & $1.057$ & $1.002$ & $0.975$ & $1.046$ \\
4 & 512 & 768 & $1.120$ & $1.077$ & $1.008$ & $0.953$ & $1.095$ \\
6 & 512 & 768 & $1.163$ & $1.098$ & $1.017$ & $0.948$ & $1.129$ \\
4 & 512 & 512 & $1.154$ & $1.098$ & $1.016$ & $0.984$ & $1.140$ \\
12 & 768 & 512 & $1.211$ & $1.131$ & $1.027$ & $0.977$ & $1.165$ \\
6 & 256 & 768 & $1.196$ & $1.117$ & $1.023$ & $0.965$ & $1.171$ \\
6 & 256 & 512 & $1.237$ & $1.141$ & $1.030$ & $0.982$ & $1.216$ \\
8 & 256 & 512 & $1.270$ & $1.154$ & $1.034$ & $0.982$ & $1.240$ \\
\bottomrule
\end{tabular}
\caption{\textbf{Throughput and latency across workloads.} All configurations use full-model adaptation, EP32 over $4$ nodes, balanced expert assignment and BF16 communication. Throughput speedup divides EchoMoE throughput by the mean throughput of the uncompressed runs. Latency speedups use the inverse ratio, so values above $1$ indicate improvements in  throughout. Decode TPOT uses the $10\%$ of requests with the latest first token (\Cref{app:serving-setup}).}
\label{tab:serving-workloads}
\end{table}

\subsection{Decode Speedup Across Load and Context Length}
\label{app:serving-load}

\begin{figure}[h]
\centering
\includegraphics[width=\linewidth]{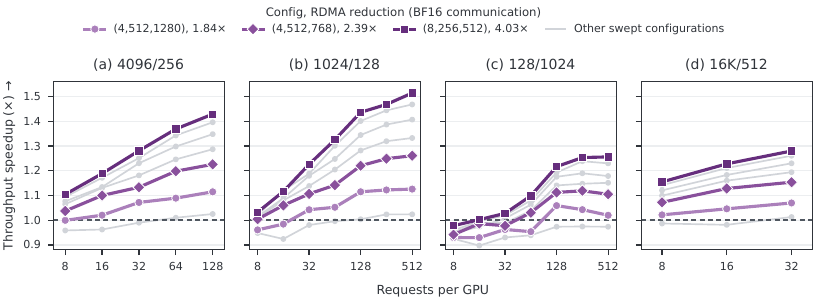}
\caption{\textbf{Throughput speedup grows with requests per GPU.} BF16 communication across four workloads, with the three highlighted configurations and the other swept configurations in grey. The dashed line marks the uncompressed model.}
\label{fig:throughput-concurrency}
\end{figure}

\begin{figure}[h]
\centering
\includegraphics[width=\linewidth]{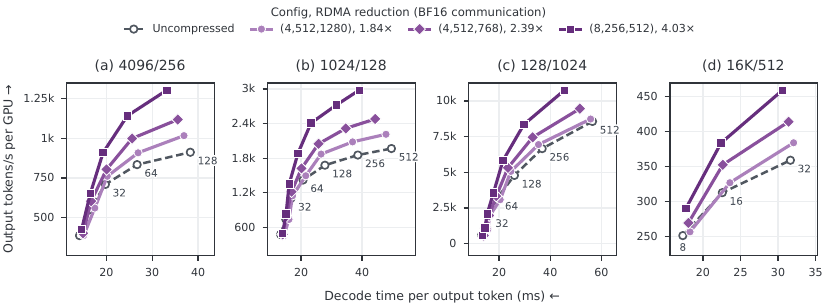}
\caption{\textbf{Throughput against decode latency.} Output tokens per second per GPU against time per output token, BF16 communication. Labels give requests per GPU for the uncompressed model. Curves above and to the left of the uncompressed model deliver more throughput at the same latency.}
\label{fig:latency-throughput}
\end{figure}

\textbf{Decode gains grow with load, not with context.} \Cref{tab:decode-grid} varies requests per GPU at each workload. At a fixed number of requests per GPU, the decode speedup barely changes from $128/1024$ to $16384/512$ input/output tokens. It rises steadily with the number of requests, reaching $1.24$--$1.27\times$ at $512$ for $(8,256,512)$ and $1.10$--$1.12\times$ for $(4,512,768)$. The KV cache limits the $128000/512$ workload to $4$ requests per GPU, so its decode speedup of $0.93$--$0.98\times$ matches shorter workloads at low concurrency. Faster prefill still improves its throughput by up to $1.10\times$ and TTFT by up to $1.13\times$ (\Cref{tab:serving-workloads}). With only a few tokens per rank, each dispatch or combine call costs a nearly fixed time at every width, so narrower codes save little until a step carries a few hundred tokens per rank (\Cref{app:deepep-scaling}). Throughput follows the same trend (\Cref{fig:throughput-concurrency}).

\textbf{TTFT improves at low concurrency and TPOT at higher concurrency.} At $4096/256$ with $8$ requests per GPU, TTFT improves by $1.16$--$1.61\times$ (\Cref{tab:low-concurrency}). With more requests per GPU, EchoMoE also achieves higher throughput and lower TPOT than the uncompressed model, and more aggressive compression  does so at lower concurrency (\Cref{tab:faster-both}). In this regime, EchoMoE achieves higher throughput at the same TPOT (\Cref{fig:latency-throughput}).

\begin{table}[t]
\centering\tablestyle[5pt]
\begin{tabular}{@{}lrrrrrrr@{}}
\toprule
\multirow{2}{*}{Workload} & \multicolumn{7}{c}{Requests per GPU} \\
\cmidrule(lr){2-8}
 & $8$ & $16$ & $32$ & $64$ & $128$ & $256$ & $512$ \\
\midrule
\multicolumn{8}{@{}l}{\emph{$(4,512,1280)$}} \\
128/1024 & $0.93$ & $0.93$ & $0.96$ & $0.95$ & $1.06$ & $1.04$ & $1.01$ \\
1024/128 & $0.93$ & $0.93$ & $0.97$ & $0.96$ & $1.05$ & $1.05$ & $1.04$ \\
4096/256 & $0.94$ & $0.93$ & $0.98$ & $0.99$ & $1.04$ & -- & -- \\
16384/512 & $0.95$ & $0.96$ & $0.99$ & -- & -- & -- & -- \\
\midrule
\multicolumn{8}{@{}l}{\emph{$(4,512,768)$}} \\
128/1024 & $0.94$ & $0.98$ & $0.97$ & $1.03$ & $1.11$ & $1.11$ & $1.10$ \\
1024/128 & $0.95$ & $0.99$ & $0.99$ & $1.03$ & $1.08$ & $1.11$ & $1.12$ \\
4096/256 & $0.95$ & $0.98$ & $0.99$ & $1.04$ & $1.08$ & -- & -- \\
16384/512 & $0.96$ & $1.00$ & $1.01$ & -- & -- & -- & -- \\
\midrule
\multicolumn{8}{@{}l}{\emph{$(8,256,512)$}} \\
128/1024 & $0.98$ & $1.00$ & $1.02$ & $1.09$ & $1.21$ & $1.24$ & $1.24$ \\
1024/128 & $0.95$ & $0.99$ & $1.03$ & $1.09$ & $1.20$ & $1.22$ & $1.27$ \\
4096/256 & $0.97$ & $0.99$ & $1.03$ & $1.09$ & $1.15$ & -- & -- \\
16384/512 & $0.98$ & $1.01$ & $1.03$ & -- & -- & -- & -- \\
\bottomrule
\end{tabular}
\caption{\textbf{Decode TPOT speedup by workload and requests per GPU.} All runs use EP32 over $4$ nodes, balanced expert assignment and BF16 communication. Rows give input/output tokens within each configuration $(s,r_d,r_c)$. Dashes mark workloads not run because the KV cache limits concurrency at longer contexts.}
\label{tab:decode-grid}
\end{table}

\begin{table}[t]
\centering\tablestyle[6pt]
\begin{tabular}{@{}lcccc@{}}
\toprule
\multirow{2}{*}{\makecell[b]{Configuration\\$(s, r_d, r_c)$}} & \multicolumn{2}{c}{4096/256 ($8$ requests per GPU)} & \multicolumn{2}{c}{128000/512 ($4$ requests per GPU)} \\
\cmidrule(lr){2-3}\cmidrule(lr){4-5}
 & Throughput & TTFT & Throughput & TTFT \\
\midrule
$(4,512,1280)$ & $1.00$ & $1.16$ & $1.01$ & $1.03$ \\
$(4,512,768)$ & $1.04$ & $1.31$ & $1.05$ & $1.08$ \\
$(8,256,512)$ & $1.10$ & $1.61$ & $1.10$ & $1.13$ \\
\bottomrule
\end{tabular}
\caption{\textbf{Throughput and TTFT speedups at low concurrency.} All runs use EP32 over $4$ nodes, balanced expert assignment and BF16 communication. Column groups give input/output tokens with requests per GPU in parentheses. Speedups are relative to the mean of the uncompressed runs.}
\label{tab:low-concurrency}
\end{table}

\begin{table}[!ht]
\centering\tablestyle[6pt]
\begin{tabular}{@{}lcccc@{}}
\toprule
\multirow{2}{*}{\makecell[b]{Configuration\\$(s, r_d, r_c)$}} & \multicolumn{4}{c}{Input / output tokens} \\
\cmidrule(lr){2-5}
 & 128/1024 & 1024/128 & 4096/256 & 16384/512 \\
\midrule
$(4,512,1280)$ & $128$ & $128$ & $128$ & -- \\
$(4,512,768)$ & $64$ & $64$ & $64$ & $32$ \\
$(8,256,512)$ & $32$ & $32$ & $32$ & $16$ \\
\bottomrule
\end{tabular}
\caption{\textbf{Lowest concurrency at which EchoMoE improves both throughput and TPOT.} EchoMoE has higher throughput and lower TPOT than the uncompressed model at and above the listed number of requests per GPU. All runs use EP32 over $4$ nodes, balanced expert assignment and BF16 communication. The dash marks no such count up to $32$ requests per GPU, the largest that fits in the KV cache at $16384/512$.}
\label{tab:faster-both}
\end{table}

\subsection{Communication and Computation Costs}
\label{app:serving-profile}

\begin{figure}[htbp]
\centering
\includegraphics[width=\linewidth]{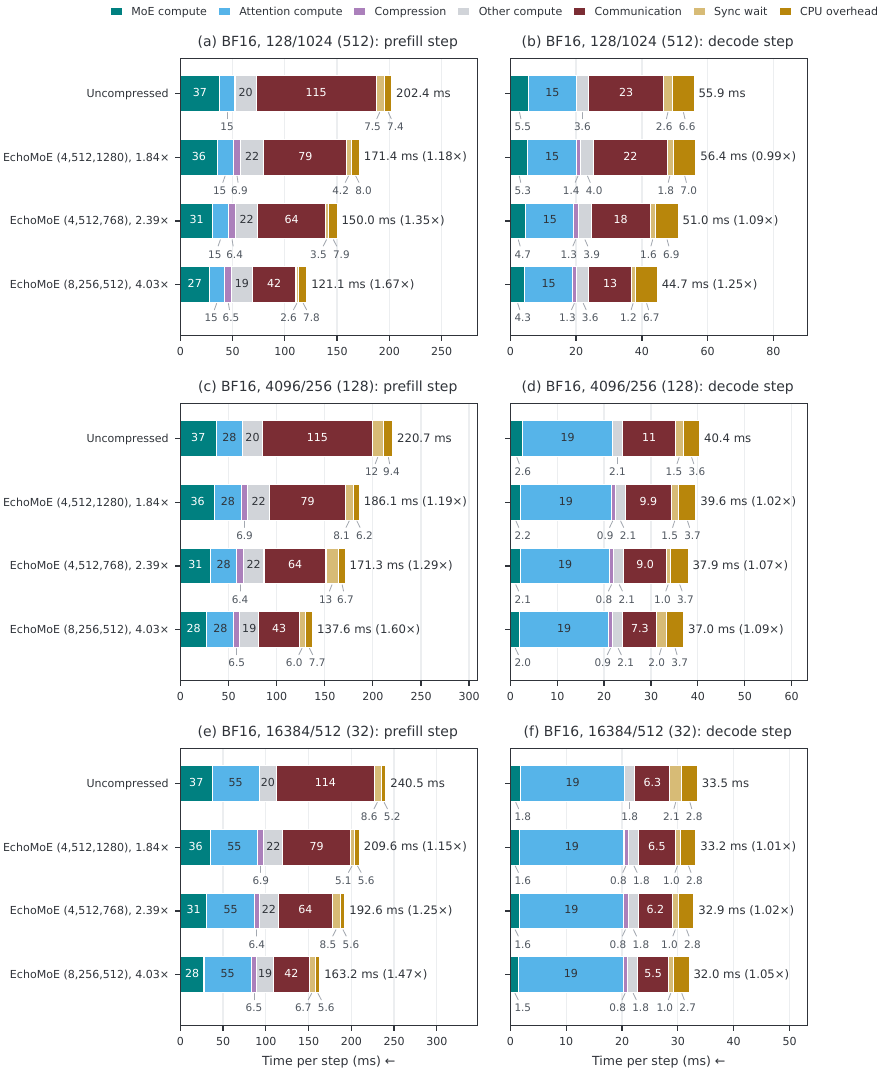}
\caption{\textbf{Prefill and decode step breakdown across workloads.} All runs use EP32 over $4$ nodes, balanced expert assignment and BF16 communication. Panel titles give input/output tokens and requests per GPU. Communication includes the reference collective. Labels give each component's time in ms and each step's speedup over the uncompressed model.}
\label{fig:breakdown}
\end{figure}

We profile the uncompressed model and three configurations at three workloads: $128/1024$ with $512$ requests per GPU, $4096/256$ with $128$ and $16384/512$ with $32$ (\Cref{fig:breakdown}). We remove the decode-graph launch overhead that the PyTorch profiler adds ($1.5$--$1.9$ ms per step for the uncompressed model and $0.1$--$0.3$ ms for EchoMoE). After removing this overhead, the profiled decode speedups match the measured TPOT.

\textbf{Communication dominates prefill.} Communication takes $48$--$57\%$ of an uncompressed prefill step at every workload. EchoMoE removes $31$--$63\%$ of this time, including the reference collective. Compression kernels add $6.4$--$6.9$ ms per prefill step and $0.8$--$1.4$ ms per decode step.

\textbf{Attention limits the relative benefit at long context.} At $16384/512$, attention takes $55$ ms of an uncompressed prefill step, compared with $15$ ms at $128/1024$. The same communication savings therefore reduces the step by $20\%$ for $(4,512,768)$ at $16384/512$ instead of $26\%$ at $128/1024$. In decode, attention grows with context from $26\%$ to $56\%$ of the step, while communication grows with requests per GPU from $19\%$ to $41\%$.

\textbf{Reference collective offset more of the communication savings during decode.} In prefill, the reference collective costs $17$--$50\%$ of the dispatch and combine time it saves (\Cref{tab:reference-saving}). In decode, it costs $25$--$93\%$ at $128$ or more requests per GPU, the most for $(4,512,1280)$. At $16384/512$ with $32$ requests per GPU, the reference offsets at least $95\%$ of the decode saving for both stride $4$ configurations. Stride $8$ halves the reference traffic and keeps $45\%$ of the saving.

\textbf{Narrower expert outputs also reduce computation.} Folding narrows the expert down projection and output sum, so MoE computation in a prefill step at $128/1024$ falls from $37.1$ ms to $35.5$, $31.0$ and $27.5$ ms for the three configurations. Throughput improvements therefore come from both communication and computation.

\begin{table}[t]
\centering\tablestyle[5pt]
\begin{tabular}{@{}lrrrrrr@{}}
\toprule
\multirow{2}{*}{\makecell[b]{Configuration\\($s, r_c, r_d$)}} & \multicolumn{3}{c}{Prefill (ms)} & \multicolumn{3}{c}{Decode (ms)} \\
\cmidrule(lr){2-4}\cmidrule(lr){5-7}
 & Saved & Reference & $\Delta$ & Saved & Reference & $\Delta$ \\
\midrule
\multicolumn{7}{@{}l}{\emph{128/1024, $512$ requests per GPU}} \\
$(4,512,1280)$ & $71.4$ & $35.8$ & $35.6$ & $8.6$ & $8.0$ & $0.6$ \\
$(4,512,768)$ & $82.1$ & $31.5$ & $50.6$ & $11.9$ & $6.9$ & $5.0$ \\
$(8,256,512)$ & $87.8$ & $14.9$ & $72.9$ & $13.5$ & $3.4$ & $10.1$ \\
\midrule
\multicolumn{7}{@{}l}{\emph{4096/256, $128$ requests per GPU}} \\
$(4,512,1280)$ & $71.0$ & $35.8$ & $35.2$ & $5.2$ & $3.7$ & $1.6$ \\
$(4,512,768)$ & $81.7$ & $31.6$ & $50.2$ & $6.0$ & $3.5$ & $2.4$ \\
$(8,256,512)$ & $87.2$ & $15.2$ & $72.0$ & $6.0$ & $1.9$ & $4.1$ \\
\midrule
\multicolumn{7}{@{}l}{\emph{16384/512, $32$ requests per GPU}} \\
$(4,512,1280)$ & $70.8$ & $35.5$ & $35.3$ & $2.0$ & $2.2$ & $-0.2$ \\
$(4,512,768)$ & $81.5$ & $31.2$ & $50.3$ & $2.3$ & $2.2$ & $0.1$ \\
$(8,256,512)$ & $87.3$ & $15.0$ & $72.4$ & $1.9$ & $1.0$ & $0.8$ \\
\bottomrule
\end{tabular}
\caption{\textbf{Communication time saved and reference cost per step.} All runs use EP32 over $4$ nodes, balanced expert assignment and BF16 communication. Saved denotes the dispatch and combine time removed by the narrower codes, Reference denotes the time of the reference collective and $\Delta$ denotes their difference.}
\label{tab:reference-saving}
\end{table}

\subsection{FP8 Communication End to End}
\label{app:fp8-serving}

We serve both models with FP8 activations in every dispatch and reference transfer and in the inter-node part of every combine, on the same hardware and workloads as the BF16 results.

\textbf{EchoMoE and FP8 compose.} Over the uncompressed BF16 model, EchoMoE with FP8 communication improves throughput by $1.33\times$ for $(4,512,768)$ and $1.51\times$ for $(8,256,512)$, compared with $1.18\times$ from FP8 alone (\Cref{tab:fp8-summary}). Switching an adapted model to FP8 needs no retraining and changes overall task performance by at most $0.29$ pp (\Cref{tab:fp8}). Each configuration's gap to the uncompressed model under FP8 is within $0.4$ pp of its BF16 gap.

\begin{figure}[t]
\centering
\includegraphics[width=\linewidth]{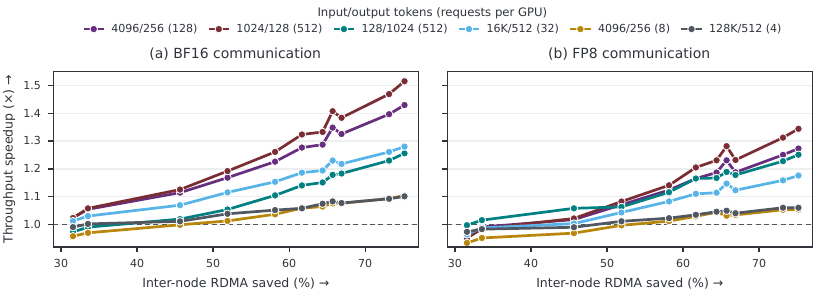}
\caption{\textbf{Throughput speedup against RDMA savings across workloads.} Different configurations with \textbf{(a)} BF16 and \textbf{(b)} FP8 communication relative to the uncompressed model with the same communication precision.}
\label{fig:throughput-workloads}
\end{figure}

\begin{figure}[t]
\centering
\includegraphics[width=\linewidth]{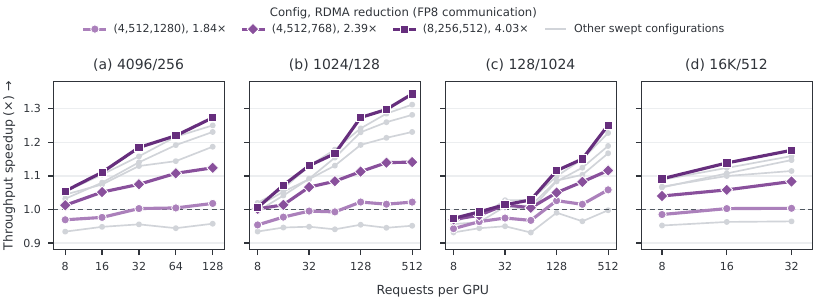}
\caption{\textbf{Throughput speedup against requests per GPU with FP8 communication.} Speedups are relative to the uncompressed FP8 model, with the three highlighted configurations and the other swept configurations in grey. The dashed line marks the uncompressed model.}
\label{fig:throughput-concurrency-fp8}
\end{figure}

\begin{figure}[!t]
\centering
\includegraphics[width=\linewidth]{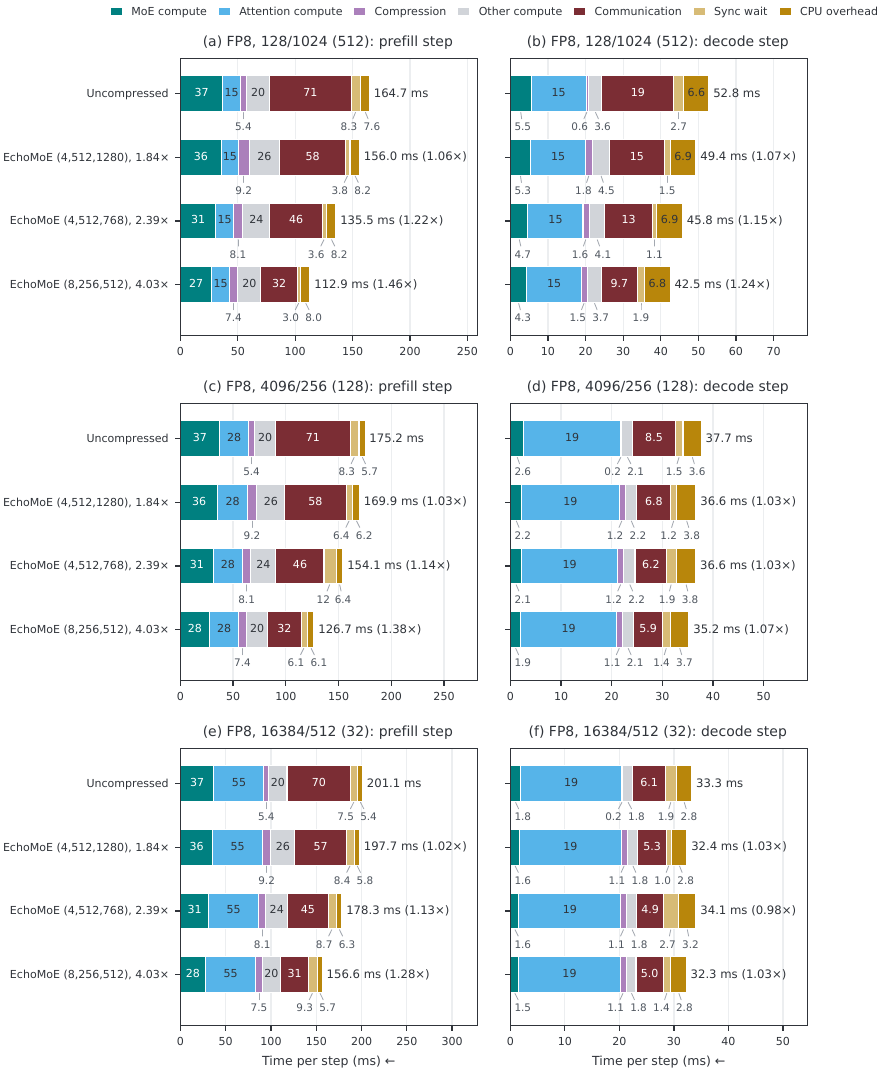}
\caption{\textbf{Prefill and decode step breakdown with FP8 communication.} All runs use EP32 over $4$ nodes, balanced expert assignment and FP8 communication. Panel titles give input/output tokens and requests per GPU. Communication includes the reference collective. Labels give each component's time in ms and each step's speedup over the uncompressed model.}
\label{fig:breakdown-fp8}
\end{figure}

\textbf{The gains hold across workloads.} Over the uncompressed FP8 model, $(8,256,512)$ improves throughput up to $1.34\times$ and TTFT by $1.41\times$, and $(4,512,768)$ up to $1.14\times$ and $1.17\times$ (\Cref{tab:fp8-workloads} and \Cref{fig:throughput-workloads}). Decode gains grow with concurrency similar to BF16 up to $1.24\times$ (\Cref{fig:throughput-concurrency-fp8}). The $(4,512,1280)$ configuration reaches only $1.06\times$ for a $2.0$ pp drop, so with FP8 communication, $(4,512,768)$ or stronger compression is the useful range. \Cref{fig:breakdown-fp8} breaks down the FP8 steps.

\textbf{Memory and computation.} EchoMoE takes $1.2$--$1.7$ GiB more per GPU to load, including the reference buffer. However, the KV cache that vLLM allocates afterwards changes by only $-0.8\%$ to $+2.3\%$ (\Cref{tab:fp8-memory}). Folding also gives $(4,512,768)$ and $(8,256,512)$ fewer matrix MACs than the uncompressed model (\Cref{tab:local-costs}).

\begin{table}[h]
\centering\tablestyle[5pt]
\begin{tabular}{@{}lrrr@{}}
\toprule
\multirow{2}{*}{\makecell[b]{Configuration\\($s, r_c, r_d$)}} & & \multicolumn{2}{c}{Throughput speedup ($\times$)} \\
\cmidrule(lr){3-4}
 & Overall (\%) & vs.\ FP8 & vs.\ BF16 \\
\midrule
\emph{Qwen3-30B-A3B ($d = 2048$, $128$ experts)} & 66.83 & $1.00$ & $1.18$ \\
$(4,512,1280)$ & 64.86\,\scorechange{$-$2.0} & $1.02$ & $1.21$ \\
$(4,512,768)$ & 62.99\,\scorechange{$-$3.8} & $1.12$ & $1.33$ \\
$(8,256,512)$ & 55.56\,\scorechange{$-$11.3} & $1.27$ & $1.51$ \\
\bottomrule
\end{tabular}
\caption{\textbf{EchoMoE with FP8 communication.} All rows use FP8 dispatch, reference and combine transfers. Overall task performance is from \Cref{tab:fp8}. Throughput uses EP32 over $4$ nodes, balanced expert assignment and $4096/256$ with $128$ requests per GPU, and is relative to the uncompressed Qwen3-30B-A3B model with FP8 or BF16 communication.}
\label{tab:fp8-summary}
\end{table}

\begin{table}[t]
\centering\tablestyle[4pt]
\begin{tabular}{@{}lrrrrrr@{}}
\toprule
& \multicolumn{6}{c}{Input / output tokens (requests per GPU)} \\
\cmidrule(lr){2-7}
Configuration & \makecell[b]{1024/128\\(512)} & \makecell[b]{4096/256\\(128)} & \makecell[b]{16384/512\\(32)} & \makecell[b]{128000/512\\(4)} & \makecell[b]{128/1024\\(512)} & \makecell[b]{4096/256\\(8)} \\
\midrule
\multicolumn{7}{@{}l}{\emph{Throughput speedup ($\times$)}} \\
$(4,512,1280)$ & $1.02$ & $1.02$ & $1.00$ & $0.99$ & $1.06$ & $0.97$ \\
$(4,512,768)$ & $1.14$ & $1.12$ & $1.08$ & $1.02$ & $1.12$ & $1.01$ \\
$(8,256,512)$ & $1.34$ & $1.27$ & $1.18$ & $1.06$ & $1.25$ & $1.05$ \\
\midrule
\multicolumn{7}{@{}l}{\emph{TTFT speedup ($\times$)}} \\
$(4,512,1280)$ & $1.02$ & $1.02$ & $1.01$ & $0.99$ & $1.02$ & $1.02$ \\
$(4,512,768)$ & $1.16$ & $1.17$ & $1.14$ & $1.03$ & $1.16$ & $1.16$ \\
$(8,256,512)$ & $1.40$ & $1.38$ & $1.31$ & $1.08$ & $1.41$ & $1.39$ \\
\midrule
\multicolumn{7}{@{}l}{\emph{Decode TPOT speedup ($\times$)}} \\
$(4,512,1280)$ & $1.03$ & $1.01$ & $0.99$ & $0.98$ & $1.06$ & $0.95$ \\
$(4,512,768)$ & $1.09$ & $1.04$ & $1.01$ & $0.98$ & $1.11$ & $0.97$ \\
$(8,256,512)$ & $1.20$ & $1.09$ & $1.02$ & $0.99$ & $1.24$ & $0.97$ \\
\bottomrule
\end{tabular}
\caption{\textbf{Throughput and latency across workloads with FP8 communication.} Speedups are relative to the uncompressed model with the same FP8 communication. All runs use EP32 over $4$ nodes and balanced expert assignment.}
\label{tab:fp8-workloads}
\end{table}

\begin{table}[t]
\centering\tablestyle[6pt]
\begin{tabular}{@{}lrrrr@{}}
\toprule
\multirow{2}{*}{\makecell[b]{Configuration\\($s, r_c, r_d$)}} & & \multicolumn{3}{c}{KV cache} \\
\cmidrule(lr){3-5}
 & Model load (GiB) & Memory (GiB) & Tokens & Change \\
\midrule
\emph{Qwen3-30B-A3B ($d = 2048$, $128$ experts)} & $4.57$ & $57.80$ & $630{,}080$ & -- \\
$(4,512,1280)$ & $6.26$ & $57.34$ & $625{,}056$ & $-0.8\%$ \\
$(4,512,768)$ & $6.03$ & $58.55$ & $638{,}320$ & $+1.3\%$ \\
$(8,256,512)$ & $5.77$ & $59.13$ & $644{,}640$ & $+2.3\%$ \\
\bottomrule
\end{tabular}
\caption{\textbf{Memory per GPU.} Values reported by vLLM for the FP8 runs at EP32 and $4096/256$. Model load includes the reference buffer, and vLLM sizes the KV cache from the memory that remains. Change is relative to the uncompressed KV cache.}
\label{tab:fp8-memory}
\end{table}

\begin{table}[!t]
\centering\tablestyle[6pt]
\begin{tabular}{@{}lcc@{}}
\toprule
\multirow{2}{*}{\makecell[b]{Configuration\\$(s, r_d, r_c)$}} & \multicolumn{2}{c}{Throughput speedup ($\times$)} \\
\cmidrule(lr){2-3}
& InfiniBand & TCP \\
\midrule
$(4,512,1280)$ & $1.08\times$ & $2.04\times$ \\
$(4,512,768)$ & $1.09\times$ & $1.87\times$ \\
$(8,256,512)$ & $1.05\times$ & $2.28\times$ \\
\bottomrule
\end{tabular}
\caption{\textbf{Switching an adapted model from BF16 to FP8 transfers.} Throughput with FP8 dispatch, reference and combine relative to the same adapted model with BF16 transfers. All runs use EP32 over $4$ nodes, balanced expert assignment and $4096/256$, with $128$ requests per GPU over InfiniBand and $8$ over TCP.}
\label{tab:fp8-switch}
\end{table}

\begin{figure}[!t]
\centering
\includegraphics[width=\linewidth]{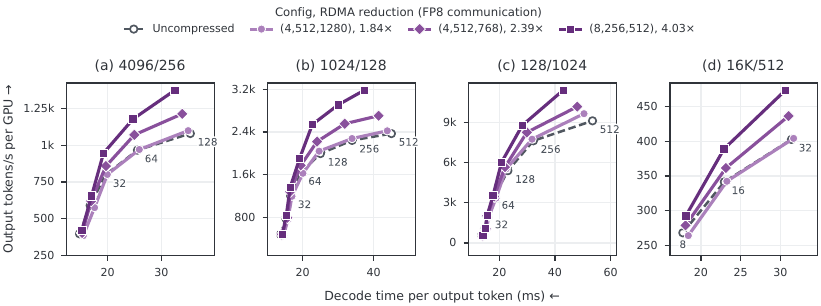}
\caption{\textbf{Throughput against decode latency with FP8 communication.} Output tokens per second per GPU against time per output token, BF16 communication. Labels give requests per GPU for the uncompressed model. Curves above and to the left of the uncompressed model deliver more throughput at the same latency.}
\label{fig:latency-throughput-fp8}
\end{figure}

\subsection{Slower Networks}
\label{app:serving-network}

\textbf{Slower networks gain more.} We cap the InfiniBand NICs near $200$ and $160$ Gb/s, and serve over TCP on Ethernet with about $23$ Gb/s per node. Across different configurations, EchoMoE's speedup grows as the network slows, up to $1.73\times$ with NICs capped near $160$ Gb/s and $2.38\times$ over TCP with BF16 communication (\Cref{tab:networks}). With FP8 communication in both models, the speedup over TCP reaches $3.24\times$. Over TCP, the random-weight larger models listed in \Cref{app:serving-large} reach $1.87$--$1.97\times$ speed up at $8$ requests per GPU.

\begin{table}[t]
\centering\tablestyle[6pt]
\begin{tabular}{@{}llrr@{}}
\toprule
& & \multicolumn{2}{c}{Throughput speedup ($\times$)} \\
\cmidrule(lr){3-4}
Network & Bandwidth & BF16 & FP8 \\
\midrule
InfiniBand & $400$ Gb/s per NIC & $1.02$--$1.43$ & $0.96$--$1.27$ \\
InfiniBand, capped & ${\sim}200$ Gb/s per NIC & $1.03$--$1.40$ & $0.96$--$1.33$ \\
InfiniBand, capped & ${\sim}160$ Gb/s per NIC & $1.11$--$1.73$ & $1.06$--$1.43$ \\
TCP over Ethernet & ${\sim}23$ Gb/s per node & $1.49$--$2.38$ & $1.86$--$3.24$ \\
\bottomrule
\end{tabular}
\caption{\textbf{Throughput speedup across networks.} Ranges over the $11$ configurations at $4096/256$ with $128$ requests per GPU ($8$ over TCP), each relative to the uncompressed model with the same network and communication precision. All runs use EP32 over $4$ nodes and balanced expert assignment.}
\label{tab:networks}
\end{table}

\begin{figure}[t]
\centering
\includegraphics[width=0.55\linewidth]{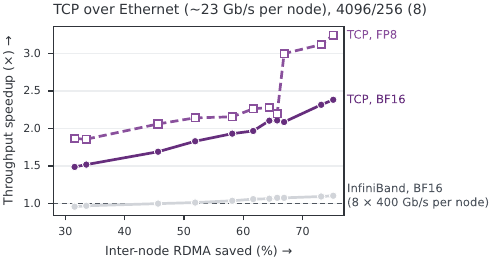}
\caption{\textbf{Throughput speedup over TCP.}  Speedup relative to the uncompressed model with the same communication precision at $4096/256$ with $8$ requests per GPU. The InfiniBand BF16 curve is shown for reference.}
\label{fig:tcp}
\end{figure}

\begin{figure}[t]
\centering
\includegraphics[width=\linewidth]{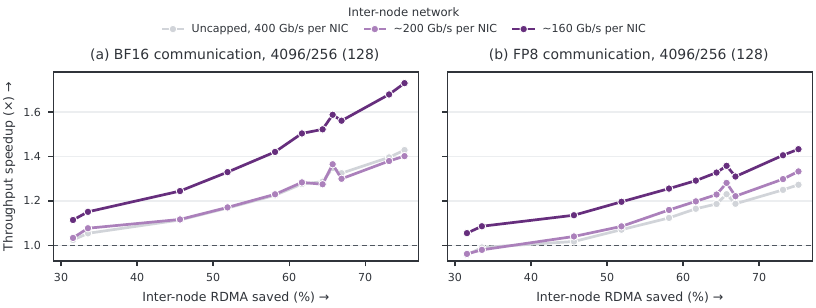}
\caption{\textbf{Throughput speedup with capped InfiniBand against inter-node RDMA saved.} Speedup relative to the uncompressed model on the same network against inter-node RDMA saved with BF16 and FP8 communication at $4096/256$ with $128$ requests per GPU.}
\label{fig:capped}
\end{figure}

\subsection{Online Serving on Real Prompts}
\label{app:serving-online}

\textbf{Setup.} We serve $5{,}017$ held-out Dolci-Instruct prompts at their original lengths and let each reply run till end of sequence.  Both models run on vLLM's server with DP32 and EP32 over $32$ H100 GPUs. Requests arrive as a Poisson stream at $16$ to $512$ per second for $240$ s per rate. We compare $(4,512,768)$ with the uncompressed model under BF16 and FP8 communication, using either each model's own routing or balanced routing. TTFT is measured over requests arriving after the server is full, and TPOT over requests decoding while arrivals continue.

\begin{figure}[t]
\centering
\includegraphics[width=\linewidth]{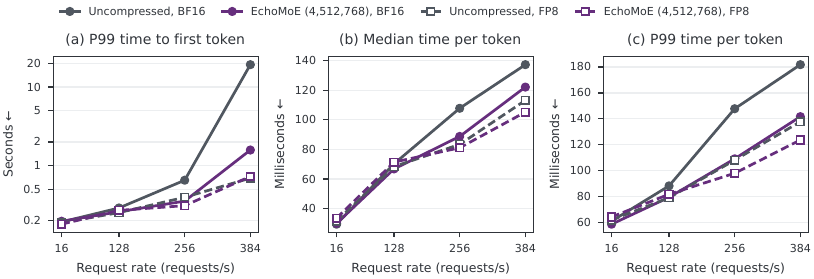}
\caption{\textbf{Online serving on real prompts with each model's own routing.} \textbf{(a)} P99 time to first token, \textbf{(b)} median and \textbf{(c)} P99 time per output token against request rate, uncompressed and EchoMoE $(4,512,768)$ with BF16 and FP8 communication.}
\label{fig:serving-online}
\end{figure}

\textbf{EchoMoE sustains a higher request rate.} At $16$ requests per second, the two models match with P99 TTFT of $193$ and $197$ ms and median TPOT of $30$ ms. At $256$, EchoMoE reduces P99 TTFT from $654$ to $355$ ms and P99 TPOT from $148$ to $109$ ms. At $384$, the uncompressed model saturates with P99 TTFT of $19$ s, while EchoMoE stays at $1.6$ s. With P99 TTFT of at most $2$ s and P99 TPOT of at most $150$ ms, EchoMoE sustains $384$ requests per second compared with $256$ for the uncompressed model (\Cref{tab:serving-online} and \Cref{fig:serving-online}).

\begin{figure}[t]
\centering
\includegraphics[width=\linewidth]{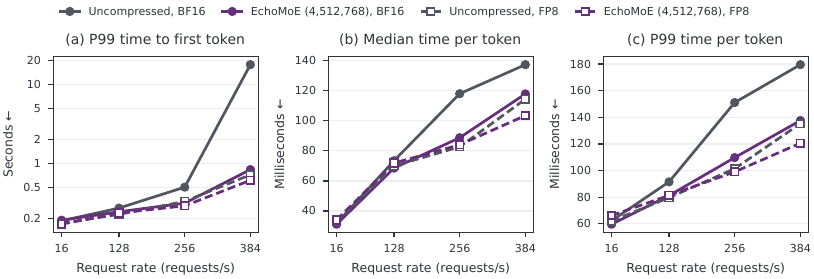}
\caption{\textbf{Online serving with balanced routing and fixed replies.} \textbf{(a)} P99 time to first token, \textbf{(b)} median and \textbf{(c)} P99 time per output token against request rate, uncompressed and EchoMoE $(4,512,768)$ with BF16 and FP8 communication.}
\label{fig:serving-online-balanced}
\end{figure}

\textbf{The gains hold when both models do identical work.} With balanced routing and each reply fixed to the uncompressed model's length, EchoMoE at $256$ requests per second reduces median TPOT from $117.8$ to $88.6$ ms and P99 TTFT from $501$ to $314$ ms (\Cref{fig:serving-online-balanced}).

\textbf{FP8 alone reaches the same request rate.} With FP8 communication, both models sustain $384$ requests per second. At this rate, EchoMoE lowers median TPOT by $7$--$10\%$ and P99 TPOT by $10$--$11\%$. At $256$ requests per second and below, the two are within $3$ ms. At $384$, the uncompressed FP8 model (median $113$--$114$ ms per token) is faster than EchoMoE with BF16 communication ($118$--$122$ ms).

\textbf{Online gains are smaller than offline decode gains.} Under continuous arrivals, most engine steps mix prefill and decode. Our engine uses CUDA graphs only for pure-decode steps and does not use \texttt{torch.compile}, so mixed steps run without graphs and communication is a smaller share of each step. With real routing, EchoMoE's replies are about $5\%$ longer and the comparison includes this additional overhead.

\begin{table}[t]
\centering\tablestyle[5pt]
\begin{tabular}{@{}rrrrr@{}}
\toprule
& \multicolumn{4}{c}{Uncompressed / EchoMoE $(4,512,768)$} \\
\cmidrule(lr){2-5}
Requests/s & P99 TTFT (ms) & Median TPOT (ms) & P99 TPOT (ms) & Output (k tokens/s) \\
\midrule
\multicolumn{5}{@{}l}{\emph{BF16, real routing}} \\
$16$ & $193$ / $197$ & $29.6$ / $30.1$ & $61.1$ / $58.8$ & $5.5$ / $5.9$ \\
$128$ & $290$ / $260$ & $70.3$ / $66.8$ & $88.3$ / $79.5$ & $44.2$ / $45.9$ \\
$256$ & $654$ / $355$ & $107.8$ / $88.8$ & $147.8$ / $109.2$ & $81.6$ / $86.9$ \\
$384$ & $19{,}255$ / $1{,}584$ & $137.2$ / $122.1$ & $181.8$ / $141.7$ & $111.4$ / $122.3$ \\
$512$ & $100{,}558$ / $61{,}027$ & $137.8$ / $118.2$ & $183.2$ / $147.9$ & $109.9$ / $129.6$ \\
\midrule
\multicolumn{5}{@{}l}{\emph{BF16, balanced routing, fixed replies}} \\
$16$ & $188$ / $191$ & $31.7$ / $31.3$ & $62.3$ / $59.6$ & $5.5$ / $5.5$ \\
$128$ & $272$ / $245$ & $73.4$ / $68.5$ & $91.5$ / $81.2$ & $43.8$ / $44.1$ \\
$256$ & $501$ / $314$ & $117.8$ / $88.6$ & $151.1$ / $109.7$ & $80.7$ / $85.8$ \\
$384$ & $17{,}735$ / $835$ & $137.1$ / $117.7$ & $179.6$ / $137.6$ & $111.6$ / $119.5$ \\
$512$ & $97{,}354$ / $49{,}810$ & $136.3$ / $117.6$ & $176.7$ / $146.0$ & $111.1$ / $131.5$ \\
\midrule
\multicolumn{5}{@{}l}{\emph{FP8, real routing}} \\
$16$ & $188$ / $180$ & $31.7$ / $33.6$ & $62.0$ / $64.4$ & $5.6$ / $5.9$ \\
$128$ & $254$ / $271$ & $68.4$ / $71.4$ & $79.2$ / $81.9$ & $44.6$ / $45.1$ \\
$256$ & $397$ / $311$ & $83.4$ / $81.1$ & $108.2$ / $98.0$ & $86.8$ / $89.6$ \\
$384$ & $695$ / $731$ & $113.1$ / $105.2$ & $137.8$ / $123.7$ & $120.0$ / $123.6$ \\
$512$ & $41{,}778$ / $39{,}702$ & $111.9$ / $109.2$ & $142.6$ / $135.7$ & $136.8$ / $140.9$ \\
\midrule
\multicolumn{5}{@{}l}{\emph{FP8, balanced routing, fixed replies}} \\
$16$ & $175$ / $170$ & $32.8$ / $34.1$ & $62.0$ / $66.2$ & $5.5$ / $5.5$ \\
$128$ & $227$ / $236$ & $70.0$ / $71.8$ & $79.7$ / $81.3$ & $43.9$ / $43.8$ \\
$256$ & $331$ / $290$ & $82.6$ / $84.0$ & $101.4$ / $99.0$ & $86.9$ / $87.3$ \\
$384$ & $719$ / $611$ & $114.2$ / $103.3$ & $135.2$ / $120.5$ & $120.3$ / $124.1$ \\
$512$ & $40{,}899$ / $32{,}236$ & $112.4$ / $108.8$ & $139.9$ / $132.6$ & $137.6$ / $142.1$ \\
\bottomrule
\end{tabular}
\caption{\textbf{Online serving on real prompts.} All runs use vLLM's server with DP32 and EP32 on $32$ H100 GPUs, $5$K held-out Dolci-Instruct prompts and Poisson arrivals for $240$ s per rate. Real routing keeps each model's own routing and reply lengths. Balanced routing uses vLLM's routing simulator and fixes each reply to the uncompressed model's length under real routing, so both models prefill and decode the same number of tokens.}
\label{tab:serving-online}
\end{table}

\subsection{Agent and Long-Reasoning Workloads}
\label{app:serving-agent}

Our models are fine-tuned on Dolci-Instruct without tool-call samples (\Cref{app:recipes}), so we do not evaluate them on agentic tasks. Instead, we measure inference cost on two types of workload: agent turns with long contexts and short outputs and long reasoning with short prompts and long outputs.

\textbf{EchoMoE improves throughput and TTFT on agent turns.} At $32$K/$256$ and $64$K/$256$ input/output tokens, throughput improves by $1.04$--$1.28\times$ with BF16 communication and $1.00$--$1.16\times$ with FP8 (\Cref{tab:agent-shapes}). The gain comes from prefill with TTFT improving by up to $1.36\times$ while TPOT stays within $4\%$.

\textbf{EchoMoE improves TTFT on long reasoning with throughput near parity.} We use $1$K/$8$K and $1$K/$16$K input/output tokens, matching our evaluation's generation limits (\Cref{app:tasks}). Decode dominates and throughput is $0.98$--$1.07\times$ with both BF16 and FP8, while TTFT improves by $1.15$--$1.64\times$ with BF16.

\begin{figure}[t]
\centering
\includegraphics[width=\linewidth]{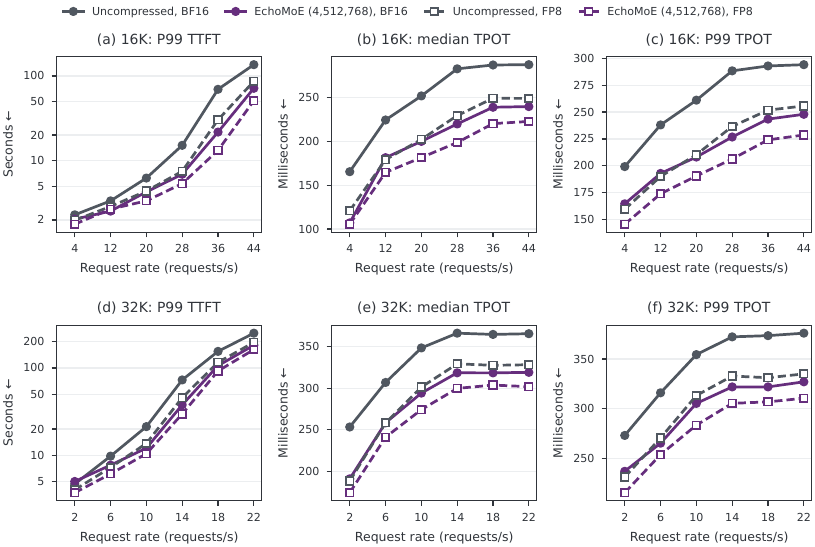}
\caption{\textbf{Agent-like online serving.} P99 TTFT, median TPOT and P99 TPOT against request rate for $16$K-token (top) and $32$K-token (bottom) conversation histories with replies of at most $256$ tokens. Uncompressed model and EchoMoE $(4,512,768)$ each use their own routing and an FP8 KV cache.}
\label{fig:serving-agent}
\end{figure}

\textbf{EchoMoE lowers latency in agent-like online serving.} We serve $16$K and $32$K-token held-out conversation histories with replies of at most $256$ tokens with requests arriving as a Poisson stream. With BF16 communication, EchoMoE $(4,512,768)$ improves median TPOT by $1.15$--$1.54\times$ at every request rate and lowers P99 TTFT before saturation by $11$--$54\%$, except at $32$K and $2$ requests per second, where it is $7\%$ higher (\Cref{fig:serving-agent}). Since almost every engine step includes a prefill chunk that all GPUs wait for, faster prefill helps even at low request rates. With FP8 communication, the gains shrink to $1.07$--$1.15\times$ and $7$--$28\%$.

\begin{table}[htbp]
\centering\tablestyle[4pt]
\begin{tabular}{@{}lrrrrrr@{}}
\toprule
& \multicolumn{3}{c}{BF16 communication} & \multicolumn{3}{c}{FP8 communication} \\
\cmidrule(lr){2-4}\cmidrule(lr){5-7}
Configuration & Throughput & TTFT & TPOT & Throughput & TTFT & TPOT \\
\midrule
\multicolumn{7}{@{}l}{\emph{Agent turn: $32$K/$256$ with $16$ requests per GPU}} \\
$(4,512,1280)$ & $1.08$ & $1.11$ & $0.98$ & $1.00$ & $1.00$ & $0.98$ \\
$(4,512,768)$ & $1.16$ & $1.20$ & $1.01$ & $1.08$ & $1.10$ & $1.01$ \\
$(8,256,512)$ & $1.28$ & $1.36$ & $1.03$ & $1.16$ & $1.21$ & $1.02$ \\
\midrule
\multicolumn{7}{@{}l}{\emph{Agent turn: $64$K/$256$ with $8$ requests per GPU}} \\
$(4,512,1280)$ & $1.04$ & $1.06$ & $0.96$ & $1.00$ & $1.00$ & $0.99$ \\
$(4,512,768)$ & $1.10$ & $1.13$ & $0.97$ & $1.05$ & $1.06$ & $0.99$ \\
$(8,256,512)$ & $1.19$ & $1.23$ & $0.99$ & $1.10$ & $1.13$ & $0.99$ \\
\midrule
\multicolumn{7}{@{}l}{\emph{Long reasoning: $1$K/$8$K with $64$ requests per GPU}} \\
$(4,512,1280)$ & $0.99$ & $1.15$ & $0.99$ & $0.98$ & $1.02$ & $0.98$ \\
$(4,512,768)$ & $1.03$ & $1.31$ & $1.03$ & $1.01$ & $1.17$ & $1.00$ \\
$(8,256,512)$ & $1.07$ & $1.60$ & $1.06$ & $1.03$ & $1.39$ & $1.03$ \\
\midrule
\multicolumn{7}{@{}l}{\emph{Long reasoning: $1$K/$16$K with $32$ requests per GPU}} \\
$(4,512,1280)$ & $0.98$ & $1.16$ & $0.98$ & $0.99$ & $1.02$ & $0.99$ \\
$(4,512,768)$ & $0.99$ & $1.32$ & $0.99$ & $1.01$ & $1.18$ & $1.01$ \\
$(8,256,512)$ & $1.02$ & $1.64$ & $1.02$ & $1.02$ & $1.41$ & $1.02$ \\
\bottomrule
\end{tabular}
\caption{\textbf{Agent and long-reasoning workloads.} Speedups are relative to the uncompressed model with the same communication precision. All runs use EP32 over $4$ nodes, balanced expert assignment and fixed output lengths and the KV cache sets the requests per GPU. Throughput speedup divides EchoMoE throughput by the uncompressed model's throughput, and latency speedups use the inverse ratio. TPOT uses the $10\%$ of requests with the latest first token (\Cref{app:serving-setup}).}
\label{tab:agent-shapes}
\end{table}

\subsection{Inference on Larger MoE Architectures}
\label{app:serving-large}

We test EchoMoE on bigger models such as Qwen3-235B-A22B, DeepSeek-V3 and Kimi-K2.6 with $s=4$, $r_d=d/4$ and $r_c=d/2$ (\Cref{tab:serving-large-settings}). Due to limited compute resources, we do not adapt these models and instead use the architecture configuration with randomly initialized weights. The results compare only throughput with the same KV cache size for both models. Since every run uses balanced expert assignment, random weights do not affect throughput.

\begin{table}[t]
\centering\tablestyle[4pt]
\begin{tabular}{@{}lccll@{}}
\toprule
& \multicolumn{2}{c}{Code width} & & \\
\cmidrule(lr){2-3}
Model & \makecell[b]{Dispatch\\$r_d$} & \makecell[b]{Combine\\$r_c$} & \makecell[b]{Expert\\weights} & \makecell[b]{KV cache\\per GPU} \\
\midrule
Qwen3-235B-A22B & $1024$ & $2048$ & BF16 & $28$ GiB \\
DeepSeek-V3 & $1792$ & $3584$ & FP8 & $12$ GiB \\
Kimi-K2.6 & $1792$ & $3584$ & INT4 & $9$ GiB \\
\bottomrule
\end{tabular}
\caption{\textbf{Inference settings for larger MoE architectures.} All configurations use stride $4$, $r_d=d/4$ and $r_c=d/2$. EchoMoE and the uncompressed model use the same expert weight precision and KV cache size. For Kimi-K2.6, only the routed expert weights are INT4 while activations and compression parameters remain in BF16.}
\label{tab:serving-large-settings}
\end{table}

\textbf{Execution settings.} All runs use random weights with balanced expert assignment, EP32, TP1 and DP32 over $4$ nodes, $4096/256$ input/output tokens, $128$ requests per GPU and at most $2048$ tokens per GPU in each forward step. Expert computation uses Triton BF16 kernels for Qwen3-235B-A22B, Triton block-scaled FP8 kernels for DeepSeek-V3 and Marlin INT4 kernels for Kimi-K2.6. EchoMoE and the uncompressed model share the expert backend and CUDA graph settings, and both choose dispatch and combine kernels per call from a cost table measured for each model's hidden size, expert count and top-$k$. At reference layers, EchoMoE dispatches the full-width states to the selected experts with the chosen kernels and all-gathers them to all ranks with NCCL on a second stream, which overlaps the expert computation. DeepSeek-V3 sends these states in FP8 with per-block scales. The shared experts in DeepSeek-V3 and Kimi-K2.6 run on the main stream in both models. EchoMoE gives data-parallel padding rows no routes while the uncompressed model routes them under the routing simulator.  The KV cache holds only $32$ to $42$ full requests per GPU depending on the model, so most of the $128$ requests per GPU wait for earlier ones to finish.

\textbf{EchoMoE improves throughput by $1.09$--$1.24\times$ on larger architectures.} At about $2\times$ RDMA reduction, Qwen3-235B-A22B, DeepSeek-V3 and Kimi-K2.6 improve throughput by $1.24\times$, $1.15\times$ and $1.09\times$ with BF16 communication and by $1.21\times$, $1.19\times$ and $1.10\times$ with FP8 (\Cref{tab:serving-large}). FP8 communication reduces the time for every batch and changes EchoMoE's speedup by at most $0.04$. For comparison, Qwen3-30B-A3B $(4,512,1024)$ reaches $1.17\times$ with BF16 communication at a similar $2.08\times$ reduction (\Cref{tab:serving-workloads}).

\begin{table}[t]
\centering\tablestyle[4pt]
\begin{tabular}{@{}lrrrrrr@{}}
\toprule
& \multicolumn{3}{c}{BF16 communication} & \multicolumn{3}{c}{FP8 communication} \\
\cmidrule(lr){2-4}\cmidrule(lr){5-7}
Model (RDMA reduction) & Uncompressed & EchoMoE & Speedup & Uncompressed & EchoMoE & Speedup \\
\midrule
Qwen3-235B-A22B ($2.07\times$) & $256.5$ & $207.7$ & $1.235$ & $244.2$ & $202.2$ & $1.208$ \\
DeepSeek-V3 ($2.06\times$) & $378.7$ & $329.3$ & $1.150$ & $332.0$ & $279.5$ & $1.188$ \\
Kimi-K2.6 ($2.07\times$) & $259.8$ & $238.3$ & $1.090$ & $249.8$ & $226.6$ & $1.102$ \\
\bottomrule
\end{tabular}
\caption{\textbf{Throughput for larger MoE architectures.} Full-sized models with random weights run $4096/256$ input/output tokens with $128$ requests per GPU, using EP32 over $4$ nodes and balanced expert assignment. Uncompressed and EchoMoE columns report batch time in seconds, the time to finish all requests averaged over $2$ repetitions. Speedup compares EchoMoE with the uncompressed model under the same communication precision.}
\label{tab:serving-large}
\end{table}

\begin{figure}[t]
\centering
\includegraphics[width=0.6\linewidth]{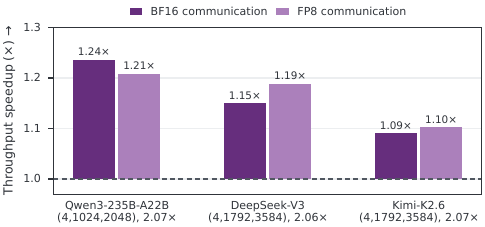}
\caption{\textbf{Throughput speedup for larger MoE architectures.} Full-sized models with random weights run $4096/256$ input/output tokens with $128$ requests per GPU, using EP32 over $4$ nodes and balanced expert assignment. Speedups compare EchoMoE with the uncompressed model under the same communication precision.}
\label{fig:larger-models}
\end{figure}

\subsection{Inference on Kanana-2 and Gemma 4}
\label{app:serving-adapted}

We serve the adapted Kanana-2 $(4,512,1024)$ and Gemma 4 26B-A4B $(4,768,1408)$ from \Cref{tab:domain-results} with the setup described in \Cref{app:serving-setup}. In both the uncompressed and EchoMoE variants, Kanana-2's shared experts and Gemma 4's dense MLP run on a second stream, concurrently with the routed experts, and both use the same KV cache size. DeepEP's combine requires a multiple of $256$ values, so we zero-pad Gemma 4's combine codes from $1408$ to $1536$, which sends $9\%$ more combine bytes without changing the outputs.

\textbf{EchoMoE speeds up Kanana-2 and Gemma 4 by up to $1.16\times$.} Across different workloads and communication precisions, EchoMoE improves throughput by $1.07$--$1.16\times$ and TTFT by $1.09$--$1.17\times$ (\Cref{tab:serving-adapted}). At $4096/256$ with BF16 communication, Kanana-2 reaches $1.15\times$ similar to the $1.17\times$ of Qwen3-30B-A3B with the same configuration $(4,512,1024)$. Decode follows the same trend as Qwen3-30B-A3B, improving at $128$ requests per GPU or more and slowing slightly at $4$ requests per GPU.

\begin{table}[t]
\centering\tablestyle[4pt]
\begin{tabular}{@{}lrrrrrr@{}}
\toprule
& \multicolumn{3}{c}{BF16 communication} & \multicolumn{3}{c}{FP8 communication} \\
\cmidrule(lr){2-4}\cmidrule(lr){5-7}
Workload (requests per GPU) & Throughput & TTFT & TPOT & Throughput & TTFT & TPOT \\
\midrule
\multicolumn{7}{@{}l}{\emph{Kanana-2-30B-A3B, $(4,512,1024)$}} \\
1024/128 ($512$) & $1.15$ & $1.16$ & $1.13$ & $1.16$ & $1.17$ & $1.10$ \\
4096/256 ($128$) & $1.15$ & $1.17$ & $1.06$ & $1.14$ & $1.17$ & $1.04$ \\
16384/512 ($32$) & $1.11$ & $1.15$ & $1.00$ & $1.10$ & $1.15$ & $0.99$ \\
128/1024 ($512$) & $1.13$ & $1.17$ & $1.12$ & $1.11$ & $1.17$ & $1.10$ \\
\midrule
\multicolumn{7}{@{}l}{\emph{Gemma 4 26B-A4B, $(4,768,1408)$}} \\
1024/128 ($512$) & $1.13$ & $1.15$ & $1.07$ & $1.11$ & $1.13$ & $1.06$ \\
4096/256 ($128$) & $1.14$ & $1.17$ & $1.07$ & $1.12$ & $1.15$ & $1.04$ \\
16384/512 ($32$) & $1.12$ & $1.16$ & $1.01$ & $1.10$ & $1.14$ & $1.00$ \\
128000/512 ($4$) & $1.09$ & $1.11$ & $0.95$ & $1.07$ & $1.09$ & $0.96$ \\
128/1024 ($512$) & $1.08$ & $1.17$ & $1.07$ & $1.10$ & $1.16$ & $1.10$ \\
\bottomrule
\end{tabular}
\caption{\textbf{Throughput and latency for adapted Kanana-2 and Gemma 4.} All runs use EP32 over $4$ nodes and balanced expert assignment. Throughput speedup divides EchoMoE throughput by the uncompressed model's throughput, and latency speedups use the inverse ratio. TPOT uses the $10\%$ of requests with the latest first token (\Cref{app:serving-setup}). Kanana-2 supports $32$K positions, so it doesn't include $128000/512$ workload.}
\label{tab:serving-adapted}
\end{table}

\clearpage

\section{Task Performance of All Configurations}
\label{app:pertask}

\Cref{tab:pertask-models} reports the per-benchmark results behind \Cref{tab:domain-results} for each model's adapted configuration. While consistent parsing and evaluation pipeline is used across models, several scores mainly reflect output format rather than exact task performance (marked in \Cref{tab:pertask-models}). Excluding these rows changes each model's overall difference marginally in both directions: from $-3.8$ to $-4.1$ pp for Qwen3-30B-A3B, $-4.4$ to $-3.8$ for Gemma 4, $-2.4$ to $-2.6$ for Kanana-2, $-4.6$ to $-3.5$ for Nemotron 3.5 and $-5.9$ to $-6.3$ for Moonlight. We report the full result in the paper without excluding any of them.

\begin{table}[htbp]
\centering\tablestyle[2pt]\footnotesize
\resizebox{\linewidth}{!}{%
\begin{tabular}{@{}llrrrrrrrrrr@{}}
\toprule
 & & \multicolumn{2}{c}{Qwen3} & \multicolumn{2}{c}{Gemma 4} & \multicolumn{2}{c}{Kanana-2} & \multicolumn{2}{c}{Nemotron 3.5} & \multicolumn{2}{c}{Moonlight} \\
\cmidrule(lr){3-4}\cmidrule(lr){5-6}\cmidrule(lr){7-8}\cmidrule(lr){9-10}\cmidrule(lr){11-12}
Benchmark & Prompt & Unc. & $(4,512,768)$ & Unc. & $(4,768,1408)$ & Unc. & $(4,512,1024)$ & Unc. & $(4,768,1408)$ & Unc. & $(4,512,1024)$ \\
\midrule
\multicolumn{12}{@{}l}{\emph{Knowledge and reasoning}} \\
MMLU & Few-shot & $83.3$ & $-6.2$ & $78.4$ & $-6.4$ & $76.9$ & $-3.9$ & $80.1$ & $-3.9$ & $70.5$ & $-6.8$ \\
ARC-Easy & Few-shot & $88.7$ & $-0.9$ & $91.7$ & $-3.3$ & $88.8$ & $-1.5$ & $90.3$ & $-1.3$ & $87.0$ & $-3.4$ \\
ARC-Challenge & Few-shot & $67.7$ & $-4.4$ & $73.5$ & $-7.9$ & $66.4$ & $-2.0$ & $69.9$ & $-1.9$ & $66.1$ & $-7.9$ \\
HellaSwag & Few-shot & $82.7$ & $-5.4$ & $84.6$ & $-4.0$ & $82.7$ & $-3.5$ & $85.9$ & $-3.1$ & $81.4$ & $-6.6$ \\
MMLU-Pro & Few-shot & $62.8$ & $-9.8$ & $52.8$ & $-8.1$ & $55.1$ & $-7.1$ & $66.8$ & $-12.6$$^\S$ & $43.0$ & $-10.5$$^\S$ \\
MMLU-Redux & Few-shot & $85.9$ & $-5.4$ & $79.5$ & $-5.4$ & $77.6$ & $-3.0$ & $73.5$ & $-13.3$ & $73.7$ & $-7.0$ \\
BBH & Few-shot & $63.6$$^\ddagger$ & $+11.2$$^\ddagger$ & $76.8$ & $-4.7$ & $77.1$ & $-4.0$ & $70.8$ & $+1.8$ & $65.1$ & $-9.5$ \\
MMLU & Chat & $82.8$ & $-5.4$ & $79.0$ & $-8.2$ & $70.8$ & $-1.4$ & $79.1$ & $-2.2$ & $63.1$ & $-9.1$ \\
MMLU-Pro & Chat & $66.0$ & $-5.7$ & $61.2$ & $-9.9$ & $48.4$ & $-0.8$ & $56.7$ & $-4.5$ & $41.3$ & $-6.4$ \\
MMLU-Redux & Chat & $87.1$ & $-5.0$ & $82.2$ & $-7.4$ & $75.0$ & $-1.5$ & $83.1$ & $-2.4$ & $66.8$ & $-7.6$ \\
ARC-Easy & Chat & $98.2$ & $-0.5$ & $97.3$ & $-1.1$ & $97.0$ & $-1.2$ & $96.5$ & $+0.2$ & $83.5$ & $-3.6$ \\
ARC-Challenge & Chat & $95.7$ & $-2.0$ & $92.7$ & $-2.7$ & $90.3$ & $-2.0$ & $92.7$ & $-1.4$ & $78.5$ & $-5.8$ \\
HellaSwag & Chat & $84.6$ & $-6.2$$^\S$ & $51.1$ & $+3.6$ & $59.8$ & $-2.5$ & $55.6$ & $-7.5$ & $42.2$ & $-9.5$ \\
BBH & Chat & $66.0$ & $-7.0$ & $54.2$ & $-4.3$ & $52.0$ & $-2.5$ & $59.9$ & $-1.8$ & $47.2$ & $-5.1$ \\
GPQA-Diamond & Chat & $48.3$ & $-7.6$ & $32.4$ & $-2.4$ & $5.8$$^\dagger$ & $+10.6$$^\dagger$ & $13.5$$^\dagger$ & $-2.1$$^\dagger$ & $34.4$ & $-5.4$ \\
\midrule
\multicolumn{12}{@{}l}{\emph{Mathematics}} \\
GSM8K & Few-shot & $90.1$ & $-4.5$ & $85.5$ & $-2.4$ & $80.1$ & $-0.5$ & $89.0$ & $-3.9$ & $73.9$ & $-0.5$ \\
MATH-500 & Few-shot & $52.6$$^\dagger$ & $-12.6$$^\dagger$ & $43.4$$^\dagger$ & $-19.6$$^\dagger$ & $42.0$$^\dagger$ & $-10.0$$^\dagger$ & $56.8$$^\dagger$ & $-9.6$$^\dagger$ & $34.0$$^\dagger$ & $-9.2$$^\dagger$ \\
GSM8K & Chat & $92.4$ & $-0.1$ & $89.8$ & $+0.2$ & $91.1$ & $-2.3$ & $85.0$ & $-31.2$$^\dagger$ & $70.1$ & $+4.7$$^\dagger$ \\
MATH-500 & Chat & $81.2$ & $-3.0$ & $61.6$ & $-6.2$ & $66.8$ & $-10.2$ & $82.4$ & $-6.8$ & $58.4$ & $-14.4$ \\
AIME 2024 & Chat & $20.4$ & $-3.6$$^\S$ & $7.1$ & $-1.9$ & $7.2$ & $-1.2$ & $19.5$ & $-5.1$ & $3.3$ & $-1.7$ \\
AIME 2025 & Chat & $12.9$ & $-0.6$ & $1.8$ & $-0.3$ & $2.3$ & $-0.8$ & $13.2$ & $-2.6$ & $1.8$ & $-1.4$ \\
HMMT Feb.~2025 & Chat & $5.0$ & $+0.2$ & $0.7$ & $-0.4$ & $1.2$ & $-0.6$ & $7.7$ & $-2.7$ & $1.2$ & $-0.8$ \\
\midrule
\multicolumn{12}{@{}l}{\emph{Coding}} \\
HumanEval+ & Chat & $77.4$ & $-6.1$ & $72.0$ & $-0.6$ & $72.6$ & $-2.4$ & $79.3$ & $-1.8$ & $68.3$ & $-9.8$ \\
MBPP+ & Chat & $73.8$ & $-4.8$ & $68.8$ & $-6.6$ & $67.2$ & $-6.6$ & $70.4$ & $-2.1$ & $63.2$ & $-9.5$ \\
LiveCodeBench & Chat & $26.5$ & $-3.7$ & $20.8$ & $-2.1$ & $20.5$ & $-2.6$ & $9.3$$^\dagger$ & $-0.7$$^\dagger$ & $9.4$$^\dagger$ & $-2.9$$^\dagger$ \\
\midrule
\multicolumn{12}{@{}l}{\emph{Instruction following}} \\
IFEval & Chat & $80.4$ & $-1.1$ & $81.9$ & $-1.7$ & $78.6$ & $-0.6$ & $79.9$ & $-1.5$ & $77.3$ & $-5.9$ \\
IFBench & Chat & $30.7$ & $-3.3$ & $32.7$ & $-4.7$ & $27.3$ & $-0.3$ & $30.0$ & $-0.3$ & $25.3$ & $-2.7$ \\
\midrule
Overall & & $66.9$ & $-3.8$ & $61.2$ & $-4.4$ & $58.5$ & $-2.4$ & $62.8$ & $-4.6$ & $53.0$ & $-5.9$ \\
\bottomrule
\end{tabular}}
\caption{\textbf{Per-benchmark task performance after full-model adaptation across models.} Uncompressed (Unc.) score of each model and the change for each EchoMoE configuration in percentage points, using each benchmark's primary metric in percent. Each model uses its configuration from \Cref{tab:domain-results}, with $(4,512,768)$ for Qwen3-30B-A3B. $^\dagger$More than $20\%$ of answers could not be extracted in the uncompressed or EchoMoE model, or the unparseable share changed by more than $10$ pp, so the score mainly reflects output format. $^\ddagger$Uncompressed Qwen3-30B-A3B often keeps writing on the answer line, so lm-eval's strict exact match captures extra text: $11.3\%$ of its answers are captured with more than $40$ characters, against $1.6\%$ for EchoMoE, while correct answers are typically $3$ characters. This scoring artifact accounts for most of the gain. $^\S$EchoMoE's own output length changes sharply: it is cut off at the generation limit at least $10$ pp more often, or its median output is at least $3\times$ longer or shorter, for example answering without chain-of-thought. These losses reflect the adapted model's behavior.}
\label{tab:pertask-models}
\end{table}

\Cref{tab:allgroups,tab:alladapted} report task performance for all $40$ joint configurations trained in dev mode and the $19$ with full model adaptation. Dispatch-only and combine-only results are in \Cref{tab:dispatch-grid,tab:correction}. Domain scores average the per-task scores while overall is the mean of all $27$ tasks and prompt format pairs (\Cref{tab:tasks}).

{\tablestyle[3pt]\renewcommand{\arraystretch}{1.0}\captionsetup{position=below}
\begin{longtable}{@{}ccccrrrr@{\hspace{10pt}}l@{}}
\toprule
\multicolumn{3}{c}{Configuration} & & \multicolumn{5}{c}{Task Performance (\%) $\uparrow$} \\
\cmidrule(lr){1-3}\cmidrule(lr){5-9}
\makecell[b]{Stride\\$s$} & \makecell[b]{Dispatch\\$r_d$} & \makecell[b]{Combine\\$r_c$} & \makecell[b]{RDMA\\Reduction} & \makecell[b]{Knowledge\\\&\ Reasoning} & Math & Code & \makecell[b]{Instruction\\Following} & \makecell[b]{Overall\\\textcolor{tabchange}{($\Delta$)}} \\
\midrule
\endfirsthead
\multicolumn{9}{@{}l}{\emph{\Cref{tab:allgroups}, continued}}\\[2pt]
\toprule
\multicolumn{3}{c}{Configuration} & & \multicolumn{5}{c}{Task Performance (\%) $\uparrow$} \\
\cmidrule(lr){1-3}\cmidrule(lr){5-9}
\makecell[b]{Stride\\$s$} & \makecell[b]{Dispatch\\$r_d$} & \makecell[b]{Combine\\$r_c$} & \makecell[b]{RDMA\\Reduction} & \makecell[b]{Knowledge\\\&\ Reasoning} & Math & Code & \makecell[b]{Instruction\\Following} & \makecell[b]{Overall\\\textcolor{tabchange}{($\Delta$)}} \\
\midrule
\endhead
\midrule
\multicolumn{9}{r@{}}{\emph{continued on the next page}}\\
\endfoot
\bottomrule
\caption{\textbf{Task performance for all dev mode joint configurations.} $\Delta$ denotes the change in overall task performance from the uncompressed model ($66.92\%$), in percentage points. RDMA reduction compares estimated BF16 bytes at EP32 over $4$ nodes, including reference transfers.}\label{tab:allgroups}
\endlastfoot
\multicolumn{3}{@{}l}{\emph{Qwen3-30B-A3B ($d=2048$)}} & $1.00\times$ & 77.6 & 50.7 & 59.2 & 55.6 & 66.9 \\
\midrule
\multicolumn{9}{@{}l}{\emph{EchoMoE}} \\
4 & 1280 & 1280 & $1.46\times$ & 75.8 & 50.0 & 56.9 & 53.8 & 65.4\,\scorechange{$-$1.5} \\
6 & 1280 & 1280 & $1.50\times$ & 75.5 & 49.3 & 55.6 & 53.2 & 64.8\,\scorechange{$-$2.1} \\
8 & 1280 & 1280 & $1.53\times$ & 74.8 & 49.6 & 56.0 & 54.6 & 64.7\,\scorechange{$-$2.2} \\
12 & 1280 & 1280 & $1.55\times$ & 74.7 & 48.9 & 55.5 & 53.4 & 64.3\,\scorechange{$-$2.6} \\
4 & 1024 & 1280 & $1.57\times$ & 75.7 & 50.5 & 55.9 & 54.9 & 65.4\,\scorechange{$-$1.5} \\
8 & 768 & 1280 & $1.83\times$ & 71.2 & 48.3 & 53.7 & 54.6 & 62.0\,\scorechange{$-$4.9} \\
4 & 512 & 1280 & $1.84\times$ & 73.8 & 47.9 & 55.1 & 52.0 & 63.4\,\scorechange{$-$3.5} \\
8 & 1024 & 1024 & $1.86\times$ & 71.5 & 46.6 & 54.9 & 54.6 & 61.9\,\scorechange{$-$5.0} \\
4 & 768 & 1024 & $1.89\times$ & 73.1 & 48.5 & 54.2 & 53.0 & 63.1\,\scorechange{$-$3.8} \\
12 & 1024 & 1024 & $1.90\times$ & 69.7 & 46.7 & 53.2 & 51.9 & 60.6\,\scorechange{$-$6.3} \\
6 & 768 & 1024 & $2.01\times$ & 71.0 & 45.4 & 51.1 & 54.0 & 60.9\,\scorechange{$-$6.0} \\
6 & 1024 & 768 & $2.05\times$ & 70.0 & 45.2 & 49.8 & 51.8 & 59.9\,\scorechange{$-$7.0} \\
8 & 768 & 1024 & $2.07\times$ & 69.0 & 45.8 & 53.4 & 51.9 & 60.0\,\scorechange{$-$6.9} \\
4 & 512 & 1024 & $2.08\times$ & 71.9 & 46.6 & 53.3 & 53.9 & 62.0\,\scorechange{$-$5.0} \\
8 & 1024 & 768 & $2.11\times$ & 68.2 & 44.5 & 51.0 & 53.4 & 59.0\,\scorechange{$-$7.9} \\
12 & 768 & 1024 & $2.14\times$ & 67.6 & 43.7 & 51.3 & 50.8 & 58.3\,\scorechange{$-$8.6} \\
12 & 1024 & 768 & $2.16\times$ & 66.7 & 44.0 & 48.9 & 51.0 & 57.7\,\scorechange{$-$9.2} \\
8 & 768 & 768 & $2.38\times$ & 65.4 & 43.4 & 49.4 & 52.8 & 56.9\,\scorechange{$-$10.0} \\
4 & 512 & 768 & $2.39\times$ & 68.2 & 43.7 & 48.3 & 52.0 & 58.4\,\scorechange{$-$8.5} \\
12 & 768 & 768 & $2.47\times$ & 63.7 & 41.9 & 48.0 & 49.5 & 55.2\,\scorechange{$-$11.7} \\
6 & 256 & 1024 & $2.54\times$ & 64.5 & 38.5 & 44.7 & 51.0 & 54.6\,\scorechange{$-$12.4} \\
6 & 512 & 768 & $2.61\times$ & 65.6 & 40.2 & 45.5 & 51.2 & 55.7\,\scorechange{$-$11.2} \\
4 & 512 & 512 & $2.81\times$ & 62.8 & 37.8 & 42.6 & 49.4 & 53.1\,\scorechange{$-$13.8} \\
12 & 512 & 768 & $2.87\times$ & 60.8 & 36.9 & 43.7 & 50.8 & 52.0\,\scorechange{$-$14.9} \\
12 & 768 & 512 & $2.92\times$ & 58.9 & 36.8 & 42.9 & 46.5 & 50.5\,\scorechange{$-$16.4} \\
6 & 256 & 768 & $3.02\times$ & 61.6 & 35.7 & 40.3 & 49.1 & 51.6\,\scorechange{$-$15.3} \\
6 & 512 & 512 & $3.12\times$ & 60.5 & 36.4 & 43.3 & 48.3 & 51.4\,\scorechange{$-$15.5} \\
6 & 256 & 512 & $3.72\times$ & 56.4 & 32.2 & 37.5 & 46.2 & 47.3\,\scorechange{$-$19.7} \\
8 & 256 & 512 & $4.03\times$ & 54.0 & 29.4 & 38.2 & 46.6 & 45.3\,\scorechange{$-$21.6} \\
8 & 512 & 256 & $4.16\times$ & 47.0 & 26.7 & 32.2 & 39.5 & 39.5\,\scorechange{$-$27.4} \\
\midrule
\multicolumn{9}{@{}l}{\emph{Plain compression}} \\
-- & 1280 & 1280 & $1.60\times$ & 71.6 & 47.5 & 53.6 & 52.7 & 61.9\,\scorechange{$-$5.0} \\
-- & 1280 & 1024 & $1.78\times$ & 69.5 & 45.7 & 53.4 & 51.9 & 60.2\,\scorechange{$-$6.7} \\
-- & 1024 & 1024 & $2.00\times$ & 65.5 & 43.2 & 50.8 & 49.8 & 56.9\,\scorechange{$-$10.0} \\
-- & 1280 & 768 & $2.00\times$ & 65.8 & 44.2 & 50.6 & 50.3 & 57.3\,\scorechange{$-$9.6} \\
-- & 1024 & 768 & $2.29\times$ & 62.2 & 40.8 & 45.3 & 51.3 & 54.0\,\scorechange{$-$12.9} \\
-- & 1024 & 512 & $2.67\times$ & 58.8 & 36.8 & 43.7 & 47.4 & 50.6\,\scorechange{$-$16.3} \\
-- & 768 & 768 & $2.67\times$ & 59.1 & 36.1 & 41.7 & 47.3 & 50.3\,\scorechange{$-$16.6} \\
-- & 768 & 512 & $3.20\times$ & 53.9 & 33.1 & 38.9 & 44.1 & 46.1\,\scorechange{$-$20.8} \\
-- & 512 & 512 & $4.00\times$ & 47.9 & 26.9 & 33.8 & 42.3 & 40.4\,\scorechange{$-$26.5} \\
-- & 768 & 256 & $4.00\times$ & 45.7 & 25.3 & 28.8 & 39.9 & 38.1\,\scorechange{$-$28.8} \\
\end{longtable}
}

\clearpage
{\tablestyle[2pt]\captionsetup{position=below}
\begin{longtable}{@{}ccccrrrr@{\hspace{6pt}}lr@{}}
\toprule
\multicolumn{3}{c}{Configuration} & & \multicolumn{5}{c}{Task Performance (\%) $\uparrow$} &  \\
\cmidrule(lr){1-3}\cmidrule(lr){5-9}
\makecell[b]{Stride\\$s$} & \makecell[b]{Dispatch\\$r_d$} & \makecell[b]{Combine\\$r_c$} & \makecell[b]{RDMA\\Reduction} & \makecell[b]{Knowledge\\\&\ Reasoning} & Math & Code & \makecell[b]{Instruction\\Following} & \makecell[b]{Overall\\\textcolor{tabchange}{($\Delta$)}} & \makecell[b]{$\Delta_{\mathrm{adapt}}$\\(pp)} \\
\midrule
\endfirsthead
\multicolumn{10}{@{}l}{\emph{\Cref{tab:alladapted}, continued}}\\[2pt]
\toprule
\multicolumn{3}{c}{Configuration} & & \multicolumn{5}{c}{Task Performance (\%) $\uparrow$} &  \\
\cmidrule(lr){1-3}\cmidrule(lr){5-9}
\makecell[b]{Stride\\$s$} & \makecell[b]{Dispatch\\$r_d$} & \makecell[b]{Combine\\$r_c$} & \makecell[b]{RDMA\\Reduction} & \makecell[b]{Knowledge\\\&\ Reasoning} & Math & Code & \makecell[b]{Instruction\\Following} & \makecell[b]{Overall\\\textcolor{tabchange}{($\Delta$)}} & \makecell[b]{$\Delta_{\mathrm{adapt}}$\\(pp)} \\
\midrule
\endhead
\midrule
\multicolumn{10}{r@{}}{\emph{continued on the next page}}\\
\endfoot
\bottomrule
\caption{\textbf{Task performance for all adapted joint configurations.} $\Delta$ denotes the change in overall task performance from the uncompressed model and $\Delta_{\mathrm{adapt}}$ the change from dev mode for the same configuration, in percentage points.}\label{tab:alladapted}
\endlastfoot
\multicolumn{3}{@{}l}{\emph{Qwen3-30B-A3B ($d=2048$)}} & $1.00\times$ & 77.6 & 50.7 & 59.2 & 55.6 & 66.9 & -- \\
\midrule
\multicolumn{10}{@{}l}{\emph{EchoMoE}} \\
4 & 1280 & 1280 & $1.46\times$ & 76.2 & 49.1 & 58.1 & 54.5 & 65.5\,\scorechange{$-$1.4} & $+0.2$ \\
6 & 1280 & 1280 & $1.50\times$ & 75.6 & 49.1 & 57.4 & 53.8 & 65.1\,\scorechange{$-$1.8} & $+0.3$ \\
12 & 1280 & 1280 & $1.55\times$ & 73.7 & 49.6 & 57.4 & 53.1 & 64.1\,\scorechange{$-$2.8} & $-0.2$ \\
4 & 512 & 1280 & $1.84\times$ & 75.3 & 49.3 & 56.8 & 53.7 & 64.9\,\scorechange{$-$2.0} & $+1.5$ \\
6 & 768 & 1024 & $2.01\times$ & 74.0 & 47.8 & 55.2 & 52.3 & 63.5\,\scorechange{$-$3.4} & $+2.6$ \\
4 & 512 & 1024 & $2.08\times$ & 74.3 & 47.8 & 55.2 & 53.5 & 63.8\,\scorechange{$-$3.1} & $+1.8$ \\
4 & 512 & 768 & $2.39\times$ & 73.6 & 47.2 & 54.4 & 53.3 & 63.1\,\scorechange{$-$3.8} & $+4.7$ \\
6 & 256 & 1024 & $2.54\times$ & 70.4 & 45.8 & 51.8 & 52.6 & 60.6\,\scorechange{$-$6.3} & $+6.1$ \\
6 & 512 & 768 & $2.61\times$ & 71.7 & 46.4 & 52.4 & 51.9 & 61.5\,\scorechange{$-$5.4} & $+5.8$ \\
4 & 512 & 512 & $2.81\times$ & 70.0 & 45.6 & 53.4 & 52.5 & 60.6\,\scorechange{$-$6.4} & $+7.5$ \\
12 & 768 & 512 & $2.92\times$ & 68.3 & 45.5 & 52.1 & 51.8 & 59.4\,\scorechange{$-$7.6} & $+8.9$ \\
6 & 256 & 768 & $3.02\times$ & 68.3 & 42.5 & 50.4 & 50.1 & 58.3\,\scorechange{$-$8.6} & $+6.7$ \\
6 & 256 & 512 & $3.72\times$ & 65.8 & 41.6 & 51.6 & 51.2 & 56.9\,\scorechange{$-$10.1} & $+9.6$ \\
8 & 256 & 512 & $4.03\times$ & 64.2 & 40.9 & 50.7 & 51.0 & 55.7\,\scorechange{$-$11.2} & $+10.4$ \\
\midrule
\multicolumn{10}{@{}l}{\emph{Plain compression}} \\
-- & 1280 & 1024 & $1.78\times$ & 72.6 & 47.5 & 55.7 & 53.5 & 62.8\,\scorechange{$-$4.1} & $+2.5$ \\
-- & 1024 & 1024 & $2.00\times$ & 70.4 & 47.4 & 55.8 & 53.5 & 61.6\,\scorechange{$-$5.3} & $+4.6$ \\
-- & 768 & 768 & $2.67\times$ & 66.2 & 44.3 & 52.8 & 51.8 & 58.0\,\scorechange{$-$8.9} & $+7.7$ \\
-- & 512 & 512 & $4.00\times$ & 59.0 & 37.6 & 47.1 & 49.6 & 51.4\,\scorechange{$-$15.5} & $+11.0$ \\
-- & 768 & 256 & $4.00\times$ & 58.0 & 38.1 & 45.1 & 48.6 & 50.7\,\scorechange{$-$16.2} & $+12.6$ \\
\end{longtable}
}

\end{document}